\documentclass[letterpaper, 10 pt, journal, twoside]{IEEEtran}

\usepackage{siunitx}
\usepackage{relsize}
\usepackage{ifthen}
\usepackage[colorinlistoftodos]{todonotes}

\usepackage[caption=false]{subfig}

\usepackage{graphics} %
\usepackage{graphicx}
\usepackage{rotating}
\usepackage{color}
\usepackage{enumerate}
\usepackage[T1]{fontenc}
\usepackage{psfrag}
\usepackage{epsfig} %
\usepackage{booktabs}
\usepackage{graphicx,url}
\usepackage{multirow}
\usepackage{array}
\usepackage{latexsym}
\usepackage{amsfonts}
\usepackage{amsmath}
\usepackage{amssymb}
\usepackage{mathtools}
\usepackage{amsthm}
\usepackage{xstring}
\usepackage{algorithm} 
\usepackage[noEnd=true, indLines=true, commentColor=NavyBlue, endLComment=~, beginComment=//~]{algpseudocodex} 
\usepackage{xpatch}
\usepackage{multirow}
\usepackage{xcolor}
\usepackage[dvipsnames]{xcolor}
\usepackage{prettyref}
\usepackage{flexisym}
\usepackage{bigdelim}
\usepackage{listings}

\usepackage{enumitem}
\usepackage{xspace}
\usepackage{bm}
\graphicspath{{./figures/}}
\usepackage{tikz}
\usetikzlibrary{matrix,calc,arrows.meta,positioning,fit,backgrounds}
\usepackage[acronym]{glossaries}
\usepackage[bookmarks=true]{hyperref}
\usepackage[capitalize]{cleveref}
\usepackage{subcaption}
\usepackage{aliascnt}
\usepackage{threeparttable}
\usepackage[normalem]{ulem}

\usepackage{fancyhdr}

\hypersetup{hidelinks}

\usepackage[markup=underlined]{changes} %
\definechangesauthor[name=Cathy, color=teal]{Cathy}
\definechangesauthor[name=Jiarui, color=purple]{Jiarui}

\usepackage{cite}

\newrefformat{prob}{Problem\,\ref{#1}}
\newrefformat{def}{Definition\,\ref{#1}}
\newrefformat{sec}{Section\,\ref{#1}}
\newrefformat{sub}{Section\,\ref{#1}}
\newrefformat{prop}{Proposition\,\ref{#1}}
\newrefformat{app}{Appendix\,\ref{#1}}
\newrefformat{alg}{Algorithm\,\ref{#1}}
\newrefformat{cor}{Corollary\,\ref{#1}}
\newrefformat{thm}{Theorem\,\ref{#1}}
\newrefformat{lem}{Lemma\,\ref{#1}}
\newrefformat{fig}{Fig.\,\ref{#1}}
\newrefformat{tab}{Table\,\ref{#1}}

\theoremstyle{definition}
\newtheorem{theorem}{Theorem}[section]

\theoremstyle{definition}
\newtheorem{problem}{Problem} 
\crefname{problem}{Problem}{Problems}

\crefname{trule}{Rule}{Rules}

\newtheorem{corollary}{Corollary}[section]
\crefname{corollary}{Corollary}{Corollaries}

\crefname{conjecture}{Conjecture}{Conjectures}

\newtheorem{lemma}{Lemma}[section]
\crefname{lemma}{Lemma}{Lemmas}

\crefname{assumption}{Assumption}{Assumptions}

\newtheorem{definition}{Definition}[section]
\crefname{definition}{Definition}{Definitions}

\newtheorem{proposition}{Proposition}[section]
\crefname{proposition}{Proposition}{Propositions}

\newtheorem{remark}{Remark}[section]
\crefname{remark}{Remark}{Remarks}

\crefname{example}{Example}{Examples}

\makeatletter\@addtoreset{problem}{section}\makeatother      %

\newcommand{\bdmath}{\begin{dmath}}
\newcommand{\edmath}{\end{dmath}}
\newcommand{\beq}{\begin{equation}}
\newcommand{\eeq}{\end{equation}}
\newcommand{\bdm}{\begin{displaymath}}
\newcommand{\edm}{\end{displaymath}}
\newcommand{\bea}{\begin{eqnarray}}
\newcommand{\eea}{\end{eqnarray}}
\newcommand{\beal}{\beq \begin{array}{ll}}
\newcommand{\eeal}{\end{array} \eeq}
\newcommand{\beas}{\begin{eqnarray*}}
\newcommand{\eeas}{\end{eqnarray*}}
\newcommand{\ba}{\begin{array}}
\newcommand{\ea}{\end{array}}
\newcommand{\bit}{\begin{itemize}}
\newcommand{\eit}{\end{itemize}}
\newcommand{\ben}{\begin{enumerate}}
\newcommand{\een}{\end{enumerate}}

\newcommand{\calI}{{\cal I}}

\newcommand{\calM}{{\cal M}}

\newcommand{\calU}{{\cal U}}

\newcommand{\hide}[1]{}

\newcommand{\hiddenText}{{\color{gray} hidden text.}}
\newcommand{\hideWithText}[1]{\hiddenText}

\newcommand{\norm}[1]{\left\| #1 \right\|}

\newcommand{\blue}[1]{{\color{blue}#1}}

\newcommand{\linkToPdf}[1]{\href{#1}{\blue{(pdf)}}}
\newcommand{\linkToPpt}[1]{\href{#1}{\blue{(ppt)}}}
\newcommand{\linkToCode}[1]{\href{#1}{\blue{(code)}}}
\newcommand{\linkToWeb}[1]{\href{#1}{\blue{(web)}}}
\newcommand{\linkToVideo}[1]{\href{#1}{\blue{(video)}}}
\newcommand{\linkToMedia}[1]{\href{#1}{\blue{(media)}}}
\newcommand{\award}[1]{\xspace} %

\newcommand{\myparagraph}[1]{\noindent\textbf{#1}}

\renewcommand{\norm}[1]{\left\lVert #1 \right\rVert}

\newcommand{\bmat}{\left[ \begin{array}}
\newcommand{\emat}{\end{array}\right]}

\newcommand{\hx}{\hat{x}}

\newcommand{\eps}{\varepsilon}                    %
\newcommand{\dt}{\Delta t}                        %
\newcommand{\latt}{L_{\eps}}                      %
\newcommand{\moveset}{\calM}                      %
\newcommand{\clr}{\mathrm{cl}}                    %
\newcommand{\reps}{r_{\eps}}                      %
\newcommand{\sigk}{\sigma_{\phi}}                 %
\newcommand{\Tf}{T_{\mathrm{f}}}                  %
\newcommand{\tx}{\tilde{x}}                       %
\newcommand{\uset}{\calU}                         %
\newcommand{\freespace}{\mathcal F}               %
\newcommand{\freeconf}{\mathcal Q_{\mathrm{free}}} %

\newcommand{\forbidden}{\mathcal C}
\newcommand{\diamondbody}{\mathcal D}

\newcommand{\tup}[1]{\left( #1\right)}

\newacronym{acr:mapf}{MAPF}{Multi-Agent Path Finding}
\newacronym{acr:fico}{FICO}{Finite-Horizon Closed-Loop Factorization}
\newacronym{acr:soc}{SOC}{Sum of Cost}
\newacronym{acr:mpc}{MPC}{Model Predictive Control}
\newacronym{acr:rl}{RL}{Reinforcement Learning}
\newacronym{acr:ert}{ERT}{Execution Response Time}
\newacronym{acr:accbs}{ACCBS}{Anytime Closed-Loop Conflict-Based Search}
\newacronym{acr:cbs}{CBS}{Conflict-Based Search}
\newacronym{acr:pwm}{PWM}{Pulse-Width Modulation}
\newacronym{acr:takt}{TAKT}{Tickwise Actions to Kinodynamic Trajectories}
\newacronym{acr:kmampp}{KMAMPP}{Kinodynamic Multi-Agent Motion Planning Problem}

\graphicspath{{./figures_arxiv/}}

\newcommand{\figscale}{1}

\newboolean{anonymous}
\setboolean{anonymous}{false}

\begin{document}

\title{Real-Time Kinodynamic \\ Multi-Agent Motion Planning}

\ifthenelse{\boolean{anonymous}}{%
  \author{Anonymous Authors}%
}{
\author{Jiarui Li,
        Runyu Zhang,
        and Gioele Zardini
\thanks{The authors are with the Laboratory for Information and Decision Systems, Massachusetts Institute of Technology, Cambridge, MA, USA (e-mails: \{jiarui01, runyuzha, gzardini\}@mit.edu).}
\thanks{This work was supported by Prof. Zardini's grant from the MIT Amazon Science Hub, hosted in the Schwarzman College of Computing. Runyu Zhang was supported by the MIT Postdoctoral Fellowship
Program for Engineering Excellence.}
}
}

\maketitle

\begin{abstract}
Coordinating large robot fleets, from warehouses to quadrotor swarms, requires collision-free trajectories that satisfy dynamic constraints and stay within speed and acceleration bounds.
Existing kinodynamic multi-agent planners enforce these bounds \emph{during} the search or through trajectory optimization \emph{after} it, which limits fleet size and speed.
We present \gls{acr:takt}, a framework that accounts for the dynamics \emph{before} the search and reduces planning to a lattice search for double-integrator robots.
\gls{acr:takt} analytically bounds the clearance lost when a discrete plan is smoothed and reserves this margin on the lattice.
A closed-form convolution provably converts any valid plan into dynamically feasible trajectories with continuous-time collision avoidance.
For robots spanning many lattice vertices, we introduce a diamond formulation that permits one-tick sideways yielding and adapt the \gls{acr:mapf} solvers PIBT and LaCAM to it.
With 50 robots, our planners run over 2,000 times faster than the strongest baseline in 2D halls and over 100 times in 3D, with comparable or lower makespan and cost.
They plan about a thousand robots in seconds in 2D, and tens of seconds in 3D, and replan up to a thousand robots in real time.
\end{abstract}
\IEEEpeerreviewmaketitle

\section{Introduction}

\IEEEPARstart{L}{arge} teams of robots that share a workspace are central to applications ranging from automated warehouse fulfillment~\cite{wurman2008coordinating,d2012guest} and autonomous intersection management~\cite{chen2015cooperative} to coordinated quadrotor swarms~\cite{honig2018trajectory,preiss2017crazyswarm}, with commercial facilities deploying hundreds of robots in a single shared workspace~\cite{wurman2008coordinating}. 
In these settings, each robot must navigate from its start position to its goal without colliding with static obstacles or other robots.
Physical execution further requires that the motions obey the robot dynamics and actuation limits, which defines the \gls{acr:kmampp}.
A solution assigns one time-parametrized trajectory to each robot.
The trajectories must be \emph{conflict-free}, so that robot bodies remain disjoint from obstacles and one another at every instant, and \emph{dynamically feasible}, so that the robot dynamics and actuation limits are satisfied~\cite{donald1993kinodynamic}.

\input{figures_arxiv/fig1}

In practice, the planning runtime governs both the fleet size that can be deployed and how quickly the fleet responds to change.
First, in offline planning, i.e., before execution begins, the computational burden grows with the number of robots, so the available planning time caps the fleet size that can be coordinated.
Second, a sufficiently fast planner can update the trajectories while the robots keep moving (\cref{fig:showcase}), provided that each update returns within the lead time before it is executed.
Such \emph{real-time} planning~\cite{frazzoli2002real,hauser2012responsiveness} tightens the time budget further as the fleet grows.
Existing kinodynamic multi-agent planners remain several orders of magnitude slower than such time budgets for large fleets. 
Joint methods explore the combined state space of the fleet through sampling~\cite{le2017cooperative,li2016asymptotically} or optimize continuous trajectories under pairwise collision constraints~\cite{augugliaro2012generation}.
Both couple inter-robot coordination with dynamics-constrained motion synthesis, so the state dimension and the number of collision constraints grow with every robot, which restricts these planners to relatively small teams~\cite{moldagalieva2025dblacam}.

By contrast, solvers for the related problem of \gls{acr:mapf} scale to thousands of agents.
Standard \gls{acr:mapf} coordinates point agents on a graph through synchronized moves and waits, avoiding simultaneous vertex occupancy and opposing edge traversals~\cite{stern2019mapf}.
Within this abstraction, PIBT constructs joint moves through priority inheritance and backtracking~\cite{okumura2022priority}, and LaCAM embeds this efficient successor generation in a complete configuration search that has solved instances with thousands of agents in tens of seconds~\cite{okumura2023lacam,okumura2023improving}.

Its solutions, however, are discrete schedules that abstract away robot-body geometry and dynamics, and not the continuous, dynamically feasible trajectories required by the \gls{acr:kmampp}.
Several planners build on \gls{acr:mapf} to solve the \gls{acr:kmampp} and differ in where they enforce dynamic feasibility. 
One class incorporates the dynamics \emph{during} the search: db-CBS and db-LaCAM~\cite{moldagalieva2024dbcbs,moldagalieva2025dblacam} run \gls{acr:mapf} search over kinodynamic motion primitives, while K-CBS~\cite{kottinger2022conflict} coordinates single-robot kinodynamic planners within \gls{acr:cbs}. 
Searching over higher-dimensional kinodynamic states and primitives is expensive, however, which limits scalability to tens of robots~\cite{moldagalieva2025dblacam}. 
A second class enforces the dynamics \emph{after} the discrete search through continuous postprocessing.
MAPF/C+POST~\cite{honig2018trajectory} constructs a discrete schedule and then solves constrained trajectory optimization problems within safe corridors, and Park et al.~\cite{park2020efficient} additionally guarantee that these problems are feasible, reporting an average planning time of 6.4\,s for 64 robots.
Because postprocessing can reduce obstacle clearance and inter-robot separation, these pipelines must enforce continuous-time collision constraints during trajectory optimization, which takes several minutes for hundreds of robots~\cite{honig2018trajectory}. 
In both classes, executable motion thus requires additional computation, either to generate and evaluate kinodynamic motions during the search or to solve trajectory optimization problems after it.
As the fleet grows, this computation limits how much of the speed and scalability of \gls{acr:mapf} carries over to kinodynamic planning, particularly within the tighter time budgets of real-time planning. 

We resolve this bottleneck by accounting for the robot dynamics \emph{before} the discrete search.
Specifically, we analytically bound the clearance lost when a discrete plan is reconstructed into continuous trajectories under the prescribed speed and acceleration caps, and we reserve the corresponding margin in the discrete search.
A closed-form reconstruction, applied independently to each robot, then converts any valid discrete plan into smooth, dynamically feasible trajectories that are provably conflict-free in continuous time. 
This approach needs neither kinodynamic search nor trajectory optimization. 
The discrete search still ranges over synchronized moves and waits only, so the speed of \gls{acr:mapf} solvers carries over, and any discrete solver that returns valid plans becomes a certified kinodynamic planner.

\subsection{Statement of Contribution}

This paper presents \acrfull{acr:takt}, a framework that certifies dynamic feasibility and continuous-time collision avoidance through discrete coordination, together with two fast planners built on it.
\acrshort{acr:takt} brings kinodynamic multi-agent motion planning to real-time speeds for fleets of hundreds of robots and scales to more than a thousand robots, more than an order of magnitude beyond the largest fleets solved by existing baselines.
We model the robots as equal-radius closed balls in 2D and 3D with double-integrator dynamics and common speed and acceleration caps.
Our contributions are threefold.

First, we formulate \acrshort{acr:takt} as a pipeline of three stages, namely \emph{Discretization}, \emph{Discrete Coordination}, and \emph{Continuous Reconstruction} (\cref{sec:pwm}).
Discretization embeds a lattice in the workspace, reserves a certified clearance radius around each robot, and reduces the \gls{acr:kmampp} to a discrete search problem (\cref{prob:la-mapf}).
Discrete Coordination solves this problem with synchronized moves and waits, and Continuous Reconstruction filters the motion of each robot into a smooth trajectory.
We prove that every valid discrete plan reconstructs into conflict-free trajectories that satisfy the speed and acceleration caps, independently of the discrete solver (\cref{thm:soundness}).
We also derive sufficient clearance conditions under which a discrete plan exists, both at a fixed lattice resolution and under lattice refinement (\cref{thm:completeness,cor:resolution,cor:diamond-resolution}).

Second, we introduce a diamond formulation of the discrete problem and two planners that solve it.
On a fine lattice, a robot spans many vertices, and a Euclidean separation rule can prevent it from yielding sideways in the same tick in which another robot advances.
The diamond formulation restores this one-tick sideways yielding, and we derive a sharp condition on the diamond size that guarantees the Euclidean separation required by reconstruction (\cref{sec:diamond}).
We then adapt PIBT and LaCAM to this formulation through backtracking over simultaneous blockers, priority reshuffling in stationary groups, Euclidean candidate ordering, and memory-efficient goal-distance queries, which yields \acrshort{acr:takt}-PIBT and \acrshort{acr:takt}-LaCAM (\cref{sec:discrete}).
With 50 robots, \acrshort{acr:takt}-LaCAM is 2,200 to 2,600 times faster than MAPF/C+POST~\cite{honig2018trajectory} in four 2D halls and 105 to 415 times faster on three 3D maps, with comparable or lower makespan and cost (\cref{sec:exp-comparison}).
Both planners compute trajectories for about a thousand robots in seconds in 2D and in tens of seconds in 3D (\cref{sec:exp-scalability}), and ablation studies isolate the effect of each design element (\cref{sec:exp-ablation}).
In physics simulation, fleets of up to 1,082 differential-drive robots and 1,010 quadrotors track the planned trajectories without collisions (\cref{sec:exp-tracking}).

Third, we develop two paradigms for real-time replanning under online changes to goals and to the environment (\cref{sec:online}).
Whole-fleet replanning computes a complete new plan and, in our experiments, returns within 500\,ms for up to 763 robots in an empty hall and 335 quadrotors around a central pillar.
Incremental replanning extends the plan with \acrshort{acr:takt}-PIBT by one configuration per tick, faster than the robots execute it, for up to 1,018 robots and 671 quadrotors on the same maps (\cref{sec:exp-online}).

\emph{Organization:} \Cref{sec:related} reviews related work, and \cref{sec:problem} states the problem.
\Cref{sec:pwm} presents the \acrshort{acr:takt} framework, \cref{sec:diamond} the diamond formulation, and \cref{sec:discrete} the two planners.
\Cref{sec:theory} establishes the guarantees, \cref{sec:online} presents real-time planning, \cref{sec:experiments} reports the experiments, and \cref{sec:conclusion} concludes the paper.

\section{Related Work}\label{sec:related}

We review kinodynamic motion planning, \gls{acr:mapf}, and methods that connect them.
Here we use \emph{path} to denote a geometric route without timestamps, and \emph{trajectory} to denote a time-parameterized motion.

\subsection{Kinodynamic Motion Planning}\label{sec:related-kinodynamic}

Kinodynamic planning searches for trajectories that satisfy dynamic constraints. Sampling-based methods such as kinodynamic rapidly exploring random trees (RRT)~\cite{lavalle2001randomized} and stable sparse RRT (SST)~\cite{li2016asymptotically} generate candidate motions by integrating sampled controls through the robot dynamics. Optimization-based methods optimize a parameterized trajectory subject to dynamic constraints, as in polynomial and minimum-snap trajectory generation~\cite{mellinger2011minimum}. Conflict-free waypoints alone do not ensure continuous-time safety. Collision avoidance must be enforced along the trajectory or established by a post-hoc separate verification.

Multi-robot extensions either plan jointly or coordinate separate robot plans. Joint methods explore the combined state space through cooperative sampling~\cite{le2017cooperative} or optimize all trajectories under pairwise separation constraints, using methods such as sequential convex programming~\cite{augugliaro2012generation}. These approaches couple the robots directly, so the dimension of the search space or the number of separation constraints grows with every added robot, which restricts these planners to relatively small teams~\cite{moldagalieva2025dblacam}.
Prioritized planning coordinates separate plans by planning robots sequentially, with each avoiding the trajectories of higher-priority robots~\cite{erdmann1987multiple,cap2015prioritized}. 
Dynamic feasibility then depends on the single-robot planner. While avoiding direct coupling, the trajectories fixed for higher-priority robots can leave a later robot without a feasible trajectory~\cite{cap2015prioritized}. 

Real-time planners replan while the robot moves. Each replan starts from the state that the robot will reach after a lead time and must return within that lead time, resulting in higher runtime requirements~\cite{frazzoli2002real,hauser2012responsiveness}. 
Reported planning times of centralized kinodynamic planners range from seconds for tens of robots to minutes for hundreds~\cite{park2020efficient,honig2018trajectory}. 

Centralized kinodynamic planners account for dynamics and collision avoidance through the states, controls, or trajectory constraints of every robot. The resulting computational burden limits both the fleet size and the speed of replanning, which motivates solving the coordination problem on a simplified motion model.

\subsection{Multi-Agent Path Finding on Graphs}\label{sec:related-mapf}

\gls{acr:mapf} serves as an abstraction of such a coordination problem on graphs. 
In standard discrete-time \gls{acr:mapf}, agents move along an edge or wait at a vertex in each synchronized step. Plans avoid simultaneous occupancy of a vertex and opposite traversal of an edge~\cite{stern2019mapf}. This model abstracts away robot-body geometry and differential constraints.

\gls{acr:cbs} resolves conflicts through a high-level tree of constrained single-agent searches~\cite{sharon2015conflict}, while M* expands the joint search where agents interact~\cite{wagner2015subdimensional}. PIBT constructs one joint move at a time, in which a blocked agent passes its priority to the agent in its way~\cite{okumura2022priority}. LaCAM embeds such successor generation in a complete search over configurations and has solved instances with ten thousand agents~\cite{okumura2023lacam}. 
Closed-loop approaches repeatedly plan from the observed configuration and commit the next joint move~\cite{li2025fico,li2026adaptive,li2026certificate}.
Large-agent \gls{acr:mapf} models footprints spanning multiple vertices~\cite{li2019large}, while multi-dependency PIBT handles moves blocked by several agents~\cite{jiang2026mdpibt}.

\subsection{From Discrete Coordination to \gls{acr:kmampp}}\label{sec:related-execution}

Standard \gls{acr:mapf} discarded robot-body geometry, inter-step motion, and dynamic constraints, which are inevitable for solving \gls{acr:kmampp}, for better scalability. Two lines of work bridge the two fields. 

One line retains kinodynamic reasoning \emph{during} the search.
K-CBS coordinates single-robot kinodynamic planners within \gls{acr:cbs}~\cite{kottinger2022conflict}.
db-CBS and db-ECBS search precomputed primitives with bounded discontinuities and remove those discontinuities through joint trajectory optimization~\cite{moldagalieva2024dbcbs,moldagalieva2025dbecbs}.
db-LaCAM propagates controls from applicable primitives and coordinates the resulting motions through a kinodynamic extension of PIBT within LaCAM, with reported experiments reaching 50 robots~\cite{moldagalieva2025dblacam}.
Precomputed primitives encode local dynamics in advance, but the search still carries kinodynamic states and validates candidate motions.
Because dynamics enter the search directly through the state and transition representations, coordinating these higher-dimensional kinodynamic states limits scalability to tens of robots.

A second line constructs dynamically feasible trajectories \emph{after} discrete coordination.
MAPF/C+POST constructs safe corridors from a discrete schedule and optimizes one trajectory per robot within them~\cite{honig2018trajectory}.
Park et al.~\cite{park2020efficient} additionally guarantee feasibility of the resulting optimization problems.
These pipelines preserve obstacle clearance and inter-robot separation through continuous trajectory constraints, so executable motion still requires solving expensive post-hoc constrained optimization problems, limiting scalability.

\subsection{Summary and Positioning}\label{sec:related-summary}

The challenge addressed here is to retain simple discrete coordination while obtaining smooth, conflict-free motion under given dynamic limits through direct reconstruction.

\gls{acr:takt} framework establishes the reconstruction guarantee \emph{before} search and makes continuous feasibility a property of any discrete plan.
At a chosen lattice resolution, the prescribed speed and acceleration caps and reconstruction kernel fix the tick duration, filter window, and clearance certificate.
The search then coordinates only synchronized axial moves and waits.
Every valid plan reconstructs into smooth, collision-free trajectories satisfying the dynamic constraints through independent closed-form convolutions.
This guarantee is independent of the discrete solver, and reconstruction requires neither trajectory optimization nor collision repair.
Building on this framework, we develop two fast discrete planners whose efficiency enables real-time planning for fleets of up to a thousand robots.

\section{Problem Statement}\label{sec:problem}
\subsection{Workspace, Obstacles, and Robot Bodies}\label{sec:workspace}
In this section, we formalize the workspace, the robot model, and the planning problem.
Let $d\in\{2,3\}$ denote the workspace dimension. Throughout, $\norm{\cdot}_2$ and $\norm{\cdot}_1$ denote the Euclidean and Manhattan norms on $\mathbb R^d$, and $\operatorname{dist}_2(p,A)=\inf_{z\in A}\norm{p-z}_2$ is the Euclidean distance from a point $p$ to a nonempty set $A$. The workspace $\mathcal W\subset\mathbb R^d$ is a compact set with nonempty interior $\operatorname{int}\mathcal W$. The static obstacles form a closed set $\mathcal O\subseteq\mathcal W$.
The forbidden set~$\forbidden$ collects the obstacles and every point outside the workspace interior, and the clearance~$\clr(p)$ of a point is its distance to this set, that is,
\begin{equation}\label{eq:forbidden}
 \forbidden=\mathcal O\cup(\mathbb R^d\setminus\operatorname{int}\mathcal W),
 \qquad \clr(p)=\operatorname{dist}_2(p,\forbidden).
\end{equation}
The set $\forbidden$ is closed and nonempty, so the distance in \cref{eq:forbidden} is attained at a point of $\forbidden$, and the clearance function is $1$-Lipschitz,
\begin{equation}\label{eq:clearance-lipschitz}
 |\clr(p)-\clr(q)|\le\norm{p-q}_2\qquad\text{for all }p,q\in\mathbb R^d .
\end{equation}

The $N\ge1$ robots are closed balls of common radius $r>0$, and the position of a robot is that of its center.
A robot centered at~$p$ thus occupies~$\mathcal B_r(p)=\{z\in\mathbb R^d:\norm{z-p}_2\le r\}.$
A robot of another shape can be represented by its smallest enclosing ball. 
Because the bodies and the forbidden set are closed, intersection tests reduce to strict distance comparisons,
\begin{align}
 \mathcal B_r(p)\cap\forbidden=\varnothing
 &\iff \clr(p)>r,\label{eq:free-body}\\
 \mathcal B_r(p)\cap\mathcal B_r(q)=\varnothing
 &\iff \norm{p-q}_2>2r.\label{eq:disjoint-bodies}
\end{align}
By \cref{eq:free-body}, the free space, i.e., the set of centers for which the body avoids the forbidden set, is~$\freespace=\{p\in\mathbb R^d:\clr(p)>r\}.$

\subsection{Robot Dynamics}\label{sec:dynamics}

We model each robot as a fully actuated double integrator in $\mathbb R^d$ with bounded speed and acceleration. Let $x_i(t)\in\mathbb R^d$ be the center of robot $i$ at time $t$, let $w_i(t)\in\mathbb R^d$ be its velocity, and let $a_i(t)\in\mathbb R^d$ be its acceleration input. The dynamics are
\begin{equation}\label{eq:dynamics}
 \begin{aligned}
 \dot x_i(t)&=w_i(t),& \dot w_i(t)&=a_i(t),\\
 \norm{w_i(t)}_2&\le v,& \norm{a_i(t)}_2&\le a_{\max}.
 \end{aligned}
\end{equation}
The caps~$v>0$ and~$a_{\max}>0$ are given independently and are common to all robots. 
A trajectory that satisfies these dynamics and caps is called \emph{admissible}, as formalized next.

\begin{definition}[Admissible trajectory]\label{def:admissible}
A trajectory $x\colon[0,T]\to\mathbb R^d$ is admissible if $x\in C^1([0,T];\mathbb R^d)$, its derivative $\dot x$ is absolutely continuous, $\norm{\dot x(t)}_2\le v$ for every $t\in[0,T]$, and $\norm{\ddot x(t)}_2\le a_{\max}$ almost everywhere.
\end{definition}

The double integrator is a planning abstraction, and \cref{sec:exp-tracking} shows that differential-drive robots and quadrotors track the resulting trajectories with standard controllers.

\subsection{Configurations and Conflict-Free Configurations}\label{sec:configurations}

\begin{definition}[Fleet configuration]\label{def:configuration}
A configuration of the $N$ labeled robots is an ordered tuple~$ Q=(p_1,\ldots,p_N)\in(\mathbb R^d)^N,$
where $p_i$ is the center of robot $i$.
\end{definition}

\begin{definition}[Conflict-free configuration]\label{def:conflict-free}
A configuration $Q=(p_1,\ldots,p_N)$ is conflict-free if $\clr(p_i)>r$ and $\norm{p_i-p_j}_2>2r$ for every $i$ and every $j\ne i$.
The set of conflict-free configurations is denoted by $\freeconf\subseteq(\mathbb R^d)^N$.
\end{definition}

Conflict freedom is a condition on the physical bodies. \Cref{sec:pwm} introduces a stricter condition for lattice configurations, called \emph{validity}, in which a certificate radius~$\reps>r$ replaces~$r$.

\subsection{The Motion Planning Problem}\label{sec:mrmp}

An instance of the planning problem is a tuple~$\mathcal I=\tup{\mathcal W,\mathcal O,r,v,a_{\max},S,G,\eta}$,
in which $S=(s_1,\ldots,s_N)\in\freeconf$ is the start configuration, $G=(g_1,\ldots,g_N)\in\freeconf$ is the goal configuration, and~$\eta\ge0$ is a small endpoint tolerance. 
Robot $i$ starts within $\eta$ of $s_i$ and must come to rest within $\eta$ of $g_i$.

\begin{problem}[\acrlong{acr:kmampp}]\label{prob:mampp}
Given an instance $\mathcal I$, find a finite horizon $T>0$ and admissible trajectories (\cref{def:admissible}) $x_1,\ldots,x_N$ on $[0,T]$ such that, for every robot $i$,
\begin{equation}\label{eq:endpoint-conditions}
 \begin{gathered}
 \norm{x_i(0)-s_i}_2\le\eta,\qquad \norm{x_i(T)-g_i}_2\le\eta,\\
 \dot x_i(0)=\dot x_i(T)=0,
 \end{gathered}
\end{equation}
and, for every $t\in[0,T]$ and every $j\ne i$,
\begin{equation}\label{eq:physical-safety}
 \clr(x_i(t))>r,\qquad \norm{x_i(t)-x_j(t)}_2>2r.
\end{equation}
\end{problem}

We define the margin of a solution $X=(x_1,\ldots,x_N)$ as the smallest amount by which the inequalities in \cref{eq:physical-safety} hold over the whole horizon,
\begin{equation}\label{eq:motion-clearance}
 \begin{aligned}
 \mu_X=\min\Bigl\{
 &\min_{i}\ \min_{t\in[0,T]}\bigl(\clr(x_i(t))-r\bigr),\\
 &\min_{i<j}\ \min_{t\in[0,T]}\bigl(\norm{x_i(t)-x_j(t)}_2-2r\bigr)\Bigr\},
 \end{aligned}
\end{equation}
where the second inner minimum is omitted when $N=1$.

Every solution has margin $\mu_X>0$, since each minimum in \cref{eq:motion-clearance} is attained on the compact horizon $[0,T]$, at a time at which the strict inequalities of \cref{eq:physical-safety} hold.

Intuitively, the margin measures how much clearance a solution leaves beyond the physical minimum. 
It drives the existence results of \cref{sec:theory}, where a discrete plan is shown to exist whenever the margin exceeds the combined lattice approximation error and certificate inflation (\Cref{thm:completeness} and \Cref{cor:resolution}).

\Cref{prob:mampp} asks for a feasible motion, and we compare solutions of the same instance using two metrics.
The arrival time $\tau_i(X)$ is the earliest time after which robot $i$ remains at $x_i(T)$ for the rest of the horizon. The \emph{makespan} is defined as $\max_i\tau_i(X)$.
The \emph{cost} of a solution integrates a quadratic running cost over the horizon,
\begin{equation}\label{eq:cost-function}
 \begin{aligned}
 J(X)=\sum_{i=1}^N\int_0^T\Bigl(&q_p\norm{x_i(t)-x_i(T)}_2^2+q_v\norm{\dot x_i(t)}_2^2\\
 &+r_a\norm{\ddot x_i(t)}_2^2\Bigr)\,dt,
 \end{aligned}
\end{equation}
with weights $q_p,q_v,r_a\ge0$ that are not all zero. The three terms penalize the state deviation $(x_i-x_i(T),\dot x_i)$ from rest at the final position and the input $a_i=\ddot x_i$ of \cref{eq:dynamics}, as in linear-quadratic control~\cite{kalman1960contributions,kirk2004optimal}.

This paper targets feasible solutions of \cref{prob:mampp} at speed and at scale, and \cref{sec:experiments} reports the makespan and the cost of the solutions obtained.

\input{figures_arxiv/pipeline}

\section{\gls{acr:takt} Framework and Construction}\label{sec:pwm}

\subsection{Pipeline Overview}\label{sec:overview}

In this section, we present the \gls{acr:takt} framework, which coordinates the robots on a lattice and obtains dynamic feasibility afterwards from a fixed filter whose effect on the clearance is bounded before the search. Its pipeline comprises 
three stages (\cref{fig:pipeline}).
\emph{Discretization} places a lattice in the workspace and sets the timing and clearance requirements according to the speed and acceleration limits and the chosen smoothing filter.
\emph{Discrete Coordination} plans synchronized moves and waits on this lattice, maintaining the required obstacle clearance and inter-robot separation at every step.
\emph{Continuous Reconstruction} filters each robot's motion independently, using a common kernel, to produce smooth, conflict-free trajectories within the speed and acceleration caps.
The guarantee holds for every valid discrete plan, allowing PIBT~\cite{okumura2022priority}, LaCAM~\cite{okumura2023lacam}, or another discrete solver.

\subsection{Discretization}\label{sec:discretization}

We discretize the workspace using a lattice of pitch $\eps>0$. Fix an origin $o\in\mathbb R^d$ and an orthonormal frame $e_1,\ldots,e_d$. All coordinates and Manhattan norms below are expressed in this frame. 

\begin{definition}[Lattice vertices and elementary moves]\label{def:lattice}
For a pitch $\eps>0$, the lattice, the move set, and the tick duration are
\begin{equation}\label{eq:lattice}
 \latt=o+\eps\mathbb Z^d,\,
 \moveset=\{0,\pm\eps e_1,\ldots,\pm\eps e_d\},\,
 \dt=\eps/v.
\end{equation}
Two vertices are axial neighbors when their difference is one of the $2d$ nonzero displacements $\pm\eps e_a$. 
An elementary move is a displacement in $\moveset$, and the zero displacement is a wait. 
A tick is a time interval of duration $\dt$. 
A lattice configuration is a configuration in the sense of \cref{def:configuration} whose positions all belong to $\latt$.
\end{definition}

\gls{acr:takt} temporarily relaxes the acceleration cap: each robot either moves to an axial neighbor at speed $v$ or waits, with instantaneous velocity changes between ticks.
The pitch may be much smaller than the robot diameter, so that a body spans many lattice vertices.

Continuous Reconstruction (\cref{sec:smoothing}) restores dynamic feasibility by convolving each stop-and-go trajectory with a common kernel $\phi$.
The kernel is symmetric and nonnegative, has unit integral, and is supported on $[-\Tf/2,\Tf/2]$, where $\Tf>0$ is the filter duration. We use the box kernel throughout this paper. 
\Cref{lem:filter} bounds the acceleration of the reconstructed trajectory by $c_\phi v/\Tf$, with $c_\phi=2$ for the box kernel. Given the acceleration cap $a_{\max}$, the following definition selects the shortest filter duration that satisfies this bound and spans an even number of ticks.

\begin{definition}[Filter width and duration]\label{def:filter-duration}
The target duration, the filter width, and the filter duration are
\begin{equation}\label{eq:filter-duration}
 T_{\mathrm{acc}}=\frac{c_\phi v}{a_{\max}},\qquad
 W=2\Bigl\lceil\frac{T_{\mathrm{acc}}}{2\dt}\Bigr\rceil,\qquad
 \Tf=W\dt .
\end{equation}
\end{definition}

The width is the smallest even number of ticks whose duration reaches the target, so~$T_{\mathrm{acc}}\le\Tf<T_{\mathrm{acc}}+2\dt$.
A longer filter duration would also satisfy the acceleration cap, but it would enlarge the clearance reserve introduced next and delay the motion, so we select the minimum duration in \cref{def:filter-duration}.
Discrete coordination accounts for the clearance needed by reconstruction through the following certificate radius.
\begin{definition}[Certificate radius]\label{def:certificate-radius}
For a filter duration $\Tf$, the certificate radius is
\begin{equation}\label{eq:inflation}
 \reps^2=r^2+\eps^2/4+\kappa_\phi v^2\Tf^2 .
\end{equation}
The required clearance at a lattice vertex $p$ is $\clr(p)>\reps$, and the required separation of simultaneous positions $p,q$ is $\norm{p-q}_2>2\reps$.
\end{definition}

Here, $\kappa_\phi$ denotes the normalized variance of the kernel, defined by 
\begin{equation}\label{eq:moments}
 \sigk^2=\textstyle \int s^2\phi(s)\,ds=\kappa_\phi\Tf^2 .
\end{equation}
For the box kernel,~$\kappa_\phi=1/12$.
Intuitively, the \emph{tick reserve} $\eps^2/4$ covers the motion between two ticks, and the \emph{smoothing reserve} $\kappa_\phi v^2\Tf^2$ covers the deviation introduced by filtering, so that validity at the ticks implies conflict freedom of the reconstructed trajectories (\cref{thm:soundness}).
In the 2D experiments of \cref{sec:experiments}, $r=0.5$\,m, $v=1$\,m/s, $\eps=0.101$\,m, and $\Tf=0.606$\,s give $\reps=0.532$\,m, so the two reserves add 3.2\,cm to the robot radius. In the 3D experiments, $r=0.2$\,m gives $\reps=0.237$\,m.

The reference graph retains the vertices with the required obstacle clearance and connects them by elementary moves.  
\begin{definition}[Reference vertices and reference graph]\label{def:reference-graph}
The reference vertex set and the reference graph are
\begin{equation}\label{eq:admissible-sites}
 V_\eps=\{p\in\latt:\clr(p)>\reps\}, \quad \mathcal G_\eps=(V_\eps,E_\eps),\end{equation}
\begin{equation}\label{eq:reference-graph}
 E_\eps=\bigl\{\{p,q\}:p,q\in V_\eps,\ q-p\in\moveset\bigr\}.
\end{equation}
\end{definition}

The Discrete Coordination stage then operates on $\mathcal G_\eps$, scheduling synchronized elementary moves with the pairwise separation required by the certificate radius $\reps$.

\subsection{Discrete Coordination}\label{sec:requirements}

The Discrete Coordination stage returns a synchronized sequence of lattice configurations~$ Q^{0:K}=(Q^0,\ldots,Q^K)$, where~$Q^k=(p_1^k,\ldots,p_N^k)$ holds the positions of the robots at tick~$k$. 
Between consecutive ticks, each robot executes an elementary move over $[k\dt,(k+1)\dt]$, either traveling at constant velocity $(p_i^{k+1}-p_i^k)/\dt\in\{0,\pm ve_a\}$ to an axial neighbor or waiting at rest, that is,
\begin{equation}\label{eq:transition}
 \{p_i^k,p_i^{k+1}\}\in E_\eps\qquad(0\le k<K,\ 1\le i\le N),
\end{equation}
and every tick configuration satisfies the clearance and separation conditions in \cref{def:certificate-radius}:
\begin{equation}\label{eq:tick-clearance}
 p_i^k\in V_\eps,\ \norm{p_i^k-p_j^k}_2>2\reps
 \qquad(i\ne j,\ 0\le k\le K).
\end{equation}
These conditions are imposed at the ticks only, and \Cref{thm:soundness} extends them to continuous time. 
The configurations satisfying both conditions form the reference configuration set
\begin{equation}\label{eq:reference-configurations}
\begin{gathered}
 \mathcal Q_\eps=\bigl\{(p_1,\ldots,p_N)\in V_\eps^N:\ \\
 \norm{p_i-p_j}_2>2\reps\ \forall i<j\bigr\}.
\end{gathered}
\end{equation}

The goal of Discrete Coordination is a sequence of such configurations from the start to the goal. 
This problem admits a large-agent \gls{acr:mapf} (LA-MAPF) formulation~\cite{li2019large} with conflicts defined at ticks. 
At a reference vertex~$p$, a robot occupies the closed ball~$\mathcal B_{\reps}(p)$, and two robots have a conflict when these balls intersect at the same tick, i.e., when their centers are at distance at most~$2\reps$.

\begin{definition}[Valid configurations and plans]\label{def:validity}
A lattice configuration is valid if it belongs to $\mathcal Q_\eps$. A plan $Q^{0:K}$ is valid if every configuration is valid and every transition satisfies \cref{eq:transition}.
\end{definition}

The Discrete Coordination stage starts and ends at the anchor configurations $S^\eps=(s_1^\eps,\ldots,s_N^\eps)$ and $G^\eps=(g_1^\eps,\ldots,g_N^\eps)$ in $\mathcal Q_\eps$, which satisfy $\norm{s_i^\eps-s_i}_2\le\eta$ and $\norm{g_i^\eps-g_i}_2\le\eta$ for every $i$. 
The resulting LA-MAPF instance is~$\mathcal I_\eps=(\mathcal G_\eps,\reps,S^\eps,G^\eps)$.

\begin{problem}[Discrete Coordination Problem]\label{prob:la-mapf}
Given $\mathcal I_\eps$, find an integer $K\ge0$ and a valid plan $Q^{0:K}$ with $Q^0=S^\eps$ and $Q^K=G^\eps$.
\end{problem}

We refer to \cref{prob:la-mapf} as the \emph{reference problem}, since any solver that returns valid plans can be used within \gls{acr:takt}.
In this paper, we solve it with the diamond formulation of \cref{sec:diamond} and the two planners of \cref{sec:discrete}.

\subsection{Continuous Reconstruction}\label{sec:smoothing}

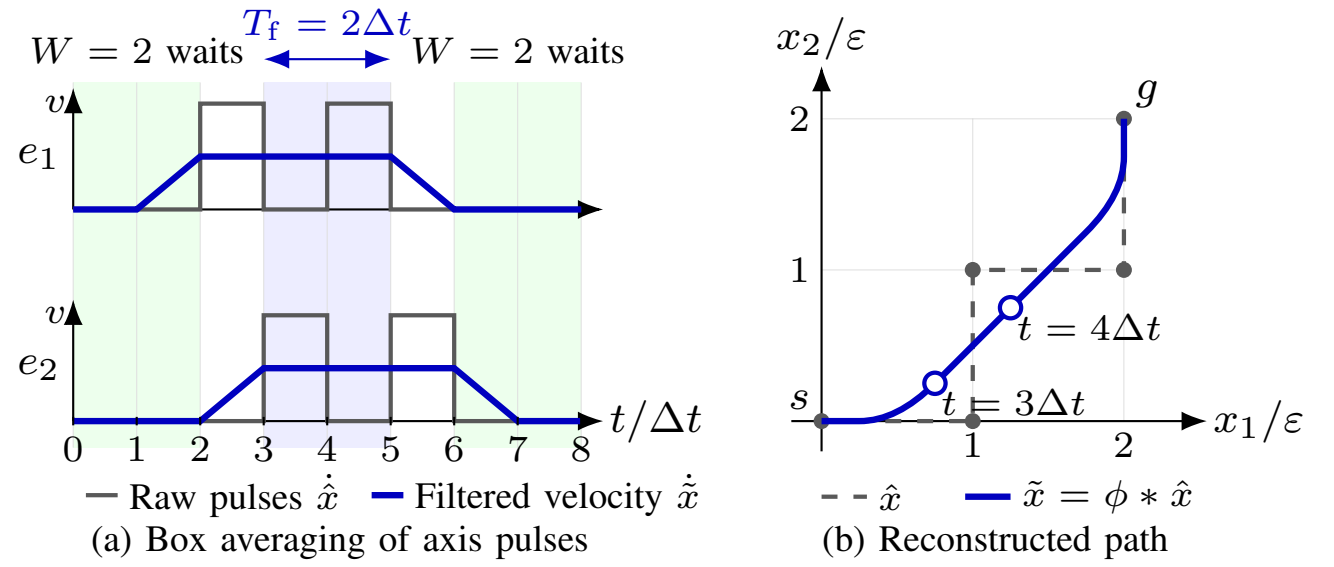
\begin{figure}[t]
\centering
\scalebox{\figscale}{
\begin{tikzpicture}[>=Latex,font=\footnotesize]
\begin{scope}[x=0.42cm,y=0.7cm]
 \fill[green!7] (0,-0.25) rectangle (2,3.2);
 \fill[green!7] (6,-0.25) rectangle (8,3.2);
 \fill[blue!7] (3,-0.25) rectangle (5,3.2);
 \foreach \t in {0,...,8} {
   \draw[black!12,very thin] (\t,-0.25)--(\t,3.2);
 }
 \node[align=center,font=\scriptsize] at (1,3.5) {$W=2$ waits};
 \node[align=center,font=\scriptsize] at (7,3.5) {$W=2$ waits};
 \draw[<->,blue!75!black] (3,3.42)--node[above,font=\scriptsize] {$\Tf=2\dt$}(5,3.42);
 \draw[->] (0,2)--(8.35,2);
 \draw[->] (0,2)--(0,3.15);
 \draw[->] (0,0)--(8.35,0) node[right,inner sep=1pt] {$t/\dt$};
 \draw[->] (0,0)--(0,1.15);
 \node[anchor=east,inner sep=1pt] at (-0.1,2.5) {$e_1$};
 \node[anchor=east,inner sep=1pt] at (-0.1,0.5) {$e_2$};
 \node[left,inner sep=1pt,font=\scriptsize] at (0,3) {$v$};
 \node[left,inner sep=1pt,font=\scriptsize] at (0,1) {$v$};
 \draw[black!65,thick]
   (0,2)--(2,2)--(2,3)--(3,3)--(3,2)--(4,2)--(4,3)--(5,3)--(5,2)--(8,2);
 \draw[black!65,thick]
   (0,0)--(3,0)--(3,1)--(4,1)--(4,0)--(5,0)--(5,1)--(6,1)--(6,0)--(8,0);
 \draw[blue!75!black,very thick]
   (0,2)--(1,2)--(2,2.5)--(5,2.5)--(6,2)--(8,2);
 \draw[blue!75!black,very thick]
   (0,0)--(2,0)--(3,0.5)--(6,0.5)--(7,0)--(8,0);
 \foreach \t in {0,...,8} {
   \draw (\t,0.04)--(\t,-0.04);
   \node[below,inner sep=2pt,font=\scriptsize] at (\t,-0.04) {$\t$};
 }
 \draw[black!65,thick] (0.2,-0.7)--(0.7,-0.7);
 \node[right,inner sep=2pt,font=\scriptsize] at (0.7,-0.7) {Raw pulses $\dot{\hx}$};
 \draw[blue!75!black,very thick] (4.7,-0.7)--(5.2,-0.7);
 \node[right,inner sep=2pt,font=\scriptsize] at (5.2,-0.7) {Filtered velocity $\dot{\tx}$};
 \node[font=\scriptsize] at (4.2,-1.15) {(a) Box averaging of axis pulses};
\end{scope}
\begin{scope}[shift={(4.95,0)},x=1.0cm,y=1.0cm]
 \draw[step=1,black!12,very thin] (0,0) grid (2,2);
 \draw[->] (-0.2,0)--(2.55,0) node[right,inner sep=1pt] {$x_1/\eps$};
 \draw[->] (0,-0.2)--(0,2.35) node[above,inner sep=1pt] {$x_2/\eps$};
 \draw[black!65,thick,dashed] (0,0)--(1,0)--(1,1)--(2,1)--(2,2);
 \foreach \p in {(0,0),(1,0),(1,1),(2,1),(2,2)} {
   \fill[black!65] \p circle (1.5pt);
 }
 \draw[blue!75!black,very thick]
   (0,0)--(0.25,0)
   .. controls ({5/12},0) and ({7/12},{1/12}) .. (0.75,0.25)
   --(1.75,1.25)
   .. controls ({23/12},{17/12}) and (2,{19/12}) .. (2,1.75)
   --(2,2);
 \filldraw[fill=white,draw=blue!75!black,thick] (0.75,0.25) circle (2pt);
 \filldraw[fill=white,draw=blue!75!black,thick] (1.25,0.75) circle (2pt);
 \node[below right,inner sep=1pt,font=\scriptsize] at (0.75,0.25) {$t=3\dt$};
 \node[below right,inner sep=1pt,font=\scriptsize] at (1.25,0.75) {$t=4\dt$};
 \node[above left,inner sep=2pt] at (0,0) {$s$};
 \node[above right,inner sep=2pt] at (2,2) {$g$};
 \foreach \a in {1,2} {
   \node[below,inner sep=2pt,font=\scriptsize] at (\a,0) {$\a$};
   \node[left,inner sep=2pt,font=\scriptsize] at (0,\a) {$\a$};
 }
 \draw[black!65,thick,dashed] (0,-0.5)--(0.3,-0.5);
 \node[right,inner sep=2pt,font=\scriptsize] at (0.3,-0.5) {$\hx$};
 \draw[blue!75!black,very thick] (0.95,-0.5)--(1.25,-0.5);
 \node[right,inner sep=2pt,font=\scriptsize] at (1.25,-0.5) {$\tx=\phi*\hx$};
 \node[font=\scriptsize] at (1.15,-0.805) {(b) Reconstructed path};
\end{scope}
\end{tikzpicture}}
\caption{Exact box-filter reconstruction of four axis moves with filter width $W=2$.
(a) The raw velocity pulses and their box averages, with the inserted waits (green) and the filter window at $t=4\dt$ (blue).
(b) The stop-and-go trajectory and its box average, which is piecewise quadratic.}
\label{fig:dac}
\end{figure}

A solution of \cref{prob:la-mapf} defines a \emph{stop-and-go trajectory} for each robot, which passes through the tick positions~$p_i^0,\ldots,p_i^K$ at the times $k\dt$.
Its velocity is a square wave, with a pulse~$\pm ve_a$ of duration $\dt$ for each move and a zero-velocity interval for each wait.
Continuous Reconstruction applies the kernel~$\phi$ of \cref{sec:discretization} as a low-pass filter that averages these pulses over a window of duration $\Tf$.
The filtered velocity is determined by the time spent moving in each signed direction within the window, in analogy with pulse-width modulation, and it lies in the convex hull of $\{0,\pm ve_a:a=1,\ldots,d\}$.
\Cref{fig:dac} shows the pulses, the filtered velocities, and the reconstructed paths in a planar example.

\begin{definition}[Padded reconstruction]\label{def:smoothing}
Let $Q^{0:K}$ be a valid plan in the sense of \cref{def:validity}. Prepend and append $W$ waits to the position sequence of each robot, which gives the duration~$T_p=(K+2W)\dt=K\dt+2\Tf$.
Let $\hx_i$ be the stop-and-go trajectory of the padded sequence of robot $i$, extended constantly outside $[0,T_p]$. The reconstructed trajectory is
\begin{equation}\label{eq:dac}
 \tx_i(t)=(\phi*\hx_i)(t)
       =\textstyle \int_{-\Tf/2}^{\Tf/2}\phi(s)\hx_i(t-s)\,ds,
\end{equation}
where $t\in[0,T_p]$. Every robot uses the same kernel, width, padding, and clock.
\end{definition}

The padded waits ensure that the reconstructed trajectories start and end at rest at the endpoints of the plan.
For $a>0$ write $B_a(s)=a^{-1}\mathbf1_{[-a/2,a/2]}(s)$. The box kernel is $\phi=B_{\Tf}$. With the primitive $F_i'=\hx_i$,
\begin{equation}\label{eq:closed-form-dac}
 \tx_i(t)=\bigl[F_i(t+\Tf/2)-F_i(t-\Tf/2)\bigr]/\Tf,
\end{equation}
so each reconstructed trajectory consists of $O(K+W)$ quadratic pieces. The coefficients are obtained directly from prefix integrals of the stop-and-go trajectory, so reconstruction can be computed independently for each robot in parallel. 

\begin{lemma}[Regularity and derivative bounds of the reconstruction]\label{lem:filter}
Let $\hx$ be the stop-and-go trajectory of a robot and~$\tx=\phi*\hx$ its box reconstruction. Then $\tx$ is $C^1$, $\dot{\tx}$ is absolutely continuous, and
\begin{equation}\label{eq:filter-bounds}
 \begin{gathered}
 \norm{\dot{\tx}(t)}_2\le v\quad\text{for all }t,\\
 \norm{\ddot{\tx}(t)}_2\le\frac{c_\phi v}{\Tf}=\frac{2v}{\Tf}\quad\text{almost everywhere}.
 \end{gathered}
\end{equation}
When $\Tf\ge T_{\mathrm{acc}}$, $\tx$ is admissible in the sense of \cref{def:admissible}.
\end{lemma}

\begin{proof}[Proof sketch]
The reconstructed velocity $\phi*\dot{\hx}$ is a convex average of raw velocities of norm at most $v$, so it stays in the ball of radius $v$ and is continuous. For the box kernel it equals the difference quotient $(\hx(t+\Tf/2)-\hx(t-\Tf/2))/\Tf$, which is $2v/\Tf$-Lipschitz because $\hx$ is $v$-Lipschitz. 
\end{proof}
\Cref{def:filter-duration} guarantees $\Tf\ge T_{\mathrm{acc}}$, so every reconstruction of \cref{def:smoothing} is admissible.

\begin{algorithm}[t]
\caption{Pipeline of the \gls{acr:takt} framework}\label{alg:pipeline}
\small
\begin{algorithmic}[1]
\Require Instance $\mathcal I$ of \cref{prob:mampp}; lattice pitch $\eps>0$; discrete solver $\mathcal S$
\Ensure Trajectories $\tx_1,\ldots,\tx_N$
\State $(\phi, c_\phi,\kappa_\phi)\gets\Call{KernelSelection}{}$ \Comment{box kernel, \cref{eq:moments}}
\LComment{\cref{def:filter-duration,def:certificate-radius}}
\State $(T_{\mathrm{acc}}, \dt, W, \Tf, \reps)\gets \Call{ComputeParameter}{\calI, \eps, c_\phi,\kappa_\phi}$ 
\State $\mathcal I_\eps\gets\Call{Discretization}{\calI,\reps}$
\State $Q^{0:K}\gets\Call{DiscreteCoordination}{\mathcal I_\eps}$
\ForAll{robots $i$}
 \State $\hx_i\gets\Call{Padding}{p_i^{0:K},W,\dt}$
 \State $\tx_i\gets\phi*\hx_i$ \Comment{\cref{def:smoothing}}
\EndFor
\State \Return $(\tx_1,\ldots,\tx_N)$
\end{algorithmic}
\end{algorithm}

\cref{alg:pipeline} summarizes the entire pipeline.

\section{Diamond Formulation for Discrete Coordination}\label{sec:diamond}

In this section, we introduce a diamond formulation of \cref{prob:la-mapf} that supports one-tick sideways yielding.
Lattice-aligned diamond footprints define pairwise conflicts through the Manhattan norm and enforce the Euclidean separation $2\reps$ required by \cref{def:certificate-radius}.

\subsection{Motivation for Diamond Footprints}\label{sec:motivation}

PIBT~\cite{okumura2022priority} is widely used in \gls{acr:mapf} for its speed. It coordinates one joint move at a time through priority inheritance and backtracking. 
A blocked agent asks the agent in its way to yield and passes on its priority, and this request propagates through the resulting dependencies.
Yielding is straightforward in standard \gls{acr:mapf}, where each agent occupies a single vertex.
When robots span several lattice vertices, however, the conflict geometry determines which responses are available. 
\Cref{fig:yielding} illustrates this effect on a schematic example in which two robots stand side by side, robot~$A$ moves one cell to the right, and robot~$B$ yields by moving one cell up.
The center offset loses one vertex along $A$'s direction of motion and gains one vertex sideways.

\begin{figure}[t]
\centering
\tikzset{yielding/.style={>=Latex,font=\footnotesize,
 body/.style={draw=orange!80!black,thick,fill=orange!15},
 cert/.style={draw=red!70!black,thick,densely dashed},
 dia/.style={draw=blue!70!black,thick,fill=blue!8},
 off/.style={draw=black!55,thin},
 mv/.style={-{Latex[length=3.2pt,width=3pt]},line width=0.9pt,draw=black!80},
 ctr/.style={circle,fill=black,inner sep=0pt,minimum size=2.6pt},
 ctrold/.style={circle,draw=black!60,fill=white,inner sep=0pt,minimum size=2.6pt}}}
 \scalebox{\figscale}{
\begin{tikzpicture}[yielding,x=0.2500cm,y=0.2500cm]
 \begin{scope}
  \draw[step=1,black!12,very thin] (-3.44,-2.6) grid (8.44,4.6);
  \draw[body] (0,0) circle (2);
  \draw[body] (5,1) circle (2);
  \draw[cert] (0,0) circle (2.363);
  \draw[cert] (5,1) circle (2.363);
  \draw[off] (0,0)--(5,1);
  \draw[mv] (0,0)--(1,0);
  \draw[mv] (5,1)--(5,2);
  \node[ctr] at (0,0) {};
  \node[ctr] at (5,1) {};
  \node[below,inner sep=2.5pt] at (0.5,0) {$A$};
  \node[right,inner sep=2.5pt] at (5,1.5) {$B$};
 \end{scope}
 \draw[->,thick] (8.89,1)--(9.99,1);
 \begin{scope}[shift={(13.70,0)}]
  \draw[step=1,black!12,very thin] (-3.44,-2.6) grid (8.44,4.6);
  \draw[body] (1,0) circle (2);
  \draw[body] (5,2) circle (2);
  \begin{scope}
   \clip (1,0) circle (2.363);
   \fill[red!70!black,opacity=0.5] (5,2) circle (2.363);
  \end{scope}
  \draw[cert] (1,0) circle (2.363);
  \draw[cert] (5,2) circle (2.363);
  \draw[off] (1,0)--(5,2);
  \draw[mv] (0,0)--(0.8,0);
  \draw[mv] (5,1)--(5,1.8);
  \node[ctrold] at (0,0) {};
  \node[ctrold] at (5,1) {};
  \node[ctr] at (1,0) {};
  \node[ctr] at (5,2) {};
  \node[below,inner sep=2.5pt] at (0.5,0) {$A$};
  \node[right,inner sep=2.5pt] at (5,1.5) {$B$};
 \end{scope}
 \node[font=\scriptsize] at (9.35,-3.55) {(a) Euclidean rule: $\sqrt{26}\,\eps>2\reps>\sqrt{20}\,\eps$};
\end{tikzpicture}}\\[0.3ex]
\scalebox{\figscale}{
\begin{tikzpicture}[yielding,x=0.2357cm,y=0.2357cm]
 \begin{scope}
  \draw[step=1,black!12,very thin] (-3.3,-3.2) grid (9.3,5.2);
  \draw[dia] (3,0)--(0,3)--(-3,0)--(0,-3)--cycle;
  \draw[dia] (9,1)--(6,4)--(3,1)--(6,-2)--cycle;
  \draw[body] (0,0) circle (2.1213);
  \draw[body] (6,1) circle (2.1213);
  \draw[off] (0,0)-|(6,1);
  \draw[mv] (0,0)--(1,0);
  \draw[mv] (6,1)--(6,2);
  \node[ctr] at (0,0) {};
  \node[ctr] at (6,1) {};
  \node[below,inner sep=2.5pt] at (0.5,0) {$A$};
  \node[right,inner sep=2.5pt] at (6,1.5) {$B$};
 \end{scope}
 \draw[->,thick] (9.78,1)--(10.94,1);
 \begin{scope}[shift={(14.53,0)}]
  \draw[step=1,black!12,very thin] (-3.3,-3.2) grid (9.3,5.2);
  \draw[dia] (4,0)--(1,3)--(-2,0)--(1,-3)--cycle;
  \draw[dia] (9,2)--(6,5)--(3,2)--(6,-1)--cycle;
  \draw[body] (1,0) circle (2.1213);
  \draw[body] (6,2) circle (2.1213);
  \draw[off] (1,0)-|(6,2);
  \draw[mv] (0,0)--(0.8,0);
  \draw[mv] (6,1)--(6,1.8);
  \node[ctrold] at (0,0) {};
  \node[ctrold] at (6,1) {};
  \node[ctr] at (1,0) {};
  \node[ctr] at (6,2) {};
  \node[below,inner sep=2.5pt] at (0.5,0) {$A$};
  \node[right,inner sep=2.5pt] at (6,1.5) {$B$};
 \end{scope}
 \node[font=\scriptsize] at (10.26,-4.20) {(b) Diamond rule: $\norm{(6,1)}_1=\norm{(5,2)}_1=7>6$};
\end{tikzpicture}}
\caption{One-tick sideways yield under the Euclidean and diamond rules. Robot $A$ moves one cell to the right while robot $B$ moves one cell upward. Panels (a) and (b) each show the configuration before and after the tick. }
\label{fig:yielding}
\end{figure}

Under the Euclidean rule in \cref{fig:yielding}(a), the robots are discs of radius $r=2\eps$, and \cref{eq:inflation} gives $\reps\approx2.36\,\eps$ for the box kernel with $W=4$. In lattice units, the simultaneous moves change the center offset as~$ (5,1)\longrightarrow(4,2),\ \sqrt{26}\,\eps>2\reps>\sqrt{20}\,\eps$,
so the pair satisfies the separation of \cref{eq:tick-clearance} before the move and violates it after the move. Robot $B$'s sideways step does not compensate for the reduced axial separation. Robot $B$ can retreat to the right to preserve the separation, but it remains in front of $A$ and may be blocked by another robot. 
A two-tick alternative is for $B$ to move upward while $A$ waits, then for both robots to advance to the right.
However, coordinating such multi-step maneuvers increases the search effort, since each robot has $(2d+1)^L$ possible action sequences over $L$ ticks, and one sequence can cause conflicts with several robots~\cite{yukhnevich2025epibt,jiang2026mdpibt}.

Under the diamond rule (\cref{fig:yielding}(b)), a single perpendicular step restores separation and supports the response in one tick, avoiding multi-tick maneuvers.
The lattice pitch is $\eps=\sqrt2\,r/3$, so diamonds of axial radius three cells enclose the same discs, and two diamonds are disjoint exactly when the Manhattan distance of their centers exceeds six cells. The same simultaneous moves give~$(6,1)\longrightarrow(5,2),\
 \norm{(6,1)}_1=\norm{(5,2)}_1=7>6$,
so the Manhattan distance is unchanged and the diamonds remain disjoint.
The certified construction below selects the diamond size to guarantee the Euclidean separation required by reconstruction.
The diamond rule admits fewer configurations than the Euclidean rule, and a robot at the minimum permitted separation can still yield sideways within one tick. \Cref{sec:exp-ablation} measures the effect.

\subsection{Aligned Footprints and One-Tick Yielding}\label{sec:footprints}

\begin{definition}[Enclosing and planning diamonds]\label{def:diamond-policy}
Fix an integer $n\ge1$ and tie the pitch to the robot radius,
\begin{equation}\label{eq:diamond-alignment}
 n\eps=\sqrt d\,r,\qquad \eps(n)=\frac{\sqrt d\,r}{n},\qquad \dt(n)=\frac{\eps(n)}{v}.
\end{equation}
For an integer $k\ge1$, the diamond of axial radius $k\eps$ around a vertex $p$ is the cross-polytope~$\diamondbody_k(p)=\{z\in\mathbb R^d:\norm{z-p}_1\le k\eps\}$,
a diamond for $d=2$ and an octahedron for $d=3$. 
For each robot we distinguish the following two diamonds: 
The \emph{enclosing diamond} of a robot at $p$ is $\diamondbody_n(p)$, the smallest lattice-aligned diamond containing its body. Discrete Coordination uses a \emph{planning diamond} $\diamondbody_m(p)$, with integer $m\ge n$.
\end{definition}

\begin{figure}[t]
\centering
\scalebox{\figscale}{
\begin{tikzpicture}[>=Latex,font=\footnotesize]
\begin{scope}[shift={(-2.1,0)}]
 \fill[red!30] (0,0) circle (1.2151);
 \filldraw[fill=blue!6,draw=blue!70!black,thick]
   (1.591,0)--(0,1.591)--(-1.591,0)--(0,-1.591)--cycle;
 \draw[blue!60!black,dashed]
   (1.0607,0)--(0,1.0607)--(-1.0607,0)--(0,-1.0607)--cycle;
 \draw[red!70!black,thick,densely dashed] (0,0) circle (1.2151);
 \filldraw[fill=orange!15,draw=orange!80!black,thick]
   (0,0) circle (0.75);
 \fill (0,0) circle (1.2pt);
 \draw[->] (0,0)--(1.591,0);
 \node[right,inner sep=1.5pt] at (1.591,0) {$h$};
 \draw[->] (0,0)--(-1.0607,0);
 \node[above,inner sep=1.5pt] at (-0.45,0) {$n\eps$};
 \draw[->,orange!80!black] (0,0)--(0.375,-0.6495);
 \node[orange!80!black,right,inner sep=1pt] at (0.25,-0.33) {$r$};
 \draw[blue!60!black,thin] (-0.4,-0.6607)--(-1.0,-1.22) node[below left,inner sep=1pt] {$\diamondbody_n$};
 \draw[orange!80!black,thin] (-0.352,0.662)--(-0.95,1.22) node[above left,inner sep=1pt] {$\mathcal B_r$};
 \node[blue!70!black] at (1.15,1.12) {$\diamondbody_{m_*}$};
 \draw[red!70!black,thin] (0.8592,-0.8592)--(1.25,-1.25) node[below right,inner sep=1pt] {$\mathcal B_{\reps}$};
 \node at (0,1.95) {(a) $d=2$: $n=2$, $W=6$, $m_*=3$};
 \node at (0,-1.95) {$n\eps=\sqrt2\,r$,\quad $h=m_*\eps$};
\end{scope}
\begin{scope}[shift={(2.2,0)}]
 \coordinate (xp) at (1.23,-0.71014);
 \coordinate (xm) at (-1.23,0.71014);
 \coordinate (yp) at (-1.23,-0.71014);
 \coordinate (ym) at (1.23,0.71014);
 \coordinate (zp) at (0,1.42028);
 \coordinate (zm) at (0,-1.42028);
 \fill[blue!6] (zp)--(xm)--(yp)--(zm)--(xp)--(ym)--cycle;
 \filldraw[fill=orange!15,draw=orange!80!black,thick]
   (0,0) circle (1.00429);
 \draw[orange!65!black,thin,dashed]
   (1.00429,0) arc[start angle=0,end angle=180,x radius=1.00429,y radius=0.57983];
 \draw[orange!65!black,thin]
   (-1.00429,0) arc[start angle=180,end angle=360,x radius=1.00429,y radius=0.57983];
 \draw[blue!70!black,thick]
   (zp)--(xm)--(yp)--(zm)--(xp)--(ym)--cycle
   (zp)--(xp)--(yp)--cycle;
 \fill (0,0) circle (1.2pt);
 \draw[->] (0,0)--node[left,fill=orange!15,inner sep=1pt] {$r$}(-0.87,0.50215);
 \draw[->] (0,0)--node[right,fill=orange!15,inner sep=1pt] {$h$}(zp);
 \draw[orange!80!black,thin] (-0.769,-0.6455)--(-1.3,-1.22) node[below left,inner sep=1pt] {$\mathcal B_r$};
 \node[blue!70!black,anchor=west] at (0.55,1.3) {$\diamondbody_n=\diamondbody_{m_*}$};
 \node at (0,1.95) {(b) $d=3$: $m_*=n$};
 \node at (0,-1.95) {$h=n\eps=\sqrt3\,r$};
\end{scope}
\end{tikzpicture}}
\caption{The physical ball $\mathcal B_r$ (orange), the enclosing diamond $\diamondbody_n$ (dashed), and the planning diamond $\diamondbody_{m_*}$ (solid). (a) An inflated planar footprint. The certificate ball of radius $\reps$ (red, dashed) extends past the footprint near the midpoints of its edges. (b) The case $m_*=n$ in three dimensions, in which the two footprints coincide and $\mathcal B_r$ is inscribed in the octahedron.
}
\label{fig:diamond}
\end{figure}
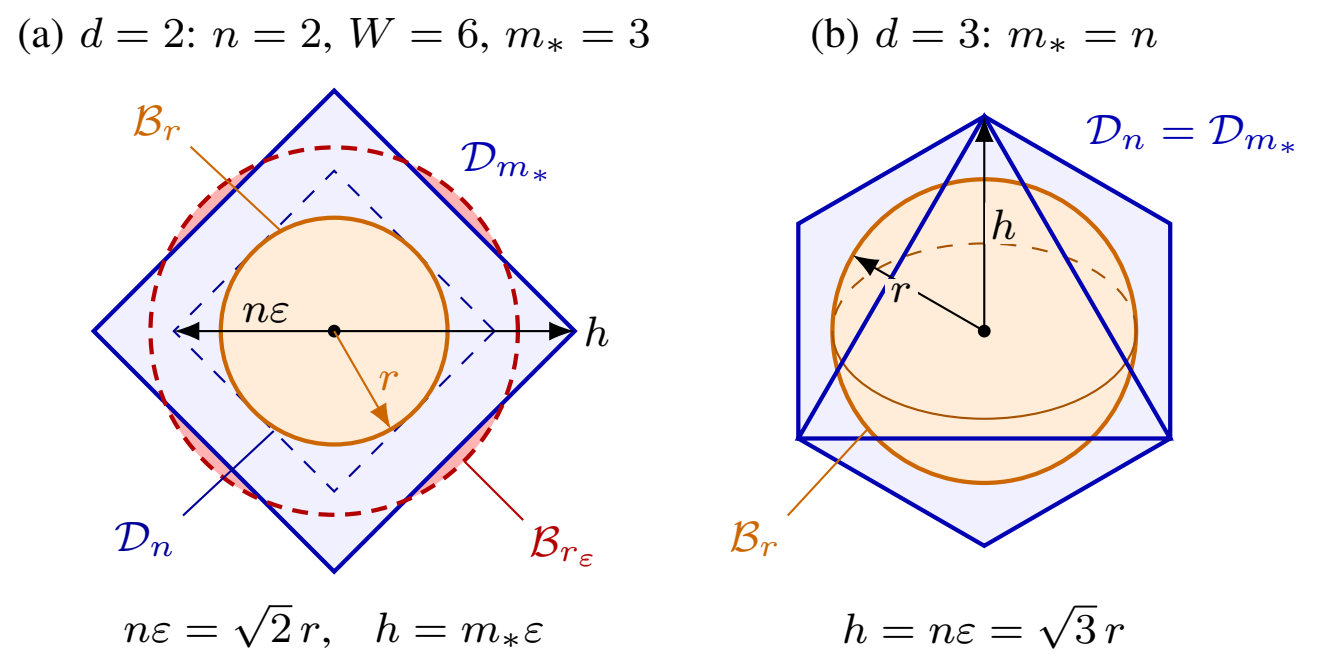

The planning diamond contains the enclosing diamond. For a given problem with fixed robot sizes, the size of the enclosing diamond is fixed, and the \emph{resolution integer}~$n$ sets the pitch~$\eps$.
Given $n$, the footprint integer $m\ge n$ sets the planning diamond's axial radius $m\eps$ and determines the pairwise conflict rule.
The geometric relationship is shown in \cref{fig:diamond}. The certificate ball of radius $\reps$ need not lie inside the planning diamond. 
The pairwise diamond condition at a tick requires the planning diamonds to be disjoint.
The triangle inequality and the midpoint of the centers give
\begin{equation}\label{eq:diamond-policy}
 \diamondbody_m(p)\cap\diamondbody_m(q)=\varnothing
 \iff\norm{p-q}_1>2m\eps.
\end{equation}
Let $a=(p-q)/\eps\in\mathbb Z^d$ denotes the relative lattice offset. We obtain 
\begin{equation}\label{eq:diamond-gap-integer}
 \norm{p-q}_1>2m\eps\iff\norm{a}_1\ge2m+1
\end{equation}

\begin{lemma}[One-tick sideways yield]\label{lem:sidestep}
Let two robots sit at $p$ and $q$ with $q-p=(2m+1)\eps e_a$. If during one tick the first robot moves to $p'=p+\eps e_a$ and the second to $q'=q\pm\eps e_b$ with $b\ne a$, then their planning diamonds remain disjoint.
\end{lemma}
\begin{proof}[Proof sketch]
The offset after the tick is $(q'-p')/\eps=2m\,e_a\pm e_b$, whose Manhattan norm is $2m+1$, so the planning diamonds are disjoint by \cref{eq:diamond-gap-integer}. 
\end{proof}
In words, a robot at the minimum permitted axial separation can step aside in the same tick in which the other robot advances.

\subsection{Alignment and Euclidean Separation}\label{sec:certificate}

Discrete Coordination requires every pair of robots to have Euclidean center separation greater than $2\reps$ at each tick (\cref{eq:tick-clearance}). We now derive the exact condition under which disjoint planning diamonds guarantee this separation.

For an integer $s\ge0$, write $s=dq+j$ with $0\le j<d$. The smallest squared Euclidean norm of an integer offset with Manhattan norm at least~$s$ is
\begin{equation}\label{eq:integer-minimum}
 \begin{aligned}
 M_d(s)&:=\min_{a\in\mathbb Z^d:\norm{a}_1\ge s}\norm{a}_2^2\\
       &=(d-j)q^2+j(q+1)^2
        =\frac{s^2+j(d-j)}d,
 \end{aligned}
\end{equation}
attained exactly by the offsets whose coordinate magnitudes differ by at most one and sum to $s$.

\begin{theorem}[Sharp diamond certificate]\label{thm:diamond}
Let $\eps=\sqrt d\,r/n$ for an integer $n\ge1$, let $m\ge n$ be an integer, and let $j=(2m+1)\bmod d$. Every pair of lattice vertices with disjoint planning diamonds obeys
\begin{equation}\label{eq:diamond-gap}
 \norm{p-q}_2^2\ge\eps^2 M_d(2m+1).
\end{equation}
Disjoint planning diamonds guarantee $\norm{p-q}_2>2\reps$ for every pair of lattice vertices if and only if
\begin{equation}\label{eq:diamond-gate}
 4\bigl[m(m+1)-n^2\bigr]+1-d+j(d-j)>4d\kappa_\phi W^2 .
\end{equation}
\end{theorem}
\begin{proof}[Proof sketch]
A permitted pair has $\norm{a}_1\ge2m+1$ by \cref{eq:diamond-gap-integer}, so \cref{eq:integer-minimum} gives \cref{eq:diamond-gap}, and the offsets that attain the minimum also attain the bound. The alignment $\eps=\sqrt d\,r/n$ turns the threshold $(2\reps)^2$ into $\eps^2(4n^2/d+1+4\kappa_\phi W^2)$, and comparing it with $\eps^2M_d(2m+1)$ gives \cref{eq:diamond-gate}.
\end{proof}

In words, disjoint planning diamonds certify the Euclidean separation exactly when the footprint integer~$m$ is sufficiently large relative to the resolution integer~$n$ and the filter width~$W$.
For the box kernel, $\kappa_\phi=1/12$, and \cref{eq:diamond-gate} reduces to
\begin{equation}\label{eq:box-conditions}
 \begin{gathered}
 6\bigl[m(m+1)-n^2\bigr]>W^2\qquad(d=2),\\
 4\bigl[m(m+1)-n^2\bigr]-2+j(3-j)>W^2\qquad(d=3),
 \end{gathered}
\end{equation}
with $j=(2m+1)\bmod 3$ in three dimensions.
With the parameters of the 2D experiments in \cref{sec:experiments}, $n=7$ and $W=6$, the first condition holds at $m=n=7$, since $6\,[7\cdot8-7^2]=42>36$, so disjoint enclosing diamonds already guarantee the separation. With $n=8$ and $W=10$ in 3D, the second condition first holds at $m=9$.

\begin{remark}[The role of alignment]\label{rem:alignment}
\Cref{fig:alignment} compares two pairs of disjoint enclosing diamonds for the box kernel with $W=2$. In (a), $\eps=\sqrt2\,r$ gives the closest permitted offset $(1,2)$, with center distance $\sqrt5\,\eps>2\reps$. In (b), $\eps=4\sqrt2\,r/3$ permits the offset $(1,1)$, with center distance $\sqrt2\,\eps<2\reps$. Thus disjoint misaligned diamonds can fail the required Euclidean separation.
\end{remark}

\begin{figure}[t]
\centering
\scalebox{\figscale}{%
\begin{tikzpicture}[font=\footnotesize,x=0.52cm,y=0.52cm,
 diamond/.style={draw=blue!70!black,thick},
 robot/.style={draw=orange!80!black,thick,fill=orange!15},
 certificate/.style={draw=red!70!black,thick,densely dashed}]
\begin{scope}
 \foreach \x in {-1.5,-0.5,0.5,1.5}
   \draw[black!12,very thin] (\x,-2.15)--(\x,2.15);
 \foreach \y in {-2,-1,0,1,2}
   \draw[black!12,very thin] (-2.1,\y)--(2.1,\y);
 \foreach \cx/\cy in {-0.5/-1,0.5/1} {
   \begin{scope}[shift={(\cx,\cy)}]
    \filldraw[diamond,fill=blue!6] (1,0)--(0,1)--(-1,0)--(0,-1)--cycle;
    \draw[robot] (0,0) circle (0.707107);
    \draw[certificate] (0,0) circle (1.040833);
   \end{scope}
 }
 \draw[black!55,thin] (-0.5,-1)--(0.5,1);
 \fill (-0.5,-1) circle (1.3pt) node[below left,inner sep=2pt] {$p$};
 \fill (0.5,1) circle (1.3pt) node[above right,inner sep=2pt] {$q$};
 \node[anchor=west,align=left,inner sep=0pt] at (2.35,0) {(a) Aligned\\[1pt]$\eps=\sqrt2\,r$\\[1pt]$\norm{p-q}_2>2\reps$};
\end{scope}
\begin{scope}[shift={(8.75,0)}]
 \foreach \x in {-2,-0.666667,0.666667,2}
   \draw[black!12,very thin] (\x,-2.15)--(\x,2.15);
 \foreach \y in {-2,-0.666667,0.666667,2}
   \draw[black!12,very thin] (-2.1,\y)--(2.1,\y);
 \foreach \cx/\cy in {-0.666667/-0.666667,0.666667/0.666667} {
   \begin{scope}[shift={(\cx,\cy)}]
    \fill[blue!6] (1,0)--(0,1)--(-1,0)--(0,-1)--cycle;
   \end{scope}
 }
 \begin{scope}
  \clip (-0.666667,-0.666667) circle (1.239773);
  \fill[red!22] (0.666667,0.666667) circle (1.239773);
 \end{scope}
 \foreach \cx/\cy in {-0.666667/-0.666667,0.666667/0.666667} {
   \begin{scope}[shift={(\cx,\cy)}]
    \draw[diamond] (1,0)--(0,1)--(-1,0)--(0,-1)--cycle;
    \draw[robot] (0,0) circle (0.707107);
   \end{scope}
 }
 \foreach \cx/\cy in {-0.666667/-0.666667,0.666667/0.666667}
   \draw[certificate] (\cx,\cy) circle (1.239773);
 \draw[black!55,thin] (-0.666667,-0.666667)--(0.666667,0.666667);
 \fill (-0.666667,-0.666667) circle (1.3pt) node[below left,inner sep=2pt] {$p$};
 \fill (0.666667,0.666667) circle (1.3pt) node[above right,inner sep=2pt] {$q$};
 \node[anchor=west,align=left,inner sep=0pt] at (2.35,0) {(b) Misaligned\\[1pt]$\eps=4\sqrt2\,r/3$\\[1pt]$\norm{p-q}_2<2\reps$};
\end{scope}
\end{tikzpicture}}
\caption{Aligned and misaligned enclosing diamonds for the same robot radius and the box kernel with $W=2$, drawn at the same scale. Blue diamonds enclose the orange robot bodies. Red dashed circles are certificate balls, whose overlap in (b) is shaded. Both diamond pairs are disjoint.}
\label{fig:alignment}
\end{figure}

\subsection{Parameter Selection and Footprint Inflation}\label{sec:parameter-rule}
Among the footprint integers that satisfy \cref{thm:diamond}, smaller ones admit more vertices near obstacles and more joint moves.
We therefore select the smallest certified footprint integer $m_*$.

\begin{definition}[Certified footprint selection]\label{def:diamond-rule}\label{def:footprint-selection}
The least certified footprint integer at resolution $n$ and width $W$ is
\begin{equation}\label{eq:footprint-selection}
 \begin{gathered}
m_*(n,W)=\min\Bigl\{m\in\mathbb Z_{\ge n}:\
 \eps^2M_d(2m+1) >4\reps^2\Bigr\},
 \end{gathered}
\end{equation}
with~$\reps^2=r^2+\frac{\eps^2}{4}+\kappa_\phi v^2\Tf^2$. Substituting the aligned pitch \cref{eq:diamond-alignment} into \cref{def:filter-duration} gives
\begin{equation}\label{eq:width-n}
 W(n)=2\Bigl\lceil\frac{c_\phi v^2\,n}{2a_{\max}\sqrt d\,r}\Bigr\rceil ,\qquad \Tf=W(n)\dt(n).
\end{equation}
The construction uses $m=m_*(n):=m_*(n,W(n))$.
\end{definition}

The set in \cref{eq:footprint-selection} is nonempty, so $m_*$ is well defined, and $m_*(n,W)$ does not decrease as $W$ grows. If $m_*=n$, the enclosing diamond itself serves as the planning diamond. Otherwise, the footprint is inflated.

\begin{remark}[Resolution and footprint inflation]\label{rem:resolution-inflation}
For fixed speed and acceleration caps, \Cref{eq:width-n} shows that~$W$ grows linearly with~$n$, up to integer rounding.
At~$m_*=n$, the left side of \cref{eq:diamond-gate} therefore grows linearly in~$n$ and the right side quadratically, so a sufficiently fine lattice requires~$m_*>n$. 
A lower acceleration cap increases~$W(n)$ and can require inflation even at coarse resolutions.

\Cref{fig:clearance} compares the resolutions $n=2$ and $n=6$ in the plane under the same caps, for the box kernel and $a_{\max}=\sqrt2\,v^2/r$, which give $T_{\mathrm{acc}}=\sqrt2\,r/v$ and $W(n)=2\lceil n/2\rceil$. The coarse resolution selects the enclosing diamond $m_*=2$, and the fine resolution selects the inflated footprint $m_*=7$ at the same filter duration.
\end{remark}

\begin{figure}[t]
\centering
\scalebox{\figscale}{%
\begin{tikzpicture}[>=Latex,font=\footnotesize,x=0.52cm,y=0.52cm]
\begin{scope}
 \draw[step=0.707107,black!12,very thin] (-1.697,-1.697) grid (3.818,3.111);
 \foreach \cx/\cy in {0/0,2.121320/1.414214} {
   \begin{scope}[shift={(\cx,\cy)}]
   \fill[red!30] (0,0) circle (1.136515);
   \filldraw[fill=blue!8,draw=blue!70!black,thick] (1.414214,0)--(0,1.414214)--(-1.414214,0)--(0,-1.414214)--cycle;
   \draw[red!70!black,thick,densely dashed] (0,0) circle (1.136515);
   \filldraw[fill=orange!15,draw=orange!80!black,thick] (0,0) circle (1);
   \fill (0,0) circle (1.1pt);
   \end{scope}
 }
 \draw[black!70] (0,0)--(2.121320,1.414214);
 \node[below left,inner sep=1pt] at (0,0) {$p$};
 \node[above right,inner sep=1pt] at (2.121320,1.414214) {$q$};
 \node at (1.06,-2.45) {(a) $n=2$, $W=2$, $m_*=2$};
\end{scope}
\begin{scope}[shift={(7.4,0)}]
 \draw[step=0.235702,black!10,very thin] (-1.886,-1.886) grid (3.771,3.536);
 \foreach \cx/\cy in {0/0,1.885618/1.649916} {
   \begin{scope}[shift={(\cx,\cy)}]
   \fill[red!30] (0,0) circle (1.086534);
   \filldraw[fill=blue!8,draw=blue!70!black,thick] (1.649916,0)--(0,1.649916)--(-1.649916,0)--(0,-1.649916)--cycle;
   \draw[blue!60!black,dashed] (1.414214,0)--(0,1.414214)--(-1.414214,0)--(0,-1.414214)--cycle;
   \draw[red!70!black,thick,densely dashed] (0,0) circle (1.086534);
   \filldraw[fill=orange!15,draw=orange!80!black,thick] (0,0) circle (1);
   \fill (0,0) circle (1.1pt);
   \end{scope}
 }
 \draw[black!70] (0,0)--(1.885618,1.649916);
 \node[below left,inner sep=1pt] at (0,0) {$p$};
 \node[above right,inner sep=1pt] at (1.885618,1.649916) {$q$};
 \draw[->] (1.885618,1.649916)--(3.535534,1.649916);
 \node[right,inner sep=1pt] at (3.535534,1.649916) {$h$};
 \node at (0.94,-2.45) {(b) $n=6$, $W=6$, $m_*=7$};
\end{scope}
\end{tikzpicture}}

\smallskip
\scalebox{\figscale}{\footnotesize
\begin{tabular}{@{}lcc@{}}
\toprule
 & (a) & (b)\\\midrule
Filter duration $\Tf$ & $\sqrt2\,r/v$ & $\sqrt2\,r/v$\\
Half-width $h$ & $\sqrt2\,r$ & $7\sqrt2\,r/6$\\
$\reps^2/r^2$ & $31/24$ & $85/72$\\
Closest offset $a$ and $\norm{a}_2^2$ & $(3,2)$ and $13$ & $(8,7)$ and $113$\\
Certificate threshold $(2\reps/\eps)^2$ & $31/3$ & $85$\\\bottomrule
\end{tabular}}
\caption{Two resolutions with the same physical limits and filter duration, drawn at the same scale. Orange discs are robot bodies, red dashed circles are certificate balls, solid outlines are planning diamonds, and the dashed diamond in (b) is the enclosing diamond. The finer resolution in (b) requires footprint inflation.}
\label{fig:clearance}
\end{figure}

With $m=m_*(n)$, the diamond vertex set, the diamond graph that it induces in $\mathcal G_\eps$, and the configuration set are
\begin{align}
 V_D&=\{p\in V_\eps:\diamondbody_{m}(p)\cap\forbidden=\varnothing\},\quad \mathcal G_D=\mathcal G_\eps[V_D],\label{eq:diamond-sites}\\
 \mathcal Q_D&=\bigl\{Q\in V_D^N:\diamondbody_m(p_i)\cap\diamondbody_m(p_j)=\varnothing\ \forall i<j\bigr\},\label{eq:diamond-configurations}
\end{align}
A free planning diamond does not imply $\clr(p)>\reps$, so $V_D$ keeps both conditions.

\begin{proposition}[Euclidean clearance of diamond plans]\label{prop:diamond-certified}
With $m=m_*(n)$, $\mathcal Q_D\subseteq\mathcal Q_\eps$. Every sequence of configurations in $\mathcal Q_D$ that satisfies \cref{eq:transition} is therefore a valid plan in the sense of \cref{def:validity}.
\end{proposition}
\begin{proof}
By \cref{eq:diamond-sites}, every center lies in $V_\eps$. By \cref{eq:footprint-selection,thm:diamond}, every pair has Euclidean separation greater than $2\reps$. Both conditions of \cref{eq:tick-clearance} therefore hold.
\end{proof}

\subsection{Discrete Coordination with Planning Diamonds}\label{sec:formulation}

The planning diamonds give a restricted form of \cref{prob:la-mapf}. With endpoints $S^\eps,G^\eps\in\mathcal Q_D$, the corresponding instance is~$\mathcal I_D=(\mathcal G_D,\diamondbody_m,S^\eps,G^\eps)$.

\begin{problem}[Discrete Coordination with planning diamonds]\label{prob:discrete}
Given $\mathcal I_D$, find an integer $K\ge0$ and a valid plan $Q^{0:K}$ with $Q^k\in\mathcal Q_D$ for every $0\le k\le K$, $Q^0=S^\eps$, and $Q^K=G^\eps$.
\end{problem}

By \cref{prop:diamond-certified}, every solution of \cref{prob:discrete} also solves \cref{prob:la-mapf}. The simultaneous elementary moves define the configuration graph used by the solvers in \cref{sec:discrete}.

\begin{definition}[Configuration graph]\label{def:configuration-graph}
The configuration graph is the finite directed graph $\mathcal H_D=(\mathcal Q_D,\mathcal T_D)$ whose vertices are the valid diamond configurations and whose edges are the valid joint moves,
\begin{equation}\label{eq:configuration-graph}
 \mathcal T_D=\bigl\{(Q,Q')\in\mathcal Q_D^2:\ q_i-p_i\in\moveset\ \forall i\bigr\},
\end{equation}
for $Q=(p_1,\ldots,p_N)$ and $Q'=(q_1,\ldots,q_N)$.
\end{definition}

A solution of \cref{prob:discrete} is a path from $S^\eps$ to $G^\eps$ in $\mathcal H_D$.

\section{PIBT and LaCAM for Discrete Coordination}\label{sec:discrete}

In this section, we solve \cref{prob:discrete} with adaptations of PIBT~\cite{okumura2022priority} and LaCAM~\cite{okumura2023lacam}.
As in \gls{acr:mapf}, the adapted PIBT serves as the configuration successor generator, and the adapted LaCAM calls it internally.
Our adaptations handle simultaneous blockers, reshuffle priorities in stationary groups, and reuse backward A* searches for exact goal distances.
We denote by \acrshort{acr:takt}-PIBT and \acrshort{acr:takt}-LaCAM the two planners obtained when the adapted PIBT and the adapted LaCAM, respectively, serve as the discrete solver in \cref{alg:pipeline} with the diamond formulation.

\subsection{PIBT with Diamond Conflicts}\label{sec:generator}

We retain the priority inheritance and backtracking of PIBT~\cite{okumura2022priority}. 
In standard MAPF, a target vertex is blocked by at most one agent.
By contrast, a target diamond can intersect the current diamonds of up to~$3^d$ undecided robots (\cref{lem:push-cap}), which become simultaneous blockers whose responses can conflict across branches.
Our generator therefore backtracks over their joint target assignments within the same tick. Repeated generator calls give the discrete solver of \acrshort{acr:takt}-PIBT.

The candidates of robot $i$ are the vertices it can reach in one elementary move,
\begin{equation}\label{eq:candidates}
 C(i)=\{p_i+\delta: \delta\in\moveset,\ p_i+\delta\in V_D\},
\end{equation}
including the wait $p_i$. \Cref{alg:generator} first checks any fixed targets $\Pi$, which the configuration search of \cref{sec:search} prescribes, for membership in $C(i)$ and pairwise diamond nonintersection. \textsc{InitializeTargets} fixes those targets and leaves the remaining robots undecided.

\textsc{SelectRoot} chooses the highest-priority undecided robot $\rho$. \textsc{Respond} searches its candidates and the responses of all blocking robots, using the order in \cref{sec:ordering}. Each tentative target is checked against all fixed and tentative targets. The routine returns a status $s$ and tentative assignments $T$. A fresh budget $B$ bounds all candidate attempts for that root, including recursive calls and retries.
\begin{lemma}[Validity of generated successors]\label{lem:tick-validity}
Start from $Q\in\mathcal Q_D$ with no fixed targets. For every priority order, every candidate order, and every budget $B\ge1$, \cref{alg:generator} terminates and returns a configuration $Q'\in\mathcal Q_D$ with $q_i-p_i\in\moveset$ for every robot $i$.
\end{lemma}
\begin{proof}[Proof sketch]
Every candidate attempt is charged to the budget and at most $N$ root searches are opened, so the call terminates. Without fixed targets, every undecided robot can wait without conflicting with any committed target, and every committed search and every fallback preserves this property. A tentative target is accepted only when it is separated from every decided target, so the committed targets stay pairwise separated and $Q'\in\mathcal Q_D$.
\end{proof}
In words, the generator never returns an invalid configuration.

\begin{algorithm}[t]
\caption{PIBT successor generation for diamonds}\label{alg:generator}
\small
\begin{algorithmic}[1]
\Require Valid configuration $Q$; priority order $\prec$; budget $B$; fixed targets $\Pi$, possibly empty
\Ensure Valid successor $Q'$, or failure
\State \textbf{if not} \Call{CheckFixed}{$Q,\Pi$} \textbf{then return} failure
\State $Q'\gets{}$\Call{InitializeTargets}{$Q,\Pi$}
\While{some robot is undecided}
 \State $\rho\gets{}$\Call{SelectRoot}{$Q',\prec$}
 \State $(s,T)\gets{}$\Call{Respond}{$Q,Q',[\rho],B$}
 \If{$s=\mathrm{success}$}
  \State $Q'\gets{}$\Call{CommitTargets}{$Q',T$}
 \Else
  \State \Call{DiscardTargets}{$T$}
  \State \textbf{if} $\Pi\ne\varnothing$ \textbf{then return} failure
  \State $Q'\gets{}$\Call{CommitWait}{$Q,Q',\rho$}
 \EndIf
\EndWhile
\State \Return $Q'$
\end{algorithmic}
\end{algorithm}

\subsection{LaCAM Configuration Search}\label{sec:search}

LaCAM searches $\mathcal H_D$ using the dual-level configuration search of~\cite{okumura2023lacam}, with \cref{alg:generator} as its successor generator. Its constraints fix individual robot moves, and the generator fills the remaining assignments. A failed call under a partial constraint leaves its descendants available for exploration. When all moves are fixed, \textsc{CheckFixed} tests the complete assignment exactly, independently of the response budget. 

\begin{proposition}[Completeness on $\mathcal H_D$]\label{prop:search-completeness}
Given sufficient time and memory, \acrshort{acr:takt}-LaCAM returns a solution of \cref{prob:discrete} whenever $\mathcal H_D$ contains a path from $S^\eps$ to $G^\eps$.
\end{proposition}
\begin{proof}[Proof sketch]
The graph $\mathcal H_D$ is finite, and the search removes a node only after its constraint tree is exhausted, so it visits every configuration reachable from $S^\eps$. At each explored configuration, the constraint tree enumerates every complete assignment of the $N$ elementary moves, and a fully constrained generator call tests each assignment exactly, so every valid joint move is eventually found. 
\end{proof}

\subsection{Candidate Ordering and Priority Reshuffling}\label{sec:ordering}

\myparagraph{Candidate Ordering} --
Following~\cite{okumura2022priority}, let
$D_i(p)=\operatorname{dist}_{\mathcal G_D}(p,g_i^\eps)$
be the goal distance in moves, ignoring other robots.
Each nonzero axial move changes lattice parity and hence a finite goal distance by exactly one.
We try improving moves with $D_i(q)=D_i(p_i)-1$,
ordered by increasing $\norm{q-g_i^\eps}_2^2$;
the wait; and retreats with $D_i(q)=D_i(p_i)+1$.
A seeded random rank breaks remaining ties.
All candidates remain available for backtracking and undergo the same diamond conflict checks.

\myparagraph{Priority Reshuffling} --
In PIBT and LaCAM, the priority of a robot determines how freely it moves.
The priorities are initialized with the distance to the goal, so that robots far from their goals are processed first.
Subsequent equal increments preserve their relative order, so a group of mutually blocked robots can keep the same initiating robot tick after tick and remain stationary.
We resolve such situations with a simple priority reshuffling mechanism, which cyclically rotates the priorities within the group and thereby changes the initiating robot.

A robot is \emph{stationary} when it has remained away from its goal without moving for $F$ ticks, where $F\approx 6r/\eps$ is the number of ticks needed to travel three body diameters.
Two stationary robots are connected when~$\norm{p_i-p_j}_1\le(2m+2)\eps$, the separation at which their next moves could create a conflict, and each connected component with at least two robots forms a stationary group. 
Within each group, the second robot receives the highest priority, the previous initiating robot receives the lowest priority, and the remaining priorities shift cyclically. The rotated robots restart their count of stationary ticks.
Reshuffling occurs whenever LaCAM creates a new node and at every tick of \acrshort{acr:takt}-PIBT.

\subsection{Exact Goal Distances with Retained Backward A*}\label{sec:distance}
Precomputing full tables of the exact goal distances requires $N|V_D|$ entries.
Standard PIBT spreads this computation across queries using lazy backward breadth-first search (BFS). Its distance storage can nevertheless become a memory bottleneck on the fine lattices used by \gls{acr:takt}.
We instead keep a backward A* search from each goal on $\mathcal G_D$ and resume it on demand.

For a goal $g$, the first queried vertex $x_0$ fixes the heuristic. The search expands vertices in nondecreasing order of
\begin{equation}\label{eq:retained-f}
 f(s)=\tilde d_g(s)+h_{x_0}(s),\qquad h_{x_0}(s)=\frac{\norm{s-x_0}_1}{\eps},
\end{equation}
where $\tilde d_g(s)$ is the tentative path length from $g$ to $s$. The Manhattan heuristic directs the initial search toward $x_0$. A query resumes the search until its vertex has been settled and expanded, and subsequent queries retain the same heuristic, settled distances, and frontier. 

\begin{figure}[t]
\centering
\scalebox{\figscale}{%
\begin{tikzpicture}[x=0.142cm,y=0.142cm,font=\footnotesize]
\begin{scope}
 \draw[step=1,black!10,very thin] (-0.5,-0.5) grid (29.5,12.5);
 \foreach \x/\y in {21/6,22/5,22/6,22/7,23/4,23/5,23/6,23/7,23/8,24/3,24/4,24/5,24/6,24/7,24/8,24/9,25/0,25/2,25/3,25/4,25/5,25/6,25/7,25/8,25/9,25/10,25/12,26/0,26/1,26/2,26/3,26/4,26/5,26/6,26/7,26/8,26/9,26/10,26/11,26/12,27/0,27/1,27/2,27/3,27/4,27/5,27/6,27/7,27/8,27/9,27/10,27/11,27/12,28/0,28/1,28/2,28/3,28/4,28/5,28/6,28/7,28/8,28/9,28/10,28/11,28/12,29/0,29/1,29/2,29/3,29/4,29/5,29/6,29/7,29/8,29/9,29/10,29/11,29/12} \fill[black!25] (\x,\y) circle (0.4pt);
 \foreach \x/\y in {0/0,0/1,0/2,0/3,0/4,0/5,0/6,0/7,0/8,0/9,0/10,0/11,0/12,1/0,1/1,1/2,1/3,1/4,1/5,1/6,1/7,1/8,1/9,1/10,1/11,1/12,2/0,2/1,2/2,2/3,2/4,2/5,2/6,2/7,2/8,2/9,2/10,2/11,2/12,3/0,3/1,3/2,3/3,3/4,3/5,3/6,3/7,3/8,3/9,3/10,3/11,3/12,4/0,4/1,4/2,4/3,4/4,4/5,4/6,4/7,4/8,4/9,4/10,4/11,4/12,5/0,5/1,5/2,5/3,5/4,5/5,5/6,5/7,5/8,5/9,5/10,5/11,5/12,6/0,6/1,6/2,6/3,6/4,6/5,6/6,6/7,6/8,6/9,6/10,6/11,6/12,7/0,7/1,7/2,7/3,7/4,7/5,7/7,7/8,7/9,7/10,7/11,7/12,8/0,8/1,8/2,8/3,8/4,8/5,8/6,8/7,8/8,8/9,8/10,8/11,8/12,9/0,9/1,9/2,9/3,9/4,9/5,9/6,9/7,9/8,9/9,9/10,9/11,9/12,10/0,10/1,10/2,10/3,10/4,10/5,10/6,10/7,10/8,10/9,10/10,10/11,10/12,11/0,11/1,11/2,11/3,11/4,11/5,11/6,11/7,11/8,11/9,11/10,11/11,11/12,12/0,12/1,12/2,12/3,12/4,12/5,12/6,12/7,12/8,12/9,12/10,12/11,12/12,13/0,13/1,13/2,13/3,13/4,13/5,13/6,13/7,13/8,13/9,13/10,13/11,13/12,14/0,14/1,14/11,14/12,15/0,15/1,15/2,15/3,15/4,15/5,15/6,15/7,15/8,15/9,15/10,15/11,15/12,16/0,16/1,16/2,16/3,16/4,16/5,16/6,16/7,16/8,16/9,16/10,16/11,16/12,17/0,17/1,17/2,17/3,17/4,17/5,17/6,17/7,17/8,17/9,17/10,17/11,17/12,18/0,18/1,18/2,18/3,18/4,18/5,18/6,18/7,18/8,18/9,18/10,18/11,18/12,19/0,19/1,19/2,19/3,19/4,19/5,19/6,19/7,19/8,19/9,19/10,19/11,19/12,20/0,20/1,20/2,20/3,20/4,20/5,20/6,20/7,20/8,20/9,20/10,20/11,20/12,21/0,21/1,21/2,21/3,21/5,21/7,21/8,21/9,21/10,21/11,21/12,22/0,22/1,22/2,22/3,22/4,22/8,22/9,22/10,22/11,22/12,23/0,23/1,23/2,23/3,23/9,23/10,23/11,23/12,24/0,24/1,24/2,24/10,24/11,24/12,25/1,25/11} \fill[blue!65!black] (\x,\y) circle (0.95pt);
 \foreach \x/\y in {14/2,14/3,14/4,14/5,14/6,14/7,14/8,14/9,14/10} \fill[black!40] (\x-0.5,\y-0.5) rectangle (\x+0.5,\y+0.5);
 \node[circle,draw,fill=white,inner sep=0.6pt,font=\scriptsize] at (7, 6) {$g$};
 \node[circle,draw,fill=white,inner sep=0.6pt,font=\scriptsize] at (21, 4) {$x_0$};
 \node[anchor=north,inner sep=2pt,align=center] at (14.5,-0.9) {(a) Backward BFS from $g$:\\302 vertices, all distances final};
\end{scope}
\begin{scope}[shift={(31.6,0)}]
 \draw[step=1,black!10,very thin] (-0.5,-0.5) grid (29.5,12.5);
 \foreach \x/\y in {0/0,0/1,0/2,0/3,0/4,0/5,0/6,0/7,0/8,0/9,0/10,0/11,0/12,1/0,1/1,1/2,1/3,1/4,1/5,1/6,1/7,1/8,1/9,1/10,1/11,1/12,2/0,2/1,2/2,2/3,2/4,2/5,2/6,2/7,2/8,2/9,2/10,2/11,2/12,3/0,3/1,3/2,3/3,3/7,3/8,3/9,3/10,3/11,3/12,4/0,4/1,4/2,4/8,4/9,4/10,4/11,4/12,5/0,5/1,5/9,5/10,5/11,5/12,6/0,6/10,6/11,6/12,7/11,7/12,8/11,8/12,9/11,9/12,10/11,10/12,11/11,11/12,12/11,12/12,13/11,13/12,14/11,14/12,15/6,15/7,15/8,15/9,15/10,15/11,15/12,16/6,16/7,16/8,16/9,16/10,16/11,16/12,17/6,17/7,17/8,17/9,17/10,17/11,17/12,18/6,18/7,18/8,18/9,18/10,18/11,18/12,19/6,19/7,19/8,19/9,19/10,19/11,19/12,20/6,20/7,20/8,20/9,20/10,20/11,20/12,21/6,21/7,21/8,21/9,21/10,21/11,21/12,22/0,22/5,22/6,22/7,22/8,22/9,22/10,22/11,22/12,23/0,23/1,23/2,23/3,23/4,23/5,23/6,23/7,23/8,23/9,23/10,23/11,23/12,24/0,24/1,24/2,24/3,24/4,24/5,24/6,24/7,24/8,24/9,24/10,24/11,24/12,25/0,25/1,25/2,25/3,25/4,25/5,25/6,25/7,25/8,25/9,25/10,25/11,25/12,26/0,26/1,26/2,26/3,26/4,26/5,26/6,26/7,26/8,26/9,26/10,26/11,26/12,27/0,27/1,27/2,27/3,27/4,27/5,27/6,27/7,27/8,27/9,27/10,27/11,27/12,28/0,28/1,28/2,28/3,28/4,28/5,28/6,28/7,28/8,28/9,28/10,28/11,28/12,29/0,29/1,29/2,29/3,29/4,29/5,29/6,29/7,29/8,29/9,29/10,29/11,29/12} \fill[black!25] (\x,\y) circle (0.4pt);
 \foreach \x/\y in {4/4,4/5,4/6,5/3,5/4,5/5,5/6,5/7,6/2,6/3,6/4,6/5,6/6,6/7,6/8,7/1,7/2,7/3,7/4,7/5,7/7,7/8,7/9,8/1,8/2,8/3,8/4,8/5,8/6,8/7,8/8,8/9,9/1,9/2,9/3,9/4,9/5,9/6,9/7,9/8,9/9,10/1,10/2,10/3,10/4,10/5,10/6,10/7,10/8,10/9,11/1,11/2,11/3,11/4,11/5,11/6,11/7,11/8,11/9,12/1,12/2,12/3,12/4,12/5,12/6,12/7,12/8,12/9,13/1,13/2,13/3,13/4,13/5,13/6,13/7,13/8,13/9,14/1,15/1,15/2,15/3,15/4,16/1,16/2,16/3,16/4,17/1,17/2,17/3,17/4,18/1,18/2,18/3,18/4,19/1,19/2,19/3,19/4,20/1,20/2,20/3,20/4,21/1,21/2,21/3} \fill[blue!65!black] (\x,\y) circle (0.95pt);
 \foreach \x/\y in {3/4,3/5,3/6,4/3,4/7,5/2,5/8,6/1,6/9,7/0,7/10,8/0,8/10,9/0,9/10,10/0,10/10,11/0,11/10,12/0,12/10,13/0,13/10,14/0,15/0,15/5,16/0,16/5,17/0,17/5,18/0,18/5,19/0,19/5,20/0,20/5,21/0,21/5,22/1,22/2,22/3,22/4} \draw[orange!90!black,line width=0.6pt] (\x,\y) circle (0.85pt);
 \foreach \x/\y in {14/2,14/3,14/4,14/5,14/6,14/7,14/8,14/9,14/10} \fill[black!40] (\x-0.5,\y-0.5) rectangle (\x+0.5,\y+0.5);
 \node[circle,draw,fill=white,inner sep=0.6pt,font=\scriptsize] at (7, 6) {$g$};
 \node[circle,draw,fill=white,inner sep=0.6pt,font=\scriptsize] at (21, 4) {$x_0$};
 \node[anchor=north,inner sep=2pt,align=center] at (14.5,-0.9) {(b) Backward A* toward $x_0$:\\107 settled and 42 tentative vertices};
\end{scope}
\end{tikzpicture}}
\caption{First-query comparison of goal-rooted searches, where $x_0$ is the queried vertex. (a) Backward BFS expands by distance from $g$ and assigns every discovered vertex closer than $x_0$. (b) Backward A* orders expansion by \cref{eq:retained-f}, which concentrates the explored region near routes to $x_0$. 
}
\label{fig:distance}
\end{figure}
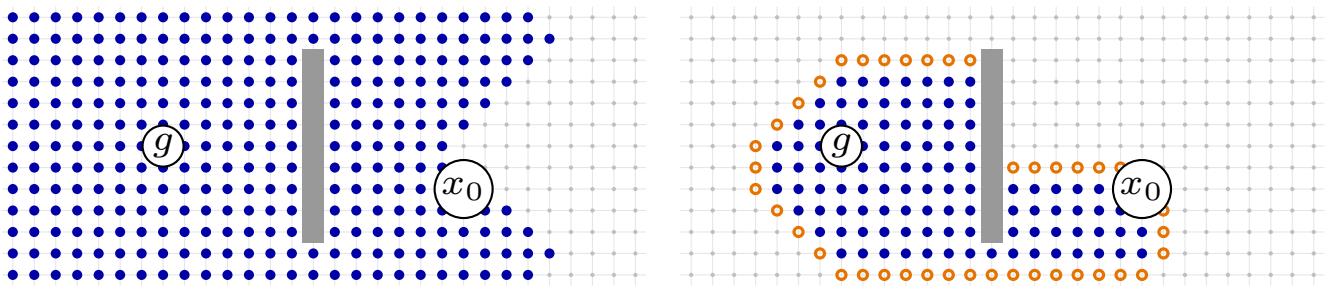

The heuristic is consistent with unit-cost axial moves, and a suspended search resumes with its queues and stored values unchanged. Every completed query therefore returns the exact goal distance $D_i$ at the queried vertex, independently of the order and timing of the queries.
\Cref{fig:distance} compares the first query with that of backward BFS and shows the memory reduction.

\subsection{Computational Cost}\label{sec:cost}

Diamond conflicts can be checked using nearby robots only. The following bound limits the number of immediate blockers of one target.
\begin{lemma}[Number of blocking robots]\label{lem:push-cap}
In a valid diamond configuration, at most $3^d$ robots have planning diamonds that intersect the planning diamond at a given target.
\end{lemma}
\begin{proof}[Proof sketch]
The centers of the blocking robots lie within Manhattan distance $2h$ of the target and are pairwise more than $2h$ apart, where $h=m\eps$. The open Manhattan balls of radius $h$ around them are disjoint and lie in the Manhattan ball of radius $3h$ around the target, whose volume is $3^d$ times theirs.
\end{proof}
Recursive dependencies may still involve many robots, so the budget $B$ remains necessary to bound the search effort.

Two targets can conflict only if their current robot positions are within Euclidean distance $(2m+2)\eps$. A spatial hash rebuilt at each tick collects these neighbors. The packing argument also bounds each neighbor list by a constant $\gamma$ depending only on the dimension $d$. Given the priority order and goal distances, one generator call costs
\begin{equation}\label{eq:tick-cost}
 O(N+N\gamma+NB\gamma)
\end{equation}
for the hash, the neighbor lists, and at most $NB$ candidate attempts, each checking $O(\gamma)$ neighbors. Sorting the priorities costs $O(N\log N)$, and priority reshuffling visits each neighbor list once. For a fixed budget $B$, one generator call therefore costs $O(N\log N)$ given the goal distances, independently of the number of lattice vertices. 
Note that \cref{eq:tick-cost} bounds a single generator call, and that TAKT-LaCAM may invoke the generator several times at one configuration.

\section{Guarantees and Resolution}\label{sec:theory}

\Cref{sec:pwm,sec:diamond,sec:discrete} reduce the \gls{acr:kmampp} to discrete coordination and reconstruct smooth trajectories from the resulting plans. 
In this section, we establish two properties in opposite directions.
First, \emph{soundness} states that every solution of the reference problem (\cref{prob:la-mapf}) reconstructs into a solution of the original problem (\cref{prob:mampp}).
Second, \emph{existence} states that a solution of \cref{prob:mampp} with sufficient margin implies that \cref{prob:la-mapf} has a solution.
Most of the analysis concerns the reference problem and is therefore independent of the discrete solver, and the results for the diamond formulation (\cref{prob:discrete}) are stated separately.

\subsection{Continuous Soundness}\label{sec:safety}

\begin{theorem}[Continuous soundness]\label{thm:soundness}
For any valid plan $Q^{0:K}$ of the reference problem, its reconstructed trajectories $\tx_1,\ldots,\tx_N$ of \cref{def:smoothing} are conflict-free throughout $[0,T_p]$, with $T_p=(K+2W)\dt$. Each trajectory is $C^1$, has absolutely continuous velocity, and satisfies the speed cap $\norm{\dot{\tx}_i(t)}_2\le v$. The trajectories start at $Q^0$ and end at $Q^K$, with zero velocity at both endpoints. If $\Tf\ge T_{\mathrm{acc}}$, they also satisfy the acceleration cap and are admissible. Under this condition, the reconstruction of a solution of \cref{prob:la-mapf} solves \cref{prob:mampp}.
\end{theorem}
\begin{proof}
Let $z=\hx_i-c$ for a fixed $c\in\forbidden$, with $(L,R)=(v,\reps)$, or let $z=\hx_i-\hx_j$, with $(L,R)=(2v,2\reps)$. In both cases, $z$ is $L$-Lipschitz and affine on each tick, and its values at the ticks have norm greater than $R$ by \cref{eq:tick-clearance}. For a point $z(s)=(1-\theta)a+\theta b$ between consecutive tick values $a$ and $b$, the identity $\norm{z(s)}_2^2=(1-\theta)\norm{a}_2^2+\theta\norm{b}_2^2-\theta(1-\theta)\norm{a-b}_2^2$ and $\norm{a-b}_2\le L\dt$ give $\norm{z(s)}_2^2>R^2-L^2\dt^2/4$.
Let $Z,Z'$ be independent random variables with density $\phi$. The variance identity and the Lipschitz bound give
\begin{equation*}
\begin{aligned}
 &\mathbb E\norm{z(t-Z)}_2^2-\norm{\mathbb E\,z(t-Z)}_2^2\\
 &\quad=\tfrac12\mathbb E\norm{z(t-Z)-z(t-Z')}_2^2\le L^2\sigk^2 .
\end{aligned}
\end{equation*}
Since $\mathbb E\,z(t-Z)=(\phi*z)(t)$, we obtain $\norm{(\phi*z)(t)}_2^2>R^2-L^2(\dt^2/4+\sigk^2)$, and the right-hand side equals $r^2$ in the first case and $4r^2$ in the second by \cref{eq:inflation}. In the first case, $(\phi*z)(t)=\tx_i(t)-c$, and taking $c$ as a point of $\forbidden$ nearest to $\tx_i(t)$ gives $\clr(\tx_i(t))>r$. In the second case, the common kernel gives $(\phi*z)(t)=\tx_i(t)-\tx_j(t)$, so $\norm{\tx_i(t)-\tx_j(t)}_2>2r$. \Cref{lem:filter} supplies the regularity and the derivative bounds. For $t\le\Tf/2$ and for $t\ge T_p-\Tf/2$, the padded waits make $\hx_i$ constant on the kernel window, so $\tx_i$ rests at $p_i^0$ and at $p_i^K$, which gives the endpoint conditions.
\end{proof}
In words, validity at the ticks alone certifies the continuous motion, whichever discrete solver produced the plan.

\begin{corollary}[Certified pipeline output]\label{cor:pipeline}
Every discrete plan produced within \cref{alg:pipeline} by either solver of \cref{sec:discrete} reconstructs into an admissible, conflict-free motion from $S^\eps$ to its final configuration $Q^K$, at rest at both ends. If $Q^K=G^\eps$, the motion solves \cref{prob:mampp}.
\end{corollary}
\begin{proof}
The selected filter duration enforces the acceleration cap. By \cref{lem:tick-validity} and the definition of $\mathcal H_D$, every returned plan is a diamond plan. By \cref{prop:diamond-certified}, every diamond plan is a valid reference plan, so \cref{thm:soundness} applies.
\end{proof}
Together, \cref{thm:soundness} and \cref{cor:pipeline} establish that the pipeline never returns an unsafe or dynamically infeasible motion.
Neither result guarantees that a discrete plan exists.
The next two subsections establish sufficient conditions for such plans to exist.

\subsection{Existence on the Reference Graph}\label{sec:reference-existence}

We next ask when a continuous solution can be represented by a valid solution of \cref{prob:la-mapf}. The axial moves and waits have velocities in~$\uset=\operatorname{conv}\{0,\pm ve_a:a=1,\ldots,d\}=\{u:\norm{u}_1\le v\}$.
Any continuous solution satisfying the Euclidean speed cap can be slowed by a common factor $\sqrt d$ so that its velocities lie in $\uset$. This slowdown preserves clearances and simultaneous separations. A greedy lattice tracker then stays within $d\eps$ of each slowed reference position at every tick. The following theorem gives a sufficient clearance condition for the tracked configurations to form a solution of \cref{prob:la-mapf}.

\begin{theorem}[Existence on the reference graph]\label{thm:completeness}
Suppose \cref{prob:mampp} has a solution $X$ from $S$ to $G$ of makespan $T$ and margin $\mu_X\ge\mu>0$. Fix a lattice of pitch $\eps$ containing the prescribed endpoints $S$ and $G$, and a positive even filter width $W$ with its associated certificate radius $\reps$. If
\begin{equation}\label{eq:completeness}
 2d\eps+2(\reps-r)<\mu,
\end{equation}
then \cref{prob:la-mapf} has a solution from $S$ to $G$. One such plan has reconstructed duration at most
\begin{equation}\label{eq:completeness-time}
 \sqrt d\,T+(2W+d+1)\dt.
\end{equation}
When $\Tf\ge T_{\mathrm{acc}}$, its reconstruction solves \cref{prob:mampp}.
\end{theorem}
\begin{proof}
Set $y_i(t)=x_i(\min\{t/\sqrt d,T\})$ and $t_k=k\dt$. Since $\norm{u}_1\le\sqrt d\,\norm{u}_2$, the slowed solution satisfies $\norm{\dot y_i}_1\le v$, so $\norm{y_i(t_{k+1})-y_i(t_k)}_1\le\eps$. Starting at $p_i^0=s_i$, each robot maintains $\norm{p_i^k-y_i(t_k)}_1\le d\eps$. The error $e=y_i(t_{k+1})-p_i^k$ satisfies $\norm{e}_1\le(d+1)\eps$. If $\norm{e}_1>d\eps$, some coordinate of $e$ exceeds $\eps$ in absolute value, and the elementary move toward $y_i(t_{k+1})$ along this coordinate reduces $\norm{e}_1$ by $\eps$. Otherwise the robot waits.
At tick $K_0=\lceil\sqrt d\,T/\dt\rceil$, the slowed solution has reached $G$. Since $p_i^{K_0}$ and $g_i$ are lattice vertices at Manhattan distance at most $d\eps$, each robot reaches its goal in at most $d$ further moves, each of which reduces this distance. At every tick, \cref{eq:clearance-lipschitz} and the margin of $X$ give $\clr(p_i^k)\ge r+\mu-d\eps$ and $\norm{p_i^k-p_j^k}_2\ge2r+\mu-2d\eps$, which exceed $\reps$ and $2\reps$ by \cref{eq:completeness}. The plan is therefore valid. It has at most $K_0+d$ ticks with $K_0\dt<\sqrt d\,T+\dt$, and the $2W$ padded waits give \cref{eq:completeness-time}. When $\Tf\ge T_{\mathrm{acc}}$, \cref{thm:soundness} applies to its reconstruction.
\end{proof}

In words, a discrete plan exists whenever the continuous solution has more margin than the lattice tracking error~$2d\epsilon$ plus the certificate inflation~$2(r_\epsilon-r)$, at the price of a slowdown by a factor of at most~$\sqrt{d}$.

\begin{figure*}[t]
\centering
\includegraphics[width=\textwidth]{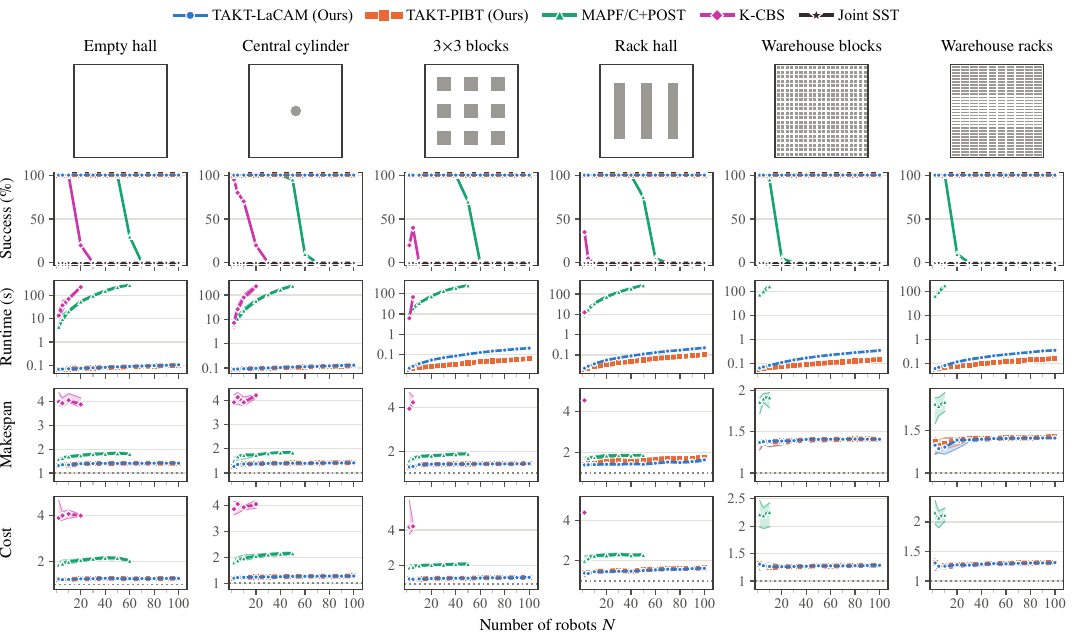}
\caption{Comparison with baselines in 2D on the six maps of the top row. Rows: success rate, runtime, and makespan and cost relative to the free-space lower bounds (dotted line at one). Lines show the median over the solved instances and bands the 25th to 75th percentile. Joint SST solves no instance.}\label{fig:exp-count-2d}
\end{figure*}

\subsection{Resolution under the Acceleration Cap}\label{sec:resolution}

We now study how the condition of \cref{thm:completeness} behaves under refinement, i.e., as~$\epsilon\to 0$ under the given speed and acceleration caps.
We first establish the limits of the parameters selected by the construction. Write
\begin{equation}\label{eq:r-acc}
 r_{\mathrm{acc}}=\sqrt{r^2+\kappa_\phi c_\phi^2\frac{v^4}{a_{\max}^2}}.
\end{equation}

\begin{proposition}[Parameter limits under refinement]\label{prop:limits}
As $\eps\to0$ with the selected filter width, $\Tf\to T_{\mathrm{acc}}$ and $\reps\to r_{\mathrm{acc}}$. Under the aligned diamond construction, the selected footprint satisfies
\begin{equation}\label{eq:footprint-limit}
 \frac{m_*(n)}{n}\to\frac{r_{\mathrm{acc}}}{r},\ 
 m_*(n)\eps(n)\to\sqrt d\,r_{\mathrm{acc}},
 \ \text{as }n\to\infty.
\end{equation}
\end{proposition}
\begin{proof}[Proof sketch]
The filter-width rounding error is less than $2\dt$, hence $\Tf\to T_{\mathrm{acc}}$ and the certificate radius tends to $r_{\mathrm{acc}}$. The diamond limit follows from the sharp footprint-selection condition, which brackets $m_*$ within one unit of $\sqrt{n^2+d\kappa_\phi W^2}$. 
\end{proof}
In words, refinement removes the tick reserve but not the smoothing reserve, which the acceleration cap sets through~$T_{\mathrm{acc}}$.
The reference certificate radius tends to $r_{\mathrm{acc}}$, while the diamond's axial radius tends to $\sqrt d\,r_{\mathrm{acc}}$, which makes it the smallest enclosing diamond for the $r_{\mathrm{acc}}$ ball. 

\begin{corollary}[Reference plans under refinement]\label{cor:resolution}
Suppose \cref{prob:mampp} has a solution $X$ of makespan $T$
with margin $\mu_X>2(r_{\mathrm{acc}}-r)$.
Then, for every sufficiently fine lattice containing $S$ and $G$, \cref{prob:la-mapf} has a solution under the selected filter width. One such plan has reconstructed duration at most $\sqrt d\,T+2\Tf+(d+1)\dt$, and its reconstruction solves \cref{prob:mampp}.
\end{corollary}
\begin{proof}
The left side of \cref{eq:completeness} tends to $2(r_{\mathrm{acc}}-r)<\mu_X$. Therefore the finite-resolution condition eventually holds, and the selected duration also enforces the acceleration cap. Apply \cref{thm:completeness}.
\end{proof}

The diamond problem imposes additional restrictions, and a stronger margin ensures both obstacle clearance of the entire footprint and pairwise footprint separation.

\begin{corollary}[Diamond plans under refinement]\label{cor:diamond-resolution}
Suppose \cref{prob:mampp} has a solution $X$ with margin
\begin{equation*}
 \mu_X>2(\sqrt d\,r_{\mathrm{acc}}-r).
\end{equation*}
Then, for all sufficiently large $n$ for which the lattice of pitch $\eps(n)$ contains $S$ and $G$, \cref{prob:discrete} has a solution with the selected planning diamonds. Its reconstruction solves \cref{prob:mampp}.
\end{corollary}
\begin{proof}[Proof sketch]
Set $h=m_*(n)\eps(n)$. By \cref{prop:limits}, $h\to\sqrt d\,r_{\mathrm{acc}}$, and a planning diamond lies inside the Euclidean ball of radius $h$ about its center. The tracked plan behind \cref{thm:completeness} keeps obstacle clearance at least $r+\mu_X-d\eps$ and pairwise separation at least $2r+\mu_X-2d\eps$, which exceed the diamond thresholds $h$ and $2h$ once $2d\eps+2(h-r)<\mu_X$. The stated margin makes this inequality hold for all sufficiently large $n$. 
\end{proof}

The additional factor $\sqrt d$ in the diamond result reflects the enclosing diamond. The increase from $r$ to $r_{\mathrm{acc}}$ reflects the smoothing reserve. With the caps of the experiments in \cref{sec:experiments}, $r_{\mathrm{acc}}=0.527$\,m for $r=0.5$\,m in 2D and $r_{\mathrm{acc}}=0.231$\,m for $r=0.2$\,m in 3D, so the smoothing reserve costs about 3\,cm of clearance in the limit. Instances below the stated margin thresholds may still have valid plans.
Together, these results close the argument for the implemented pipeline.
Under the margin condition of \cref{cor:diamond-resolution}, a diamond plan exists on every sufficiently fine lattice containing the endpoints. Given sufficient time and memory, \acrshort{acr:takt}-LaCAM finds such a plan (\cref{prop:search-completeness}). \Cref{thm:soundness} certifies its continuous reconstruction once a discrete plan is found.

\section{Real-Time Planning}\label{sec:online}

\Cref{alg:pipeline} returns a reference trajectory for every robot, with position, velocity, and acceleration at all times, which a low-level controller on the robot tracks at high frequency.
When goals or workspace conditions change during execution, the planner must update these trajectories, and we call the time that an update takes its \emph{computation time}.
Because planning takes time, a replan starts from the state of the fleet at a lead time in the future, and it must return within this lead time~\cite{frazzoli2002real,hauser2012responsiveness}. 
In this section, we present two paradigms for real-time planning with \gls{acr:takt}.
The lead time that the fleet can afford differs between the two paradigms, so each paradigm has its own criterion for real-time planning.

\begin{definition}[Real-time planning]\label{def:real-time}
Replanning is \emph{real-time} when its computation time is at most 500\,ms (2\,Hz) under whole-fleet replanning (\cref{sec:online-whole}) or one tick $\dt$ per decision cycle under incremental replanning (\cref{sec:online-incremental}).
\end{definition}

The whole-fleet criterion lies within the 0.1--1\,s cycles of existing single-robot planners~\cite{kunz2010real,werling2010optimal,mohta2018fast,petti2005safe,leven2002framework}.
Robots track the committed motion while a replacement is computed.
Incremental replanning has no trajectory beyond the supplied configurations and must append one configuration per tick: 101\,ms in our 2D experiments and 43\,ms in 3D.

\subsection{Whole-Fleet Replanning}\label{sec:online-whole}

This paradigm is the default setting for the planner-controller interaction~\cite{frazzoli2002real,kunz2010real,hauser2012responsiveness}. 
When an update is needed at time $t$, the planner computes a complete replacement plan from a future configuration on the current discrete plan. The existing and replacement sequences are joined at that configuration and reconstructed with the same filter, while the robots continue executing the committed motion.
\Cref{fig:showcase} shows an example with 24 quadrotors. After the no-fly zone appears, \gls{acr:takt} computes new reference trajectories for all robots within 30\,ms, and the controllers then switch to the new trajectories. 
\Cref{fig:exp-count-2d,fig:exp-density-2d} show that in the 2D experiments, \acrshort{acr:takt}-PIBT produces complete trajectories within 500\,ms, in the median over 20 instances, for up to around 750 robots in the empty hall and up to 200 robots in the hall with three racks. 
The incremental paradigm extends real-time planning to larger fleets by using the one-step structure of PIBT.

\subsection{Incremental Replanning}\label{sec:online-incremental}
This paradigm rests on two observations.
First, by \cref{eq:dac}, the reference position $\tx_i(t)$, velocity $\dot{\tx}_i(t)$, and acceleration $\ddot{\tx}_i(t)$ that a controller needs at time $t$ depend on the discrete motion only up to $t+\Tf/2$, so the reference trajectory can be constructed on demand as execution advances.
Second, unlike LaCAM, PIBT has a closed-loop structure and constructs its solution one configuration at a time~\cite{li2025fico,li2026adaptive}.
In every tick, the paradigm therefore runs one \emph{decision cycle}, which reads the changes that arrived since the last cycle, calls \cref{alg:generator} once, and appends the returned configuration to the discrete motion, which stays at least $\Tf/2$ ahead of the execution.
A change in the problem simply enters the next decision cycle.
Since each cycle computes one configuration in place of the entire plan, the paradigm meets the one-tick criterion for fleets for which a complete replan takes longer than 500\,ms (\cref{sec:exp-online}).
Moreover, every appended configuration is valid by \cref{lem:tick-validity}, so the executed motion remains the reconstruction of a valid plan, and the guarantee of \cref{thm:soundness} holds.

\section{Experimental Evaluation}\label{sec:experiments}
We evaluate \acrshort{acr:takt}-PIBT and \acrshort{acr:takt}-LaCAM through baseline comparisons (\cref{sec:exp-comparison}), scalability tests (\cref{sec:exp-scalability}), ablations (\cref{sec:exp-ablation}), and physics simulations of trajectory tracking and online replanning (\cref{sec:exp-tracking,sec:exp-online}) in 2D and 3D.
Baseline comparisons cover up to 100 robots, the largest fleet size solved by a baseline; scalability tests reach approximately a thousand robots.

\subsection{Experimental Setup}\label{sec:exp-setup}
\begin{figure}[t]
\centering
\includegraphics[width=\columnwidth]{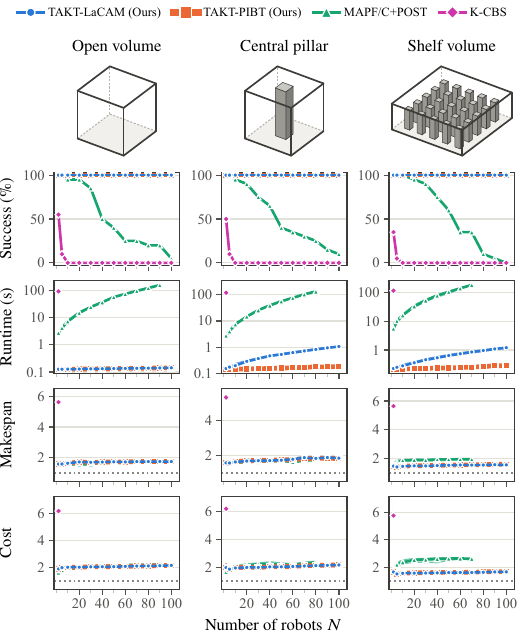}
\caption{Comparison with baselines in 3D. Rows and conventions as in \cref{fig:exp-count-2d}.}\label{fig:exp-count-3d}
\end{figure}

\myparagraph{Instances} --
The 2D instances use the six maps in the top row of \cref{fig:exp-count-2d}. Four are halls of 105.5\,m by 105.5\,m: an empty hall, a hall with a central cylinder, a hall with a $3\times3$ array of blocks, and a hall with three racks (the rack hall). Two are warehouses of 183\,m by 183\,m, one with blocks (the block warehouse) and one with racks (the rack warehouse), whose aisles are 3\,m wide. The robots are discs with $r=0.5$\,m, $v=1$\,m/s, and $a_{\max}=3.5$\,m/s$^2$. The 3D instances use the three volumes in the top row of \cref{fig:exp-count-3d}: an open cube of side 8.5\,m (the open volume), a cube of side 8.8\,m with a central pillar of 2\,m by 2\,m (the central pillar volume), and a room of 13.7\,m by 13.7\,m by 4.8\,m that holds 25 shelves separated by aisles of 1.46\,m (the shelf volume). The robots are spheres with $r=0.2$\,m, $v=1$\,m/s, and $a_{\max}=5$\,m/s$^2$, an acceleration cap below gravity.

\begin{figure*}[t]
\centering
\includegraphics[width=\textwidth]{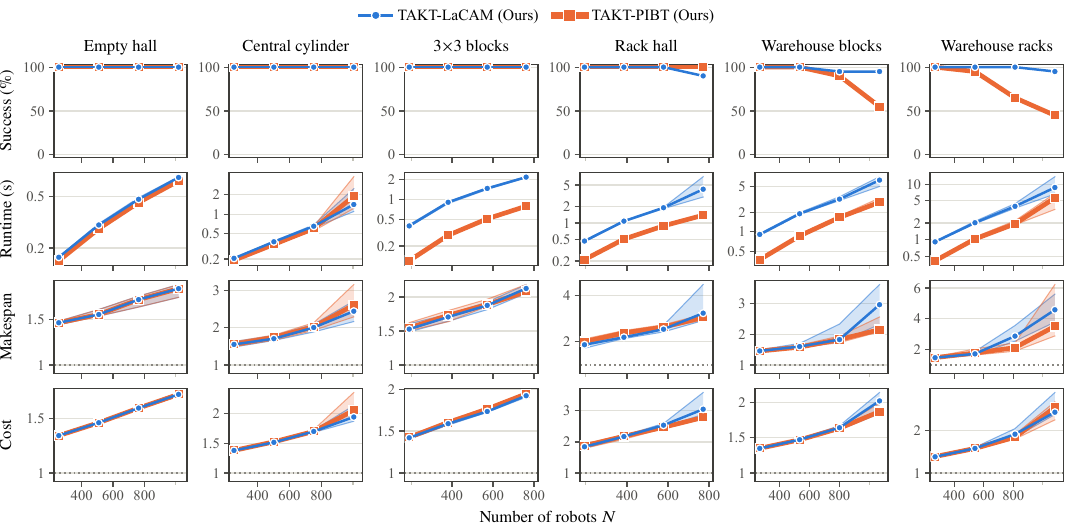}
\caption{Scalability of the \acrshort{acr:takt} planners in 2D. 
}\label{fig:exp-density-2d}
\end{figure*}

\begin{figure}[t]
\centering
\includegraphics[width=\columnwidth]{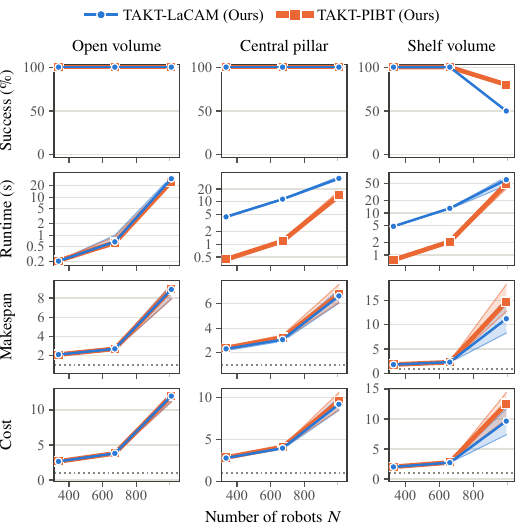}
\caption{Scalability of the \gls{acr:takt} planners in 3D. Rows and conventions as in \cref{fig:exp-count-2d}.}\label{fig:exp-density-3d}
\end{figure}

Every instance draws its start and goal positions at random. 
For the baseline comparisons (\cref{fig:exp-count-2d,fig:exp-count-3d}), we evaluate fleet sizes $N\in\{2,5,10,20,30,\ldots,100\}$, with 20 instances per map and fleet size.
For the scalability evaluation (\cref{fig:exp-density-2d,fig:exp-density-3d}), we increase the fleets to approximately 1,000 robots.

\begin{figure}[t]
\centering
\includegraphics[width=\figscale\columnwidth]{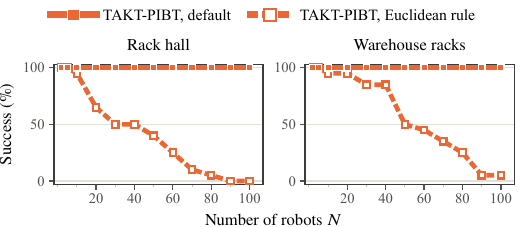}
\caption{\textbf{Effect of the diamond conflict rule.} Success rate of \acrshort{acr:takt}-PIBT under the diamond rule and under the Euclidean rule.}\label{fig:exp-abl-rule}
\end{figure}

\myparagraph{\acrshort{acr:takt} Settings} --
We reconstruct with the box kernel and select the parameters by \cref{def:diamond-rule}. In 2D experiments, $n=7$ gives $\eps=0.101$\,m, $\dt=0.101$\,s, $W=6$, and $m_*=7$. In 3D experiments, $n=8$ gives $\eps=0.043$\,m, $\dt=0.043$\,s, $W=10$, and $m_*=9$. 
The endpoint tolerance is $\eta=0.09$\,m in 2D and $\eta=0.05$\,m in 3D.

\myparagraph{Baselines} --
We compare with three planners that generate trajectories for double-integrator robots without offline precomputation.
Methods such as db-CBS and db-LaCAM~\cite{moldagalieva2024dbcbs,moldagalieva2025dblacam} rely on precomputed primitive libraries, and thus fall outside this evaluation setting.
(i) MAPF/C+POST~\cite{honig2018trajectory} plans a discrete schedule with ECBS~\cite{barer2014suboptimal} on a grid of pitch $3r$ under generalized conflicts, optimizes one trajectory per robot inside safe corridors, and stretches the time axis until the dynamic constraints hold. We run the authors' implementation with MATLAB. 
(ii) K-CBS~\cite{kottinger2022conflict} plans each robot with a kinodynamic tree search and resolves the collisions between robots through the constraint tree of \gls{acr:cbs}. 
(iii) Joint SST runs the SST planner~\cite{li2016asymptotically} in the composite state space of the fleet. 
We run K-CBS and joint SST from the benchmark suite released with db-CBS~\cite{moldagalieva2024dbcbs}. Joint SST enters the 2D comparison only.

\myparagraph{Protocol} --
All methods plan the same instances under the same speed and acceleration caps, on one laptop with an Apple M3 Pro processor and 36\,GB of memory. Every instance has a time limit of 300\,s. 
A run succeeds when it returns, within its limit, trajectories that solve \cref{prob:mampp}. 

\myparagraph{Metrics} --
We report four measures. 
The \emph{success rate} is the percentage of instances that a method solves. 
The \emph{runtime} is the planning time of a solved instance. For \acrshort{acr:takt} it covers all three stages in~\cref{sec:overview}. For MAPF/C+POST it excludes the start-up of MATLAB.
The \emph{makespan} and the \emph{cost} are those of \cref{sec:mrmp}, the cost with unit weights $q_p=q_v=r_a=1$. 
We divide both by a lower bound of the instance, in which every robot travels alone on the straight line from its start to its goal, without obstacles and under the same caps. The bound of the makespan is the largest minimum travel time over the robots, $\norm{g_i-s_i}_2/v+v/a_{\max}$ for robot $i$, and the bound of the cost is the sum of the minimum costs of the robots, which we compute numerically. 
Every curve shows the median over the instances that a method solves, with a band from the 25th to the 75th percentile, and it is drawn where the method solves at least 20\,\% of the instances.

\subsection{Comparison with Kinodynamic Planners}\label{sec:exp-comparison}

\Cref{fig:exp-count-2d,fig:exp-count-3d} show the comparisons with baselines in both 2D and 3D.

\myparagraph{Success and Runtime} --
Both \gls{acr:takt} planners solve every instance on every map.
With 100 robots, the median runtime of \acrshort{acr:takt}-LaCAM is 0.1\,s in the empty hall, 0.4\,s in the two warehouses, and 1.2\,s in the shelf volume, and \acrshort{acr:takt}-PIBT is faster on every map.

The baselines stop at small fleets, at a size that depends on the map.
MAPF/C+POST solves every instance of the four halls up to 40 robots and none beyond 60 robots, and in the two warehouses it solves none beyond 20 robots.
In 3D, it solves 8 to 12 of the 20 instances of each map with 50 robots and at most 2 with 100 robots.
K-CBS solves instances with 10 or more robots only in the empty hall and in the hall with a central cylinder, and joint SST solves none.

Where a baseline succeeds, both \gls{acr:takt} planners are orders of magnitude faster.
In the empty hall with 50 robots, MAPF/C+POST takes 230\,s and \acrshort{acr:takt}-LaCAM 0.09\,s.
At this fleet size, \acrshort{acr:takt}-LaCAM is 2,200 to 2,600 times faster in each of the four halls and 105 to 415 times faster on each of the three 3D maps.
The speedup grows with the fleet size, since the runtime of both \gls{acr:takt} planners is almost flat while the runtime of MAPF/C+POST rises toward the time limit.

\begin{figure}[t]
\centering
\includegraphics[width=\figscale\columnwidth]{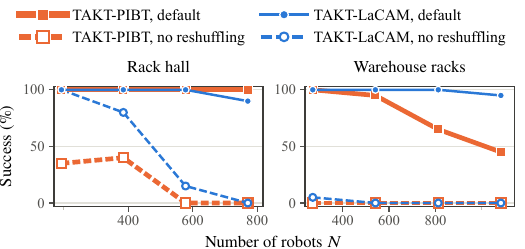}
\caption{\textbf{Effect of priority reshuffling.} Success rate with and without the reshuffling of \cref{sec:ordering} on the instances of \cref{fig:exp-density-2d}.}\label{fig:exp-abl-reshuffle}
\end{figure}

\begin{figure}[t]
\centering
\includegraphics[width=\figscale\columnwidth]{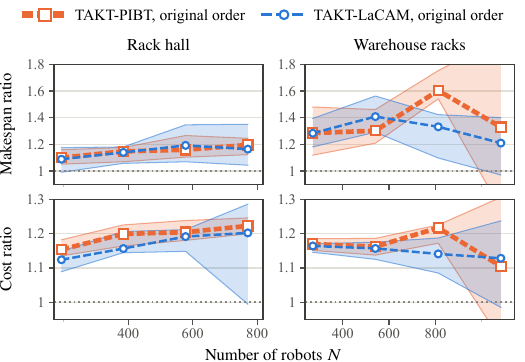}
\caption{\textbf{Effect of the candidate order.} Makespan and cost under the candidate order of the original PIBT, relative to the default order on the same instance (dotted line at one), over the instances that both solve.
}\label{fig:exp-abl-order}
\end{figure}

\begin{figure}[t]
\centering
\includegraphics[width=\figscale\columnwidth]{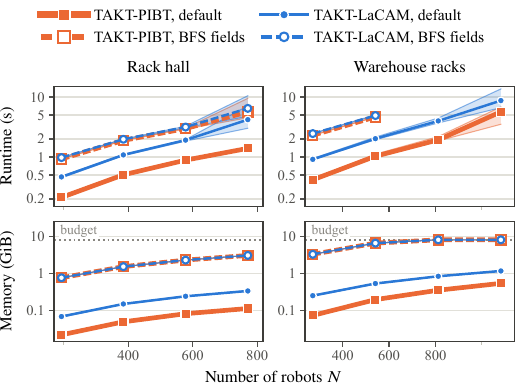}
\caption{\textbf{Effect of the retained distance search.} Runtime and peak memory of the goal-distance computation over tested instances, with the retained backward A* and with backward BFS fields under a budget of 8\,GB.
}\label{fig:exp-abl-distance}
\end{figure}

\begin{figure*}[t]
\centering
\scalebox{1}{\input{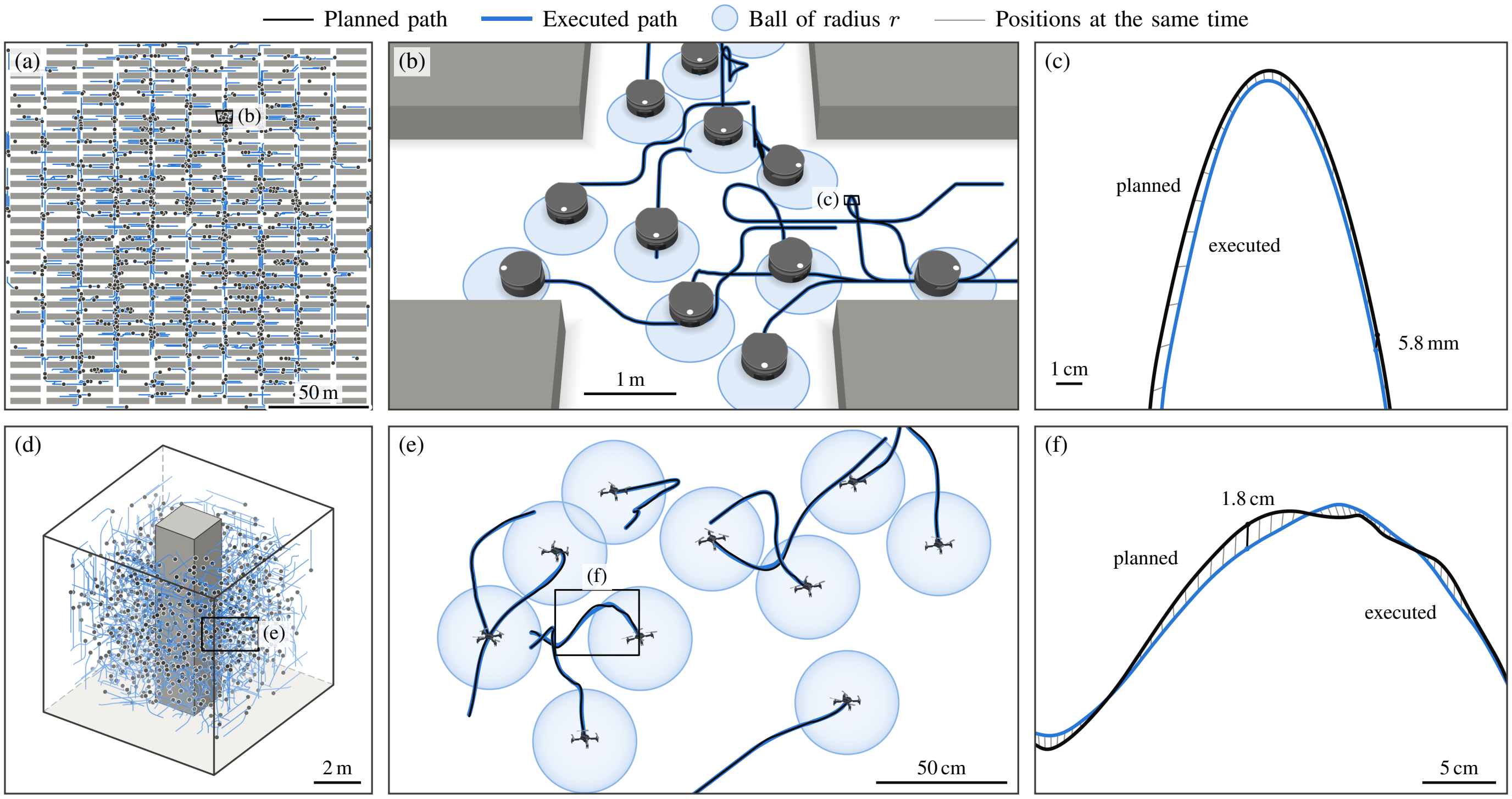}}
\caption{\textbf{Simulated robots executing plans of \acrshort{acr:takt}-LaCAM} in the physics simulator ARGoS~\cite{pinciroli2012argos}. Top: 1,082 RB-1 Base robots in the rack warehouse. Bottom: 1,006 Crazyflie quadrotors around the central pillar. (a), (d) The fleet, with one enlarged dot per robot. (b), (e) The regions marked in (a) and (d), in which each robot lies inside the ball of radius $r$ around its planned position. In (b), the white dot of a base is its tracked point. (c), (f) The windows marked in (b) and (e), labeled with the largest distance between the planned and the executed position at the same time.}\label{fig:exp-tracking}
\end{figure*}

\myparagraph{Solution Quality} --
In 2D, the makespan and the cost of the \gls{acr:takt} planners lie below those of both baselines in every panel of \cref{fig:exp-count-2d}.
In the empty hall with 50 robots, both \gls{acr:takt} planners reach 1.40 times the makespan bound and 1.27 times the cost bound, against 1.83 and 2.17 times for MAPF/C+POST.
K-CBS lies at about four times both bounds where it succeeds.

In 3D, the solution quality of the \gls{acr:takt} planners is comparable with that of MAPF/C+POST in the open volume and the central pillar volume.
In the shelf volume, both \gls{acr:takt} planners lead at every fleet size, and with 50 robots their makespan is 21\,\% shorter and their cost 38\,\% lower.

Overall, both \gls{acr:takt} planners solve every instance on every map within seconds, where the baselines stop at small fleets and take up to three orders of magnitude longer.
Their solution quality leads on every 2D map and in the shelf volume in 3D, and it is comparable with that of MAPF/C+POST on the other two 3D maps.

\subsection{Scalability}\label{sec:exp-scalability}

\Cref{fig:exp-density-2d,fig:exp-density-3d} show the \acrshort{acr:takt} planners on fleets that are about ten times larger than those of the comparison.
The largest fleets hold 762 to 1,082 robots in 2D and about 1,000 robots in 3D.

\myparagraph{2D Instances} --
The \acrshort{acr:takt} planners plan a thousand robots within seconds.
In the empty hall, both planners solve every instance with 1,018 robots, with a median runtime of 0.7\,s.
The runtime is longest in the rack warehouse, where \acrshort{acr:takt}-LaCAM plans 1,082 robots in 8.7\,s and solves 19 of the 20 instances.
\acrshort{acr:takt}-PIBT solves every instance of the four halls, and in the hall with $3\times3$ blocks and in the rack hall it is two to three times faster than \acrshort{acr:takt}-LaCAM.

The makespan and the cost grow with the fleet size on every map, since the robots wait for one another more often.
In the empty hall, the makespan grows from 1.5 to 1.8 times its bound between 254 and 1,018 robots.
In the rack warehouse, it grows from 1.5 times its bound with 271 robots to 3.5 times for \acrshort{acr:takt}-PIBT and 4.6 times for \acrshort{acr:takt}-LaCAM with 1,082 robots.

\myparagraph{3D Instances} --
In the open volume, both planners solve every instance and plan 673 robots in about 0.7\,s.
In the central pillar and the shelf volume, \acrshort{acr:takt}-PIBT plans fleets of this size within about 2\,s and \acrshort{acr:takt}-LaCAM within about 13\,s.
With about 1,000 robots, the runtime and the makespan rise on all three maps.
In the open volume, \acrshort{acr:takt}-PIBT then needs 26\,s, and the makespan grows from 2.7 to 8.9 times its bound.
In the shelf volume with nearly a thousand robots, \acrshort{acr:takt}-PIBT solves 16 of the 20 instances.

The footprint inflation of \cref{rem:resolution-inflation} contributes to this rise.
Under the cap of 5\,m/s$^2$, the certified footprint integer $m_*=9$ exceeds $n=8$, and the planning diamonds of 1,000 robots cover about 21\,\% of the lattice vertices.
In 2D, $m_*=n$ and no inflation occurs.

Overall, the \acrshort{acr:takt} planners plan about a thousand robots in 0.7\,s in the empty hall, in up to 8.7\,s in the rack warehouse, and in tens of seconds on the 3D maps.

\subsection{Ablation Studies}\label{sec:exp-ablation}

In this subsection, we change one design element at a time and plan the 2D instances with both planners.
We report each ablation on the rack hall and the rack warehouse.

\begin{figure*}[t]
\centering
\scalebox{\figscale}{\input{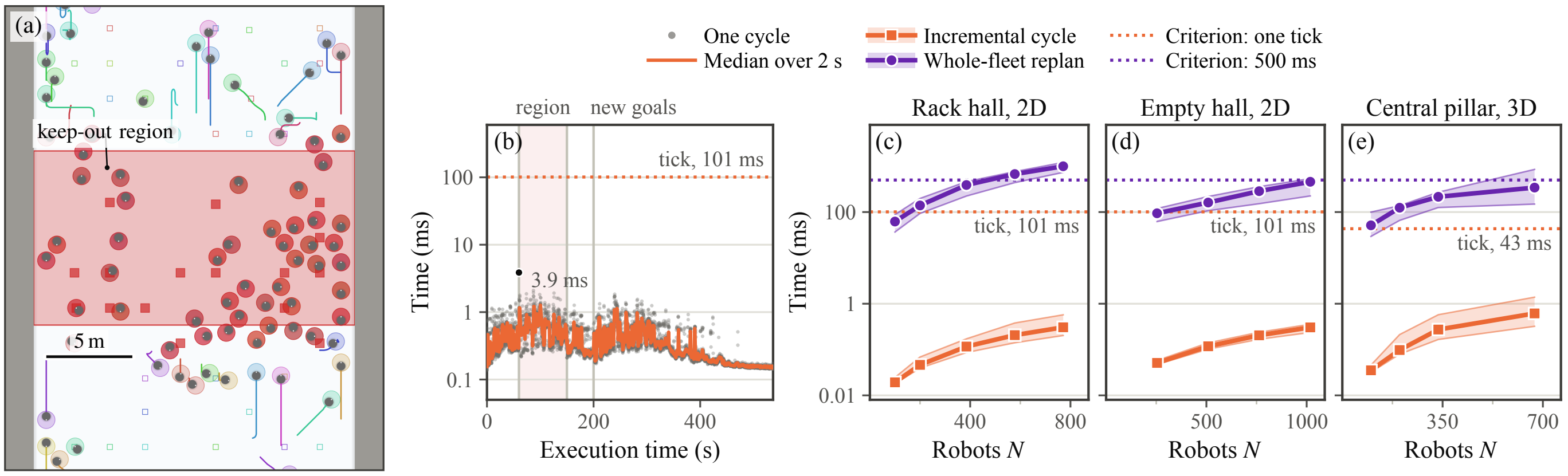}}
\caption{\textbf{Real-time planning.} (a) One incremental session with 771 simulated RB-1 Base robots in the rack hall, 4\,s after a keep-out region appears. The robots that are inside or about to enter the region are halted (red discs), and the goals inside it are suspended (red squares). (b) Computation time of every decision cycle of this session, with the interval of the region shaded. (c)--(e) Computation time against the fleet size, over ten sessions per fleet. Lines show the median and bands the 25th to 75th percentile. Dotted lines mark the real-time criterion of each paradigm (\cref{sec:online}).}\label{fig:exp-online}
\end{figure*}

\myparagraph{Diamond Conflict Rule} --
We replace the diamond conflict test of \cref{eq:diamond-configurations} with the Euclidean separation $\norm{p_i-p_j}_2>2\reps$ of \cref{eq:tick-clearance}.
According to~\cref{prop:diamond-certified}, the Euclidean rule admits every configuration that the diamond rule admits, so the planners search a larger set of configurations.
However, as shown in \cref{fig:exp-abl-rule}, the success rate falls significantly.
On the instances with 50 robots, \acrshort{acr:takt}-PIBT solves 8 of the 20 instances in the rack hall and 10 in the rack warehouse under the Euclidean rule.
With 100 robots, it solves none and one.
Under the diamond rule, it solves all of these instances.
The Euclidean rule admits additional configurations in which an axially aligned pair can be too close for a simultaneous advance and sideways step to preserve pairwise separation.
The diamond rule excludes these configurations, and a robot at the minimum permitted separation can still yield sideways within one tick (\cref{lem:sidestep}).

\myparagraph{Priority Reshuffling} --
Without the reshuffling of \cref{sec:ordering}, both planners lose most instances on the two maps shown in \cref{fig:exp-abl-reshuffle}.
In the rack hall, \acrshort{acr:takt}-PIBT solves 35 and 40\,\% of the instances at the two smaller fleet sizes and none at the two larger ones, while \acrshort{acr:takt}-LaCAM shows a similar decline.
In the rack warehouse, \acrshort{acr:takt}-PIBT solves no instances and \acrshort{acr:takt}-LaCAM solves only one of 80.
In the failed runs, stationary groups repeatedly retain the same initiating robot. Priority reshuffling changes the initiating robot and enables different yielding attempts.

\myparagraph{Candidate Order} --
The original PIBT~\cite{okumura2022priority} orders the improving moves of a robot by a random tie order alone, and \cref{sec:ordering} orders them by the Euclidean distance to the goal.
With the original order, the cost and the makespan rise on both maps and at every fleet size (\cref{fig:exp-abl-order}).
In the rack hall, the cost rises by 12 to 22\,\% and the makespan by 9 to 20\,\%.
In the rack warehouse, the cost rises by 10 to 22\,\% and the makespan by 21 to 60\,\%.
Among the improving moves, the Euclidean order prefers the move closest to the goal.
A robot therefore alternates between the axes, and its reconstructed trajectory approaches the straight line.

\myparagraph{Exact Goal Distances} --
We replace the retained backward A* of \cref{sec:distance} with the lazy backward BFS of the original PIBT, under a memory budget of 8\,GB.
Both computations return the same distances, and they differ in memory and time (\cref{fig:exp-abl-distance}).
In the rack hall with 771 robots, the BFS fields take 3.1\,GB, and the retained search takes 0.1\,GB under \acrshort{acr:takt}-PIBT and 0.3\,GB under \acrshort{acr:takt}-LaCAM.
The median runtime in the rack hall grows by a factor of 3.4 to 4.4 for \acrshort{acr:takt}-PIBT and of 1.5 to 2.1 for \acrshort{acr:takt}-LaCAM.
In the rack warehouse, the BFS fields exceed the memory budget at the two largest fleet sizes, and both planners then fail on every instance.
The retained search takes at most 1.2\,GB there.

\begin{table}[t]
\centering\footnotesize
\caption{Tracking of \acrshort{acr:takt}-LaCAM trajectories by the simulated robots. Tracking error $e$ at the 99th percentile and at its maximum. Reach: largest distance from a planned position to a point of the robot body. Gap: smallest distance between two bodies and between a body and an obstacle.}\label{tab:exp-tracking}
\setlength{\tabcolsep}{3.5pt}
\begin{tabular}{@{}lrrrrrrr@{}}
\toprule
Platform & $N$ & $e_{99}$ & $e_{\max}$ & Reach & $r$ & \multicolumn{2}{c}{Gap (cm)}\\
\cmidrule(l){7-8}
 & & (cm) & (cm) & (cm) & (cm) & Bodies & Obstacle\\
\midrule
RB-1 Base & 381--1,082 & 0.53 & 0.72 & 46.4 & 50 & 15.5 & 10.7\\
Crazyflie 2.x & 331--1,010 & 1.4 & 5.0 & 11.1 & 20 & 32.8 & 18.5\\
\bottomrule
\end{tabular}
\end{table}

\subsection{Execution by Simulated Robots}\label{sec:exp-tracking}

A plan of \acrshort{acr:takt} keeps the balls of radius $r$ around the planned positions clear of the obstacles and of one another (\cref{thm:soundness}).
A robot therefore executes its trajectory without collisions when its body stays inside the ball of radius $r$ around its planned position.
We test this condition on a differential-drive base and on a quadrotor, each of which tracks its trajectory with a standard off-the-shelf controller.

\myparagraph{Platforms} --
In 2D, every robot is a simulated Robotnik RB-1 Base, a differential-drive base with a body radius of 25.75\,cm, planned with $r=50$\,cm.
In 3D, every robot is a simulated Crazyflie~2.x quadrotor~\cite{giernacki2017crazyflie} with a body radius of 6.1\,cm, planned with $r=20$\,cm.
The RB-1 Base tracks its trajectory through a layered PI controller~\cite{deluca1998feedback}, and the quadrotor through the PX4 autopilot~\cite{meier2015px4}.

\myparagraph{Results} --
We execute six 2D and six 3D plans of \acrshort{acr:takt}-LaCAM from \cref{sec:exp-scalability}, with 331 to 1,082 robots.
\Cref{fig:exp-tracking} shows one fleet of each platform, and \cref{tab:exp-tracking} gives the measurements.
Tracking errors are on the millimeter scale for the bases and the centimeter scale for the quadrotors (\cref{fig:exp-tracking}(c) and (f)).
In every fleet, the tracking error stays within 0.72\,cm for the bases and 5.0\,cm for the quadrotors.
Every point of a robot body therefore stays inside the ball of radius $r$ around its planned position, with at least 3.6\,cm to spare for a base and 8.9\,cm for a quadrotor.
In every run, every body stays at least 10.7\,cm away from the other bodies and from the obstacles.

\subsection{Real-Time Planning}\label{sec:exp-online}

We now run the two paradigms of \cref{sec:online} while the fleet moves and its task changes to demonstrate the real-time planning performance.
The runs test temporary keep-out regions (also called no-fly zones in 3D, as in~\cref{fig:showcase}) and reassignment of every robot's goal (\cref{fig:exp-online}(a)).
We measure the computation time of each paradigm and compare it with its criterion of \cref{sec:online}, which is 500\,ms for a whole-fleet replan and one tick for an incremental decision cycle.

\myparagraph{Sessions} --
The fleets hold 100 to 771 robots in the rack hall, 254 to 1,018 robots in the empty hall, and 100 to 671 quadrotors in the central pillar volume.
Each fleet runs ten sessions per paradigm, in which the changes occur at random times.
The measured computation time covers one complete replan under whole-fleet replanning and every decision cycle under incremental replanning.

\myparagraph{Whole-Fleet Replanning} --
The computation time of whole-fleet replanning grows with fleet size (\cref{fig:exp-online}(c)--(e)). The replans meet the 500\,ms criterion for up to 200 robots in the rack hall, 763 robots in the empty hall, and 335 quadrotors in the central pillar volume.
The longest replan takes 0.36\,s, 0.46\,s, and 0.44\,s for these three fleets.

\myparagraph{Incremental Replanning} --
Every decision cycle of every session returns within its tick.
The incremental paradigm therefore extends real-time planning from 200 to 771 robots in the rack hall, from 763 to 1,018 robots in the empty hall, and from 335 to 671 quadrotors in the central pillar volume.
In 2D, the tick lasts 101\,ms, and the longest cycles are those that read a change (\cref{fig:exp-online}(b)). They take 4.9\,ms for 771 robots in the rack hall and 7.0\,ms for 1,018 robots in the empty hall.
A cycle takes 0.4\,ms and 0.3\,ms on average for these two fleets, so the planner computes the motion about 250 and 330 times faster than the robots execute it.

In 3D, the tick lasts 43\,ms. The cycle that reads a new no-fly zone takes 28 to 34\,ms for 100 to 671 quadrotors, of which 28 to 32\,ms are spent processing the zone at every fleet size.
A cycle takes 0.5\,ms on average for 335 quadrotors and 1.0\,ms for 671 quadrotors, so the planner computes the motion about 90 and 44 times faster than the quadrotors execute it.

Overall, \gls{acr:takt} achieves real-time whole-fleet replanning for hundreds of robots and, under the stricter one-tick criterion, scales real-time incremental replanning to fleets of a thousand robots.

\section{Conclusion and Future Work}
\label{sec:conclusion}
In this paper we presented \gls{acr:takt}, a framework that certifies continuous dynamic feasibility through discrete coordination for the \gls{acr:kmampp}.
\gls{acr:takt} bounds the clearance lost during continuous trajectory reconstruction before the search begins, and reserves that clearance margin on the workspace lattice.
A closed-form convolution, applied independently to each robot, then converts synchronized lattice moves and waits into smooth trajectories that avoid collisions and satisfy the speed and acceleration caps.
This soundness guarantee holds for every valid discrete plan, so any discrete solver that returns valid plans becomes a certified kinodynamic planner.
Our analysis also establishes sufficient clearance conditions under which a valid discrete plan exists at a fixed lattice resolution and under lattice refinement.
To coordinate robots that span multiple lattice vertices, we introduced a diamond formulation whose conflict rule supports sideways yielding in a single tick while guaranteeing the Euclidean separation required by reconstruction.
We adapted PIBT and LaCAM to this diamond formulation by backtracking over simultaneous blockers, reshuffling priorities in stationary groups, ordering candidate moves by Euclidean distance to the goal, and querying exact goal distances with retained backward A* searches.

With 50 robots, the resulting planners ran more than three orders of magnitude faster than the strongest baseline in the 2D halls and more than two orders of magnitude faster on the 3D maps, with comparable or lower makespan and cost.
The planners computed complete trajectories for about a thousand robots in seconds in 2D and in tens of seconds in 3D, and simulated differential-drive robots and quadrotors tracked these trajectories without collisions.
\gls{acr:takt} also replanned in real time under online changes to goals and keep-out regions.
Replanning the complete fleet returned within 500\,ms for hundreds of robots, and incremental replanning extended real-time planning to fleets of up to a thousand robots by computing one configuration per tick.

Future work includes tighter certificates and conflict rules for narrower passages and denser fleets; incorporating tracking uncertainty and communication delays, with hardware validation; and co-designing lattice resolution, reconstruction kernels, and dynamic caps to optimize clearance, runtime, and solution quality~\cite{zardini2022analysis}.

\ifthenelse{\boolean{anonymous}}{}{%
  \section*{Acknowledgment}
  We would like to thank Dr. Keisuke Okumura for providing suggestions for this project.
}

\IEEEtriggeratref{32}
{\bibliographystyle{IEEEtran}\bibliography{references}}

@string{aaai = {{AAAI} Conf. on Artificial Intelligence}}

@string{ai = {Artificial Intelligence}}

@string{ar = {Autonomous Robots}}

@string{as = {{IEEE} Trans. Aerosp.}}

@string{cdc = {{IEEE} Conf. on Decision and Control ({CDC})}}

@string{icaps = {Intl. Conf. on Automated Planning and Scheduling ({ICAPS})}}

@string{icra = {{IEEE} Intl. Conf. on Robotics and Automation ({ICRA})}}

@string{ijrr = {Intl. J. of Robotics Research}}

@string{iros = {{IEEE}/{RSJ} Intl. Conf. on Intelligent Robots and Systems ({IROS})}}

@string{jfr = {J. of Field Robotics}}

@string{ral = {{IEEE} Robotics and Automation Letters}}

@string{socs = {Intl. Symp. on Combinatorial Search ({SoCS})}}

@string{springer = {Springer}}

@string{tase = {{IEEE} Trans. Autom. Sci. Eng.}}

@string{tits = {{IEEE} Trans. Intell. Transp. Syst.}}

@string{tro = {{IEEE} Trans. Robot.}}

@article{okumura2022priority,
  title={Priority inheritance with backtracking for iterative multi-agent path finding},
  author={Okumura, Keisuke and Machida, Manao and D{\'e}fago, Xavier and Tamura, Yasumasa},
  journal=ai,
  volume={310},
  pages={103752},
  year={2022},
  publisher={Elsevier}
}

@inproceedings{okumura2023lacam,
  title={Lacam: Search-based algorithm for quick multi-agent pathfinding},
  author={Okumura, Keisuke},
  booktitle=aaai,
  volume={37},
  number={10},
  pages={11655--11662},
  year={2023}
}

@article{okumura2023improving,
  title={Improving lacam for scalable eventually optimal multi-agent pathfinding},
  author={Okumura, Keisuke},
  journal={arXiv preprint arXiv:2305.03632},
  year={2023}
}

@article{wurman2008coordinating,
  title={Coordinating hundreds of cooperative, autonomous vehicles in warehouses},
  author={Wurman, Peter R and D'Andrea, Raffaello and Mountz, Mick},
  journal={AI Magazine},
  volume={29},
  number={1},
  pages={9--9},
  year={2008}
}

@article{sharon2015conflict,
  title={Conflict-based search for optimal multi-agent pathfinding},
  author={Sharon, Guni and Stern, Roni and Felner, Ariel and Sturtevant, Nathan R},
  journal=ai,
  volume={219},
  pages={40--66},
  year={2015},
  publisher={Elsevier}
}

@article{wagner2015subdimensional,
  title={Subdimensional expansion for multirobot path planning},
  author={Wagner, Glenn and Choset, Howie},
  journal=ai,
  volume={219},
  pages={1--24},
  year={2015},
  publisher={Elsevier}
}

@article{zardini2022analysis,
  title={Analysis and control of autonomous mobility-on-demand systems},
  author={Zardini, Gioele and Lanzetti, Nicolas and Pavone, Marco and Frazzoli, Emilio},
  journal={Annual Review of Control, Robotics, and Autonomous Systems},
  volume={5},
  number={1},
  pages={633--658},
  year={2022},
  publisher={Annual Reviews}
}

@article{stern2019mapf,
  title={Multi-Agent Pathfinding: Definitions, Variants, and Benchmarks},
  author={Roni Stern and Nathan R. Sturtevant and Ariel Felner and Sven Koenig and Hang Ma and Thayne T. Walker and Jiaoyang Li and Dor Atzmon and Liron Cohen and T. K. Satish Kumar and Eli Boyarski and Roman Bartak},
  journal=socs,
  year={2019},
  pages={151--158}
}

@book{kirk2004optimal,
  title={Optimal control theory: an introduction},
  author={Kirk, Donald E},
  year={2004},
  publisher={Dover Publications},
  address={Mineola, NY},
  note={Reprint of the 1970 Prentice-Hall edition}
}

@article{kalman1960contributions,
  title={Contributions to the theory of optimal control},
  author={Kalman, Rudolf E},
  journal={Bolet{\'i}n de la Sociedad Matem{\'a}tica Mexicana},
  volume={5},
  pages={102--119},
  year={1960}
}

@article{d2012guest,
  title={Guest editorial: A revolution in the warehouse: A retrospective on {K}iva {S}ystems and the grand challenges ahead},
  author={D'Andrea, Raffaello},
  journal=tase,
  volume={9},
  number={4},
  pages={638--639},
  year={2012},
  publisher={IEEE}
}

@article{li2025fico,
  title={FICO: Finite-Horizon Closed-Loop Factorization for Unified Multi-Agent Path Finding},
  author={Li, Jiarui and Zanardi, Alessandro and Pecora, Federico and Zhang, Runyu and Zardini, Gioele},
  journal={arXiv preprint arXiv:2511.13961},
  year={2025}
}

@inproceedings{barer2014suboptimal,
  title={Suboptimal variants of the conflict-based search algorithm for the multi-agent pathfinding problem},
  author={Barer, Max and Sharon, Guni and Stern, Roni and Felner, Ariel},
  booktitle=socs,
  volume={5},
  number={1},
  pages={19--27},
  year={2014}
}

@inproceedings{moldagalieva2025dblacam,
  author    = {Moldagalieva, Akmaral and Okumura, Keisuke and Prorok, Amanda and H{\"o}nig, Wolfgang},
  title     = {{db-LaCAM}: Fast and Scalable Multi-Robot Kinodynamic Motion Planning with Discontinuity-Bounded Search and Lightweight {MAPF}},
  booktitle = icaps,
  volume    = {36},
  number    = {1},
  pages     = {738--746},
  year      = {2026}
}

@article{jiang2026mdpibt,
  author  = {Jiang, Zixiang and Zhang, Yulun and Veerapaneni, Rishi and Li, Jiaoyang},
  title   = {Planning over {MAPF} agent dependencies via multi-dependency {PIBT}},
  journal = {arXiv preprint arXiv:2603.23405},
  year    = {2026},
  note    = {Accepted to IEEE/RSJ IROS 2026}
}

@inproceedings{yukhnevich2025epibt,
  author    = {Yukhnevich, Egor and Andreychuk, Anton},
  title     = {Enhancing {PIBT} via Multi-Action Operations},
  booktitle = aaai,
  volume    = {40},
  number    = {35},
  pages     = {29865--29873},
  year      = {2026},
  doi       = {10.1609/aaai.v40i35.40233}
}

@inproceedings{li2019large,
  author    = {Li, Jiaoyang and Surynek, Pavel and Felner, Ariel and Ma, Hang and Kumar, T. K. Satish and Koenig, Sven},
  title     = {Multi-Agent Path Finding for Large Agents},
  booktitle = aaai,
  volume    = {33},
  number    = {1},
  pages     = {7627--7634},
  year      = {2019},
  doi       = {10.1609/aaai.v33i01.33017627}
}

@article{honig2018trajectory,
  author  = {H{\"o}nig, Wolfgang and Preiss, James A. and Kumar, T. K. Satish and Sukhatme, Gaurav S. and Ayanian, Nora},
  title   = {Trajectory planning for quadrotor swarms},
  journal = tro,
  volume  = {34},
  number  = {4},
  pages   = {856--869},
  year    = {2018}
}

@article{cap2015prioritized,
  author  = {{\v{C}}{\'a}p, Michal and Nov{\'a}k, Peter and Kleiner, Alexander and Seleck{\'y}, Martin},
  title   = {Prioritized Planning Algorithms for Trajectory Coordination of Multiple Mobile Robots},
  journal = tase,
  volume  = {12},
  number  = {3},
  pages   = {835--849},
  year    = {2015},
  doi     = {10.1109/TASE.2015.2445780},
  note    = {arXiv:1409.2399}
}

@article{donald1993kinodynamic,
  author  = {Donald, Bruce and Xavier, Patrick and Canny, John and Reif, John},
  title   = {Kinodynamic motion planning},
  journal = {Journal of the ACM},
  volume  = {40},
  number  = {5},
  pages   = {1048--1066},
  year    = {1993},
  doi     = {10.1145/174147.174150}
}

@article{li2026adaptive,
  author  = {Li, Jiarui and Pecora, Federico and Zhang, Runyu and Zardini, Gioele},
  title   = {Adaptive-horizon conflict-based search for closed-loop multi-agent path finding},
  journal = ral,
  volume  = {11},
  number  = {9},
  pages   = {10505--10512},
  year    = {2026},
  doi     = {10.1109/LRA.2026.3711523},
  note    = {arXiv:2602.12024}
}

@inproceedings{li2026certificate,
  author    = {Li, Jiarui and Zhang, Runyu and Zardini, Gioele},
  title     = {Certificate-driven closed-loop multi-agent path finding with inheritable factorization},
  booktitle = cdc,
  year      = {2026},
  note      = {In press. arXiv:2604.00428}
}

@inproceedings{moldagalieva2024dbcbs,
  author    = {Moldagalieva, Akmaral and Ortiz-Haro, Joaquim and Toussaint, Marc and H{\"o}nig, Wolfgang},
  title     = {{db-CBS}: Discontinuity-Bounded Conflict-Based Search for Multi-Robot Kinodynamic Motion Planning},
  booktitle = icra,
  pages     = {14569--14575},
  year      = {2024},
  doi       = {10.1109/ICRA57147.2024.10610999},
  note      = {arXiv:2309.16445}
}

@inproceedings{park2020efficient,
  author    = {Park, Jungwon and Kim, Junha and Jang, Inkyu and Kim, H. Jin},
  title     = {Efficient Multi-Agent Trajectory Planning with Feasibility Guarantee using Relative {Bernstein} Polynomial},
  booktitle = icra,
  pages     = {434--440},
  year      = {2020},
  doi       = {10.1109/ICRA40945.2020.9197162},
  note      = {arXiv:1909.10219}
}

@article{lavalle2001randomized,
  author  = {LaValle, Steven M. and Kuffner, James J.},
  title   = {Randomized Kinodynamic Planning},
  journal = ijrr,
  volume  = {20},
  number  = {5},
  pages   = {378--400},
  year    = {2001},
  doi     = {10.1177/02783640122067453}
}

@article{erdmann1987multiple,
  author  = {Erdmann, Michael and Lozano-P{\'e}rez, Tom{\'a}s},
  title   = {On Multiple Moving Objects},
  journal = {Algorithmica},
  volume  = {2},
  number  = {1--4},
  pages   = {477--521},
  year    = {1987},
  doi     = {10.1007/BF01840371}
}

@inproceedings{augugliaro2012generation,
  author    = {Augugliaro, Federico and Schoellig, Angela P. and D'Andrea, Raffaello},
  title     = {Generation of collision-free trajectories for a quadrocopter fleet: A sequential convex programming approach},
  booktitle = iros,
  pages     = {1917--1922},
  year      = {2012},
  doi       = {10.1109/IROS.2012.6385823}
}

@inproceedings{mellinger2011minimum,
  author    = {Mellinger, Daniel and Kumar, Vijay},
  title     = {Minimum snap trajectory generation and control for quadrotors},
  booktitle = icra,
  pages     = {2520--2525},
  year      = {2011},
  doi       = {10.1109/ICRA.2011.5980409}
}

@article{pinciroli2012argos,
  author  = {Pinciroli, Carlo and Trianni, Vito and O'Grady, Rehan and Pini, Giovanni and Brutschy, Arne and Brambilla, Manuele and Mathews, Nithin and Ferrante, Eliseo and Di Caro, Gianni and Ducatelle, Frederick and Birattari, Mauro and Gambardella, Luca Maria and Dorigo, Marco},
  title   = {{ARGoS}: A modular, parallel, multi-engine simulator for multi-robot systems},
  journal = {Swarm Intelligence},
  volume  = {6},
  number  = {4},
  pages   = {271--295},
  year    = {2012},
  doi     = {10.1007/s11721-012-0072-5}
}

@inproceedings{meier2015px4,
  author    = {Meier, Lorenz and Honegger, Dominik and Pollefeys, Marc},
  title     = {{PX4}: A node-based multithreaded open source robotics framework for deeply embedded platforms},
  booktitle = icra,
  pages     = {6235--6240},
  year      = {2015},
  doi       = {10.1109/ICRA.2015.7140074}
}

@inproceedings{panerati2021learning,
  author    = {Panerati, Jacopo and Zheng, Hehui and Zhou, SiQi and Xu, James and Prorok, Amanda and Schoellig, Angela P.},
  title     = {Learning to fly---a gym environment with {PyBullet} physics for reinforcement learning of multi-agent quadcopter control},
  booktitle = iros,
  pages     = {7512--7519},
  year      = {2021},
  doi       = {10.1109/IROS51168.2021.9635857}
}

@inproceedings{giernacki2017crazyflie,
  author    = {Giernacki, Wojciech and Skwierczy{\'n}ski, Mateusz and Witwicki, Wojciech and Wro{\'n}ski, Pawe{\l} and Kozierski, Piotr},
  title     = {Crazyflie 2.0 quadrotor as a platform for research and education in robotics and control engineering},
  booktitle = {Intl. Conf. on Methods and Models in Automation and Robotics (MMAR)},
  pages     = {37--42},
  year      = {2017},
  doi       = {10.1109/MMAR.2017.8046794}
}

@inproceedings{preiss2017crazyswarm,
  author    = {Preiss, James A. and H{\"o}nig, Wolfgang and Sukhatme, Gaurav S. and Ayanian, Nora},
  title     = {Crazyswarm: A large nano-quadcopter swarm},
  booktitle = icra,
  pages     = {3299--3304},
  year      = {2017},
  doi       = {10.1109/ICRA.2017.7989376}
}

@incollection{deluca1998feedback,
  author    = {De Luca, Alessandro and Oriolo, Giuseppe and Samson, Claude},
  title     = {Feedback control of a nonholonomic car-like robot},
  booktitle = {Robot Motion Planning and Control},
  editor    = {Laumond, Jean-Paul},
  series    = {Lecture Notes in Control and Information Sciences},
  volume    = {229},
  publisher = springer,
  pages     = {171--253},
  year      = {1998},
  doi       = {10.1007/BFb0036073}
}

@inproceedings{kottinger2022conflict,
  author    = {Kottinger, Justin and Almagor, Shaull and Lahijanian, Morteza},
  title     = {Conflict-Based Search for Multi-Robot Motion Planning with Kinodynamic Constraints},
  booktitle = iros,
  pages     = {13494--13499},
  year      = {2022},
  doi       = {10.1109/IROS47612.2022.9982018},
  note      = {arXiv:2207.00576}
}

@inproceedings{le2017cooperative,
  author    = {Le, Duong and Plaku, Erion},
  title     = {Cooperative Multi-Robot Sampling-Based Motion Planning with Dynamics},
  booktitle = icaps,
  volume    = {27},
  pages     = {513--521},
  year      = {2017},
  doi       = {10.1609/icaps.v27i1.13854}
}

@article{li2016asymptotically,
  author  = {Li, Yanbo and Littlefield, Zakary and Bekris, Kostas E.},
  title   = {Asymptotically Optimal Sampling-Based Kinodynamic Planning},
  journal = ijrr,
  volume  = {35},
  number  = {5},
  pages   = {528--564},
  year    = {2016}
}

@article{frazzoli2002real,
  author  = {Frazzoli, Emilio and Dahleh, Munther A. and Feron, Eric},
  title   = {Real-time motion planning for agile autonomous vehicles},
  journal = {Journal of Guidance, Control, and Dynamics},
  volume  = {25},
  number  = {1},
  pages   = {116--129},
  year    = {2002},
  doi     = {10.2514/2.4856}
}

@inproceedings{kunz2010real,
  author    = {Kunz, Tobias and Reiser, Ulrich and Stilman, Mike and Verl, Alexander},
  title     = {Real-time path planning for a robot arm in changing environments},
  booktitle = iros,
  pages     = {5906--5911},
  year      = {2010},
  doi       = {10.1109/IROS.2010.5653275}
}

@article{hauser2012responsiveness,
  author  = {Hauser, Kris},
  title   = {On responsiveness, safety, and completeness in real-time motion planning},
  journal = ar,
  volume  = {32},
  number  = {1},
  pages   = {35--48},
  year    = {2012},
  doi     = {10.1007/s10514-011-9254-z}
}

@inproceedings{werling2010optimal,
  author    = {Werling, Moritz and Ziegler, Julius and Kammel, S{\"o}ren and Thrun, Sebastian},
  title     = {Optimal trajectory generation for dynamic street scenarios in a {F}ren{\'e}t frame},
  booktitle = icra,
  pages     = {987--993},
  year      = {2010},
  doi       = {10.1109/ROBOT.2010.5509799}
}

@article{mohta2018fast,
  author  = {Mohta, Kartik and Watterson, Michael and Mulgaonkar, Yash and Liu, Sikang and Qu, Chao and Makineni, Anurag and Saulnier, Kelsey and Sun, Ke and Zhu, Alex and Delmerico, Jeffrey and Karydis, Konstantinos and Atanasov, Nikolay and Loianno, Giuseppe and Scaramuzza, Davide and Daniilidis, Kostas and Taylor, Camillo Jose and Kumar, Vijay},
  title   = {Fast, autonomous flight in {GPS}-denied and cluttered environments},
  journal = jfr,
  volume  = {35},
  number  = {1},
  pages   = {101--120},
  year    = {2018},
  doi     = {10.1002/rob.21774}
}

@inproceedings{petti2005safe,
  author    = {Petti, St{\'e}phane and Fraichard, Thierry},
  title     = {Safe motion planning in dynamic environments},
  booktitle = iros,
  pages     = {2210--2215},
  year      = {2005}
}

@article{leven2002framework,
  author  = {Leven, Peter and Hutchinson, Seth},
  title   = {A framework for real-time path planning in changing environments},
  journal = ijrr,
  volume  = {21},
  number  = {12},
  pages   = {999--1030},
  year    = {2002},
  doi     = {10.1177/0278364902021012001}
}

@article{chen2015cooperative,
  title={Cooperative intersection management: A survey},
  author={Chen, Lei and Englund, Cristofer},
  journal=tits,
  volume={17},
  number={2},
  pages={570--586},
  year={2016},
  publisher={IEEE}
}

@article{moldagalieva2025dbecbs,
  title={db-ECBS: Interaction-Aware Multirobot Kinodynamic Motion Planning},
  author={Moldagalieva, Akmaral and Ortiz-Haro, Joaquim and H{\"o}nig, Wolfgang},
  journal=tro,
  volume={42},
  pages={244--260},
  year={2025},
  publisher={IEEE}
}

\clearpage

\appendices

\section{Proofs}\label{app:proofs}

This appendix provides full proofs of the results sketched in the main text, together with the auxiliary lemmas they use. The subsections follow the order of the paper.

\subsection{Regularity of the Reconstruction}\label{app:proof-filter}

\begin{proof}[Proof of \cref{lem:filter}]
The weak derivative of $\tx$ is $\phi*\dot{\hx}$, a convex average of velocities that lie in the Euclidean ball of radius $v$. A convex average stays in the ball, which proves the Euclidean speed bound. For a plan satisfying \cref{eq:transition}, the raw velocity is almost everywhere an axis pulse of magnitude $v$ or zero, so it lies in the Manhattan ball of radius $v$ and so does its convex average. Translations of $\phi$ are continuous in $L^1$ and $\dot{\hx}$ is bounded, so $\phi*\dot{\hx}$ is continuous, which gives $\tx\in C^1$ and makes the speed bounds hold at every time.

For the box kernel,
\begin{equation*}
 \dot{\tx}(t)=\frac{\hx(t+\Tf/2)-\hx(t-\Tf/2)}{\Tf},
\end{equation*}
so the Lipschitz bound on $\hx$ gives
\begin{equation*}
 \norm{\dot{\tx}(t)-\dot{\tx}(s)}_2\le\frac{2v}{\Tf}\,|t-s| .
\end{equation*}
The reconstructed velocity is therefore Lipschitz and hence absolutely continuous, and its almost-everywhere derivative has norm at most $2v/\Tf$. When $\Tf\ge T_{\mathrm{acc}}$ the acceleration bound $c_\phi v/\Tf=2v/\Tf$ is at most $a_{\max}$, which together with the $C^1$ regularity, the absolute continuity of the velocity, and the speed bound gives admissibility.
\end{proof}

\subsection{Diamond Geometry}\label{app:proof-geometry}

\Cref{lem:integer-gap} supplies the minimum used in \cref{thm:diamond}. \Cref{lem:footprint-exists} shows that the footprint selection of \cref{def:footprint-selection} is well defined and is used again in Appendix~\ref{app:proof-resolution}.

\begin{proof}[Proof of \cref{lem:sidestep}]
The offset after the tick is $(q'-p')/\eps=2m\,e_a\pm e_b$, whose Manhattan norm is $2m+1$, so the planning diamonds at $p'$ and $q'$ are disjoint by \cref{eq:diamond-gap-integer}. During the stop-and-go motion within the tick, both robots move at speed $v$ along their axes, so the relative offset in lattice units is $(2m+1-\alpha)e_a\pm\alpha e_b$ for $0\le\alpha\le1$. Its Manhattan norm is $2m+1$ for every $\alpha$, so the planning diamonds remain disjoint throughout the tick.
\end{proof}

\begin{lemma}[Minimum norm of an integer offset]\label{lem:integer-gap}
For an integer $s\ge0$ with $s=dq+j$ and $0\le j<d$, the minimum $M_d(s)$ in \cref{eq:integer-minimum} equals $(d-j)q^2+j(q+1)^2=(s^2+j(d-j))/d$, and it is attained exactly by the offsets whose coordinate magnitudes differ by at most one and sum to $s$.
\end{lemma}
\begin{proof}[Proof of \cref{lem:integer-gap}]
If the sum of the coordinate magnitudes exceeds $s$, decreasing a positive magnitude by one reduces the squared norm and keeps the sum at least $s$, so a minimizer has sum exactly $s$. At sum $s$, two magnitudes $u\ge w+2$ can be replaced by $u-1$ and $w+1$, which decreases the squared norm by $2(u-w-1)>0$. A minimizer therefore has magnitudes that differ by at most one. Exactly $j$ of them equal $q+1$ and the others equal $q$, which gives $(d-j)q^2+j(q+1)^2$, and substituting $q=(s-j)/d$ gives $(s^2+j(d-j))/d$. Conversely, every offset with magnitudes that differ by at most one and sum to $s$ attains this value.
\end{proof}

\begin{proof}[Proof of \cref{thm:diamond}]
A permitted pair has $\norm{a}_1\ge2m+1$ by \cref{eq:diamond-gap-integer}, so \cref{lem:integer-gap} gives \cref{eq:diamond-gap}, and the balanced offsets of the lemma attain the bound. The implication therefore holds for every pair exactly when
\begin{equation*}
 \eps^2M_d(2m+1)>4\reps^2 .
\end{equation*}
The alignment $\eps=\sqrt d\,r/n$ and \cref{eq:inflation,eq:moments} give $(2\reps)^2/\eps^2=4n^2/d+1+4\kappa_\phi W^2$, so the excess of the smallest permitted squared offset over this threshold is
\begin{equation}\label{eq:diamond-margin}
 \begin{aligned}
 &M_d(2m+1)-\Bigl(\frac{4n^2}d+1+4\kappa_\phi W^2\Bigr)\\
 &\qquad=\frac{4[m(m+1)-n^2]+1-d+j(d-j)}d-4\kappa_\phi W^2,
 \end{aligned}
\end{equation}
which is positive exactly under \cref{eq:diamond-gate}.
\end{proof}

\begin{lemma}[Existence and monotonicity of the footprint integer]\label{lem:footprint-exists}
For every integer $n\ge1$ and every positive even $W$, the set in \cref{eq:footprint-selection} is nonempty, so $m_*(n,W)$ exists, and every integer $m\ge m_*(n,W)$ belongs to the set. Moreover, $m_*(n,W)$ does not decrease as $W$ grows.
\end{lemma}
\begin{proof}
By \cref{eq:diamond-margin}, an integer $m\ge n$ belongs to the set exactly when \cref{eq:diamond-gate} holds. From $m$ to $m+1$ the term $4m(m+1)$ on the left side of \cref{eq:diamond-gate} grows by $8(m+1)$, while $j(d-j)$ stays between $0$ and $2$, so the left side increases strictly and without bound with $m$. The set is therefore nonempty and contains every integer above its least element, and the search that tests \cref{eq:diamond-gate} at $m=n,n+1,\ldots$ stops at $m_*(n,W)$. The right side of \cref{eq:diamond-gate} increases with $W$, so $m_*(n,W)$ does not decrease as $W$ grows.
\end{proof}

\subsection{Successor Generation and Configuration Search}\label{app:proof-solver}

The proof of \cref{lem:tick-validity} refers to the response search of Appendix~\ref{app:response} and to the direct commitment of Appendix~\ref{app:neighbors}. \Cref{lem:neighbor-cap} proves the bound on the neighbor lists used in \cref{eq:tick-cost}.

\begin{proof}[Proof of \cref{lem:tick-validity}]
Every candidate attempt is charged to the budget, the budget bounds the attempts of the search for one root, and at most $N$ searches are opened, so the algorithm terminates.

We claim that, throughout the call, every undecided robot can wait without conflicting with any committed target. The claim holds before the first search of a tick, because without fixed targets the only committed targets are the direct commitments of Appendix~\ref{app:neighbors}, each of which is separated from every current position and from every possible target of every other robot. Suppose the claim holds when the search for a root begins. If the search succeeds, every undecided robot whose position conflicted with a target assigned in it was placed on the agenda of Appendix~\ref{app:response} and decided before \textsc{Respond} returned success, and every robot assigned its own position is separated from every undecided position because $Q$ is valid. If the search is abandoned, only the root is committed, at its own position, which is separated from every undecided position because $Q$ is valid. In both cases the claim holds when the next search begins.

Every candidate lies in $V_D$ and differs from the position of its robot by an element of $\moveset$. A tentative target is accepted only when it is separated from every decided target, so the committed targets are pairwise separated at all times. By the claim, the wait of a root is separated from every target committed before its search, it has no blocking robots because $Q$ is valid, and every target committed after it is checked against it. The wait therefore succeeds whenever the root reaches it, so a search can fail only through the budget, and the root is then committed at its position, which is legal by the claim. Every robot is decided when the tick ends, and $Q'$ is valid.
\end{proof}

\begin{proof}[Proof of \cref{prop:search-completeness}]
There are at most $|V_D|^N$ configurations and at most $(2d+1)^N$ joint moves from each. Removing the invalid configurations and transitions leaves a finite graph, and the search, which removes a node only after its constraint tree is exhausted~\cite{okumura2023lacam}, visits every reachable vertex of it. A constraint tree that fixes one more move at each level enumerates the same finite move set at full depth. A failed generator call under a partial constraint does not affect the argument, because the fully constrained leaves are still tested exactly, and heuristic successors generated at shallower levels are added earlier and are not required.
\end{proof}

\begin{lemma}[Exactness of the retained search]\label{lem:retained-search}
Every completed query for robot $i$ returns the exact goal distance $D_i$ at the queried vertex, independently of the order and timing of the queries.
\end{lemma}
\begin{proof}[Proof of \cref{lem:retained-search}]
The heuristic is consistent, because $|h_{x_0}(s)-h_{x_0}(s')|\le1$ for axial neighbors $s$ and $s'$ and every edge costs one, so $f$ does not decrease along a shortest path from $g$. The two queues pop entries in nondecreasing $f$. Suppose some vertex were settled above its true distance, and let $s$ be the first such vertex in pop order. Let $u$ be the first vertex on a shortest path from $g$ to $s$ that was not settled before $s$. The predecessor of $u$ on that path was settled with its exact distance, so its expansion stored $\tilde d_g(u)=D(u)$ and queued $u$ at $f(u)=D(u)+h_{x_0}(u)\le D(s)+h_{x_0}(s)<\tilde d_g(s)+h_{x_0}(s)$. Every entry below the $f$-value of $s$ was popped before $s$, so $u$ was settled before $s$, which contradicts its choice. Suspending the search after a settled and expanded query and resuming it later leaves the queues and the stored values unchanged, so the popped vertices form the sequence of one uninterrupted search with the fixed target $x_0$.
\end{proof}

\begin{proof}[Proof of \cref{lem:push-cap}]
For Euclidean conflicts, the positions in conflict with a target $t$ lie within $2\reps$ of $t$ and are pairwise more than $2\reps$ apart. The open balls of radius $\reps$ around them are disjoint and lie in the ball of radius $3\reps$ around $t$, whose volume is $3^d$ times theirs. For diamond conflicts, the same argument applies to the open Manhattan balls of radius $h$ around the positions, which are disjoint and lie in the Manhattan ball of radius $3h$ around $t$.
\end{proof}

The packing argument uses the axial radius $h=m_*\eps$ and remains valid when $m_*>n$, because a larger footprint also enlarges the separation of the positions.

\begin{lemma}[Bounded neighbor lists]\label{lem:neighbor-cap}
In a configuration of $\mathcal Q_\eps$, at most $7^d$ robots lie within Euclidean distance $2\reps+2\eps$ of any point. In a configuration of $\mathcal Q_D$ with $h=m_*\eps$, at most $(4\sqrt d+1)^d$ robots lie within Euclidean distance $2h+2\eps$ of any point. The neighbor lists of Appendix~\ref{app:neighbors} therefore have length at most a constant $\gamma$ that depends only on $d$.
\end{lemma}
\begin{proof}
In $\mathcal Q_\eps$, the robots within $2\reps+2\eps$ of a point $z$ are pairwise more than $2\reps$ apart, so the open balls of radius $\reps$ around them are disjoint. Since $\eps\le2\reps$ by \cref{eq:inflation}, these balls lie in the ball of radius $3\reps+2\eps\le7\reps$ around $z$, and comparing volumes bounds their number by $7^d$. In $\mathcal Q_D$, validity gives pairwise Manhattan separation greater than $2h$, so the open Manhattan balls of radius $h$ around the robots within $2h+2\eps$ of $z$ are disjoint. Since $\eps\le h$, the Euclidean ball of radius $2h+2\eps\le4h$ around $z$ lies in the Manhattan ball of radius $4\sqrt d\,h$, so these disjoint Manhattan balls lie in the Manhattan ball of radius $(4\sqrt d+1)h$ around $z$, and comparing volumes bounds their number by $(4\sqrt d+1)^d$.
\end{proof}

\subsection{Existence and Resolution}\label{app:proof-resolution}

The proof of \cref{prop:limits} uses \cref{lem:footprint-exists}, and the proof of \cref{cor:diamond-resolution} reuses the plan constructed in the proof of \cref{thm:completeness}.

\begin{proof}[Proof of \cref{prop:limits}]
By \cref{def:filter-duration}, $T_{\mathrm{acc}}\le\Tf<T_{\mathrm{acc}}+2\dt$ with $\dt=\eps/v$, so $\Tf\to T_{\mathrm{acc}}$ as $\eps\to0$. In \cref{eq:inflation} the tick reserve $\eps^2/4$ vanishes and $\kappa_\phi v^2\Tf^2\to\kappa_\phi v^2T_{\mathrm{acc}}^2=\kappa_\phi c_\phi^2v^4/a_{\max}^2$, so $\reps\to r_{\mathrm{acc}}$ by \cref{eq:r-acc}.

For the diamond limit \cref{eq:footprint-limit}, write $A=n^2+d\kappa_\phi W^2$ with $W=W(n)$. By \cref{lem:footprint-exists}, $m_*$ exists. The term $1-d+j(d-j)$ in \cref{eq:diamond-gate} equals $0$ or $-2$ for $d\in\{2,3\}$, so the condition implies $m(m+1)>A$ and is implied by $m(m+1)>A+1/2$. The least integer $m_*$ therefore satisfies
\begin{equation*}
 \sqrt A-\tfrac12<m_*\le\sqrt A+1,
\end{equation*}
where the upper bound uses $n\le\sqrt A$ when $m_*=n$ and the failure of the condition at $m_*-1$ otherwise. Multiplying by $\eps=\sqrt d\,r/n$ and using $W\eps=v\Tf$ gives
\begin{equation*}
 m_*\eps=\sqrt d\,\bigl(r^2+\kappa_\phi v^2\Tf^2\bigr)^{1/2}+O(\eps),
\end{equation*}
which tends to $\sqrt d\,r_{\mathrm{acc}}$ because $\Tf\to T_{\mathrm{acc}}$ by \cref{def:filter-duration}. Dividing by $n\eps=\sqrt d\,r$ gives the ratio in \cref{eq:footprint-limit}.
\end{proof}

\begin{proof}[Proof of \cref{cor:diamond-resolution}]
Set $h=m_*(n)\eps(n)$. The parameter limits give $h\to\sqrt d\,r_{\mathrm{acc}}$ and $\reps\to r_{\mathrm{acc}}$, so $h>\reps$ at sufficiently fine resolution. A planning diamond lies inside the Euclidean ball of radius $h$ about its center. Hence a lattice vertex with obstacle clearance greater than $h$ is admitted, and two centers separated by more than $2h$ have disjoint diamonds. The tracker used for \cref{thm:completeness} has clearance at least $r+\mu_X-d\eps$ and pairwise separation at least $2r+\mu_X-2d\eps$. Both exceed the required diamond thresholds when
\begin{equation*}
 2d\eps+2(h-r)<\mu_X.
\end{equation*}
The stated margin makes this inequality true for all sufficiently large $n$. Thus the tracked plan lies in $\mathcal Q_D$. The selected width gives $\Tf\ge T_{\mathrm{acc}}$, so \cref{thm:soundness} makes its reconstruction a solution of \cref{prob:mampp}.
\end{proof}

\section{Details of the Pipeline Modules}\label{app:modules}

This appendix specifies the parts of the modules that \cref{alg:pipeline} and \cref{sec:pwm,sec:diamond,sec:discrete} leave out. Each subsection covers one module, in the order of \cref{alg:pipeline}.

\subsection{Reconstruction Kernels}\label{app:kernels}

The kernel $\phi$ of \cref{sec:discretization} is symmetric and nonnegative, has unit integral, and is supported on $[-\Tf/2,\Tf/2]$. The box kernel is used throughout the paper. This subsection gives the constants of the triangular kernel, a second member of the class, whose trajectories have continuous acceleration.

\paragraph*{Triangular kernel}
The triangular kernel is $B_{\Tf/2}*B_{\Tf/2}$, with $(c_\phi,\kappa_\phi)=(4,1/24)$. With $F'=\hx$, $G'=F$, and $c=\Tf/2$, its reconstruction is
\begin{equation}\label{eq:closed-form-tri}
 \tx(t)=\frac{G(t+c)-2G(t)+G(t-c)}{c^2},
\end{equation}
which consists of cubic pieces. Two box averages give
\begin{equation*}
 \ddot{\tx}(t)=\frac{\hx(t+c)-2\hx(t)+\hx(t-c)}{c^2},
\end{equation*}
which is continuous and has norm at most $2vc/c^2=4v/\Tf$, so $\tx\in C^2$. Its almost-everywhere derivative has norm at most $4v/c^2=16v/\Tf^2$. \Cref{tab:kernels} lists the constants of both kernels. Under the same speed and acceleration caps, the triangular kernel has twice the target duration $T_{\mathrm{acc}}$ and twice the limiting squared smoothing reserve $\kappa_\phi v^2T_{\mathrm{acc}}^2$, since $\kappa_\phi c_\phi^2$ equals $1/3$ for the box kernel and $2/3$ for the triangular kernel. At the same filter duration, the triangular kernel has half the variance. The certificate condition \cref{eq:diamond-gate} for the triangular kernel reads $12[m(m+1)-n^2]>W^2$ for $d=2$ and $2\bigl(4[m(m+1)-n^2]-2+j(3-j)\bigr)>W^2$ for $d=3$.

\begin{table}[t]
\centering\small
\caption{The box and triangular kernels with filter duration $\Tf$. The polynomial degrees refer to stop-and-go trajectories. The derivative bounds hold for any $v$-Lipschitz path.}\label{tab:kernels}
\setlength{\tabcolsep}{4pt}\begin{tabular}{@{}lll@{}}
\toprule
 & Box & Triangular\\\midrule
$\phi(s)$ on its support & $1/\Tf$ & $\frac{2}{\Tf}(1-\frac{2|s|}{\Tf})$\\
$\kappa_\phi=\sigk^2/\Tf^2$ & $1/12$ & $1/24$\\
Trajectory & $C^1$, quadratic pieces & $C^2$, cubic pieces\\
Acceleration bound & $2v/\Tf$ a.e. & $4v/\Tf$\\
Acceleration constant $c_\phi$ & $2$ & $4$\\
Jerk bound & --- & $16v/\Tf^2$ a.e.\\\bottomrule
\end{tabular}
\end{table}

\subsection{Vertex Admission and Endpoint Anchors}\label{app:admission}

The vertex set $V_D$ of \cref{eq:diamond-sites} tests both the radial condition $\clr(p)>\reps$ and the footprint condition $\diamondbody_m(p)\cap\forbidden=\varnothing$ at every vertex, because a free planning diamond does not imply $\clr(p)>\reps$.
The inradius of the planning diamond $\diamondbody_m(p)$ is $h/\sqrt d$, so a ball of radius $\varrho$ around $p$ lies in the footprint exactly when $\varrho\le h/\sqrt d$. With $h=m\eps$, $r=n\eps/\sqrt d$, and $v\Tf=W\eps$, the certificate ball of radius $\reps$ is contained in the planning diamond exactly when
\begin{equation*}
 \frac{m^2\eps^2}{d}\ge\reps^2=\eps^2\Bigl(\frac{n^2}{d}+\frac14+\kappa_\phi W^2\Bigr),
\end{equation*}
that is, when $m^2\ge n^2+d/4+d\kappa_\phi W^2$.
The pair certificate of \cref{thm:diamond} does not require this containment. In \cref{fig:clearance}(a), $m_*=2$ satisfies the certificate, and containment would require $m\ge3$. In \cref{fig:clearance}(b), containment holds. When containment fails, the point $z=p+\lambda(1,\ldots,1)/\sqrt d$ with $h/\sqrt d<\lambda<\reps$ lies outside $\diamondbody_m(p)$ at distance $\lambda<\reps$ from $p$, so an obstacle at $z$ leaves $\diamondbody_m(p)$ free and violates $\clr(p)>\reps$. The pair certificate compares two lattice centers, whereas an obstacle point need not be a lattice center, and the vertex set $V_D$ of \cref{eq:diamond-sites} therefore requires both conditions for every footprint integer.

Let $S=(s_1,\ldots,s_N)$ and $G=(g_1,\ldots,g_N)$ be the requested start and goal configurations and let $\eta\ge0$ be the tolerance. The start anchor configuration $S^\eps=(s_1^\eps,\ldots,s_N^\eps)$ and the goal anchor configuration $G^\eps=(g_1^\eps,\ldots,g_N^\eps)$ are lattice configurations in $\mathcal Q_D$ that satisfy
\begin{equation}\label{eq:anchor-tolerance}
 \norm{s_i^\eps-s_i}_2\le\eta,\qquad
 \norm{g_i^\eps-g_i}_2\le\eta
 \qquad\text{for every }i.
\end{equation}
The vertices $s_i^\eps$ and $g_i^\eps$ are the start anchor and the goal anchor of robot $i$.
The tolerance absorbs only the offset between a requested endpoint and the lattice. Every point lies within $\sqrt d\,\eps/2=dr/(2n)$ of a lattice vertex, so $\eta$ is of the order of one pitch, and it decreases as the resolution $n$ grows. In our experiments, $\eta$ is 9\,cm for robots of radius 50\,cm in workspaces of more than 100\,m across in 2D, and 5\,cm for robots of radius 20\,cm in 3D. The displacement of an endpoint is therefore negligible at the scale of the task, and the anchors coincide with the requested endpoints for every practical purpose. Since $\eta$ is part of the instance, the reconstruction of a plan between the anchors meets the endpoint conditions of \cref{prob:mampp} exactly.

The candidate anchors of an endpoint are the admitted vertices within the tolerance $\eta$ of it, listed by increasing displacement. Two endpoints in the same configuration interact when some pair of their candidates has intersecting planning diamonds. Each connected group of interacting endpoints is assigned jointly by backtracking over its candidate lists, so that the selected configuration lies in $\mathcal Q_D$.

\subsection{Response Search and Rollback}\label{app:response}

\textsc{Respond} in \cref{alg:generator} keeps an agenda of the robots that must respond to the tentative targets of the current branch.
The blocking robots of a target are placed at the front of the agenda in the priority order, each is resolved fully before the next, and a robot that another branch has decided in the meantime is skipped, with its target constraining every later choice. The candidate orders, the distances, and the priority order are frozen for the tick. A call of \textsc{Respond} that returns failure has undone every tentative assignment it made, so the branch of its caller is restored to the state before that call, and the caller tries its next candidate. When the budget is exhausted, the interruption propagates to the root without local cleanup, and the root then discards every tentative assignment of its search and marks those robots undecided. The same budget counts the attempts of all branches of the search for one root, so trying another root target does not refresh the budget.

\subsection{Neighbor Queries and Direct Commitment}\label{app:neighbors}

One call of \cref{alg:generator} tests every candidate target against the positions and the targets of the other robots, and testing all pairs of robots would cost $O(N^2)$ per tick. Conflicts are local, because a target lies within one pitch of its robot, so only robots within a fixed distance of each other can conflict. The generator uses this locality in two ways. Neighbor queries restrict every conflict test to a list of nearby robots, whose length is bounded by a constant (\cref{lem:neighbor-cap}). Direct commitment assigns robots that have no other robot near their preferred move before any search starts, so the response search of Appendix~\ref{app:response} runs only where robots interact. Together they give the cost \cref{eq:tick-cost} of one generator call, which is linear in $N$ apart from the sort of the priorities.

Two vertices are in diamond conflict when their planning diamonds intersect, as defined in \cref{prob:discrete}. A pair in diamond conflict has $\norm{p-q}_1\le2h$ by \cref{eq:diamond-policy}, and $\norm{p-q}_2\le\norm{p-q}_1$, so the pair lies within the Euclidean conflict radius $R=2h=2m_*\eps$. Under the Euclidean rule of the ablation in \cref{sec:exp-ablation}, the conflict radius is $R=2\reps$.

A target lies within $\eps$ of its robot. The targets of two robots can therefore conflict only when the robots are within $R+2\eps$ of each other, and the current position or a target of another robot can conflict with a candidate $t$ only when that robot lies within $R+\eps$ of $t$. The neighbor list of a robot holds the robots within $R+2\eps$ of its position, and the blocking robots of a target are found in the neighbor list of the robot that proposes it. A spatial hash with cells of side $R+4\eps$ is rebuilt at every tick, and a query for the robots within a radius scans the $3^d$ cells around the query point.

Before the searches of a tick, a robot with no other robot within $R+4\eps$ of its first candidate is committed to that candidate directly. The query radius covers $R+\eps$, so the committed target is separated from the current position and from every possible target of every other robot. Fixed targets are committed before this step and remain fixed throughout the call, so the direct commitment considers undecided robots only.

\subsection{Priority Updates, Tie Orders, and \acrshort{acr:takt}-PIBT}\label{app:scheduling}

\paragraph*{Initialization and node state}
Each search node retains a priority $\pi_i$, a rank $\beta_i$, a wait counter $\omega_i$, and a candidate tie order for every robot. The scheduling key is $(\pi_i,\beta_i)$. At the initial node, $\pi_i=D_i(s_i^\eps)/N$, $\beta_i=i$, and $\omega_i=0$. The tie order of a robot is a permutation of $\moveset$ that breaks the ties left by the candidate order of \cref{sec:ordering}.

\paragraph*{Updates after a joint move}
On the edge from a node to a child, $\pi_i$ increases by one when robot $i$ ends away from its goal and is replaced by its fractional part when it ends at the goal. The counter $\omega_i$ is set to zero when robot $i$ moved on the edge or ends at its goal and increases by one otherwise. The updated triples are the state of the child before rotation.

\paragraph*{Stationary-group rotation}
At the child configuration, robot $i$ is stationary when it is away from its goal and $\omega_i\ge F$, where
\begin{equation}\label{eq:freeze-window}
 F=\max\Bigl\{1,\ \Bigl\lfloor\frac{3\tau_I}{\dt}\Bigr\rceil\Bigr\}
  =\max\Bigl\{1,\ \Bigl\lfloor\frac{6r}{\eps}\Bigr\rceil\Bigr\}
\end{equation}
is the number of ticks in three interaction times $\tau_I=2r/v$, rounded down to the nearest integer. Each stationary group of \cref{sec:ordering}, with its members $a_0,\ldots,a_{\lambda-1}$ listed in the current priority order, passes the key of $a_k$ to $a_{k+1}$ and the key of $a_{\lambda-1}$ to $a_0$, and the wait counters of the rotated robots are set to zero. 
The child’s priority order places robots away from their goals first, breaking ties by decreasing $\pi_i$ and then by decreasing $\beta_i$.
The ranks remain a permutation, so the order is total, and it is frozen for the node. Repeated rotations of an unchanged group give every member the leading key once in $\lambda$ rotations, so the group obtains a different initiating robot.

\paragraph*{Tie orders and revisits}
A new node receives seeded tie orders, which every constrained generator call at that node uses. A configuration reached again keeps the node and the state of its first visit, so repeated calls and revisits at one configuration use consistent choices.

\paragraph*{\acrshort{acr:takt}-PIBT}
The generator used tick by tick without the configuration search differs in three respects. A scheduling age replaces the priority. It starts at zero, increases by one in every tick that the robot ends away from its goal, and is reset to zero when the robot ends at its goal. The rank and the wait counter are updated by the same rules after every tick, and the stationary groups of the current configuration are rotated at the start of the next tick. The tie order of a robot is redrawn every $\tau$ ticks, where $\tau$ is one interaction time $\tau_I$ rounded to a whole number of ticks, so the ranking of tied moves is constant within a window, and all attempts within a single tick share the same tie order.

\subsection{Retained Backward A* Implementation}\label{app:distance}

The search of a goal $g$ stores the tentative distance $\tilde d_g(s)$ of every discovered vertex and expands vertices in the order of $f$ in \cref{eq:retained-f}. An axial move changes $h_{x_0}$ by exactly one and costs one, so the $f$-value of a discovered neighbor equals the $f$-value of its parent or exceeds it by two. The open list therefore consists of two first-in-first-out queues, one for the $f$-value being expanded and one for that value plus two. A popped vertex is settled when $\tilde d_g(s)+h_{x_0}(s)$ equals the $f$-value of its queue, and a stale entry is skipped. A query resumes popping vertices until the queried vertex has been settled and expanded, and the remaining entries of both queues are retained as the frontier. The first query location $x_0$ remains fixed for the life of the search, so later queries change only where the search pauses.

The tentative distances and settled bits of a search are stored sparsely, in pages allocated when a vertex of the page is first discovered, so the storage of a search grows with the region it has explored. When the admitted vertices form one rectangle, $D_i(q)$ equals the Manhattan distance $\norm{q-g_i^\eps}_1/\eps$, because a monotone lattice path between two vertices of a rectangle stays inside the rectangle, and this formula replaces the search.

\section{Experiment Details}\label{app:experiments}

This appendix gives the details of the experiments of \cref{sec:experiments}. Its subsections cover the simulated platforms and their controllers, the execution of the two replanning paradigms of \cref{sec:online}, and further measurements of the real-time sessions.

\subsection{Simulated Platforms and Tracking Controllers}\label{app:platforms}

\myparagraph{Differential-Drive Base} --
A differential-drive robot moves along its heading on two driven wheels. Each base tracks its trajectory with a point at a distance $b>0$ ahead of the wheel axle. The velocity of this point is an invertible function of the forward speed and the turn rate, so a proportional feedback on the point, with the planned velocity as feedforward, drives it onto any planned trajectory at an exponential rate~\cite{deluca1998feedback}. The feedback stays defined at rest, which suits planned motions that include waits. The axle midpoint lies at the distance $b$ behind the tracked point, so a robot stays inside the disc of radius $r$ around its planned position whenever $b$, the tracking error, and the radius of the robot body sum to less than $r$.

The simulator of the Robotnik RB-1 Base takes the mass of 27.25\,kg, the yaw inertia of 1.10\,kg\,m$^2$, the wheel radius of 7.62\,cm, and the track of 43.6\,cm from the manufacturer's robot description. It limits each wheel to 27.27\,rad/s and each motor to 9.17\,N\,m and 250\,W, assumes that the wheels do not slip, and integrates the fleet every 1\,ms. Every robot tracks its trajectory with the point $b=20$\,cm ahead of its axle, a feedback gain of 8\,s$^{-1}$, and the planned velocity read 0.015\,s ahead. Inverting the kinematics of that point gives the forward speed and the turn rate, and hence the two wheel-speed setpoints. A proportional-integral loop on each wheel, with gains of 5\,N\,m\,s/rad and 50\,N\,m/rad, sets the motor torque. Each robot starts at rest with the tracked point on its start position and its heading along its first move.

\myparagraph{Quadrotor} --
Each simulated Crazyflie~2.x is a rigid body driven by four rotors with the parameters of gym-pybullet-drones~\cite{panerati2021learning}: a mass of 27\,g, a thrust-to-weight ratio of 2.25, and up to 0.149\,N per rotor. The simulator integrates the fleet every 1\,ms. Every quadrotor tracks its trajectory with a cascaded controller based on the PX4 autopilot~\cite{meier2015px4}. Position and velocity loops with gains of 6\,s$^{-1}$ and 12\,s$^{-1}$ correct the planned velocity and acceleration, which enter as feedforward~\cite{mellinger2011minimum}. The commanded acceleration and gravity define the thrust vector, with the tilt limited to 0.5\,rad, and an attitude loop with gains of 1600\,s$^{-2}$ and 80\,s$^{-1}$ aligns the quadrotor with this thrust vector at a fixed heading.
\Cref{def:smoothing} gives the planned velocity and acceleration in closed form, so the controller reads the planned acceleration 0.06\,s ahead and each quadrotor tilts in time for every acceleration change. Fixed-pitch rotors bound the downward acceleration of a quadrotor by gravity, which is the reason for the cap of 5\,m/s$^2$ in the 3D instances.

\myparagraph{Fleets} --
\Cref{tab:exp-tracking} reports the largest errors and the smallest gaps over twelve fleets, six per platform, with the simulated states sampled every 10\,ms. The tracking is uniform across the fleets. The 99th-percentile error lies between 0.48 and 0.53\,cm for the bases and between 1.0 and 1.4\,cm for the quadrotors.

\subsection{Execution of the Replanning Paradigms}\label{app:events}

This subsection specifies how the sessions of \cref{sec:exp-online} execute the two paradigms of \cref{sec:online}.

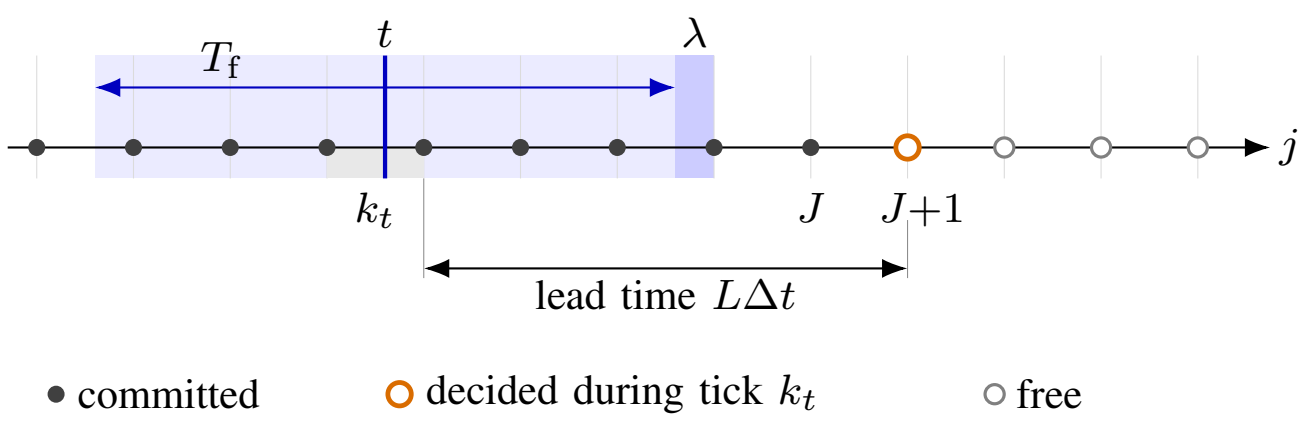
\begin{figure}[t]
\centering
\begin{tikzpicture}[>=Latex,font=\footnotesize,x=0.64cm,y=0.64cm,
 done/.style={circle,fill=black!75,inner sep=0pt,minimum size=3.2pt},
 next/.style={circle,draw=orange!85!black,line width=0.9pt,fill=white,inner sep=0pt,minimum size=4.4pt},
 free/.style={circle,draw=black!50,line width=0.6pt,fill=white,inner sep=0pt,minimum size=3.6pt}]
 \fill[blue!8] (0.6,-0.32) rectangle (6.6,0.95);
 \fill[blue!20] (6.6,-0.32) rectangle (7.0,0.95);
 \foreach \t in {0,...,12} { \draw[black!14,very thin] (\t,-0.32)--(\t,0.95); }
 \fill[black!9] (3,-0.32) rectangle (4,0);
 \draw[->] (-0.3,0)--(12.75,0) node[right,inner sep=1.5pt] {$j$};
 \draw[blue!75!black,thick] (3.6,-0.32)--(3.6,0.95);
 \node[above,inner sep=1.5pt,text=black] at (3.6,0.95) {$t$};
 \draw[<->,blue!75!black] (0.6,0.62)--node[above,inner sep=1.5pt,text=black,pos=0.22] {$\Tf$}(6.6,0.62);
 \node[above,inner sep=1.5pt] at (6.8,0.95) {$\lambda$};
 \foreach \t in {0,...,8} { \node[done] at (\t,0) {}; }
 \node[next] at (9,0) {};
 \foreach \t in {10,11,12} { \node[free] at (\t,0) {}; }
 \node[below,inner sep=2.5pt] at (3.5,-0.32) {$k_t$};
 \node[below,inner sep=2.5pt] at (8,-0.32) {$J$};
 \node[below,inner sep=2.5pt] at (9.15,-0.32) {$J{+}1$};
 \draw[<->] (4,-1.25)--node[below,inner sep=2pt] {lead time $L\dt$}(9,-1.25);
 \draw[black!45,very thin] (4,-0.32)--(4,-1.35);
 \draw[black!45,very thin] (9,-0.75)--(9,-1.35);
 \node[done] at (0.2,-2.55) {};
 \node[right,inner sep=3pt] at (0.25,-2.55) {committed};
 \node[next] at (3.75,-2.55) {};
 \node[right,inner sep=3pt] at (3.85,-2.55) {decided during tick $k_t$};
 \node[free] at (9.9,-2.55) {};
 \node[right,inner sep=3pt] at (9.95,-2.55) {free};
\end{tikzpicture}
\caption{Lead time of the incremental paradigm for $W=6$ and $\lceil\lambda/\dt\rceil=1$. The dots are the configurations of the padded sequence at their times $j\dt$. At the clock time $t$, inside tick $k_t$, the reconstruction averages the stop-and-go trajectory over the window of duration $\Tf$ around $t$ (light blue), and the controller reads it a look-ahead $\lambda$ ahead (dark blue). Until the tick ends, the robots need the configurations up to the frontier $J$. The planner decides the configuration $J+1$ during this tick, $L$ ticks before the stop-and-go trajectories reach it, and every later configuration is free.}
\label{fig:lead-time}
\end{figure}

\myparagraph{Lead Time} --
Let $P^0,P^1,\ldots$ be the padded sequence of configurations of \cref{def:smoothing}, which the stop-and-go trajectories $\hx_i$ pass at the times $j\dt$. By \cref{eq:dac}, the reconstruction at a time $t$ depends on $\hx_i$ over the window $[t-\Tf/2,t+\Tf/2]$ only. A tracking controller may also read its reference a look-ahead $\lambda\ge0$ ahead of the clock. During the tick $k_t=\lceil t/\dt\rceil$, the robots therefore need the sequence up to the index
\begin{equation}\label{eq:frontier}
 J=k_t+\frac W2+\Bigl\lceil\frac{\lambda}{\dt}\Bigr\rceil ,
\end{equation}
which we call the \emph{frontier} (\cref{fig:lead-time}). The configurations up to the frontier are committed, and every later configuration can still change. A planner that decides the configuration $P^{J+1}$ during this tick decides every configuration
\begin{equation}\label{eq:lead}
 L=\frac W2+1+\Bigl\lceil\frac{\lambda}{\dt}\Bigr\rceil
\end{equation}
ticks before the stop-and-go trajectories reach it. The duration $L\dt$ is the lead time~\cite{frazzoli2002real} of the incremental paradigm. It is the sum of half a filter duration, one tick of computation, and the look-ahead rounded up to whole ticks. By \cref{def:filter-duration}, half a filter duration lies within one tick of $c_\phi v/(2a_{\max})$, which for the box kernel is the time $v/a_{\max}$ in which a robot at speed $v$ stops under the acceleration cap. The lead time therefore follows from the dynamics of the robots, and it does not depend on the fleet size. The controllers of Appendix~\ref{app:platforms} read their references 0.015\,s and 0.06\,s ahead, so the lead time is $L=5$ ticks, or 0.51\,s, in 2D and $L=8$ ticks, or 0.35\,s, in 3D.

\myparagraph{Changes} --
A session starts from the anchor configuration $S^\eps$ and receives changes at times that it does not know in advance. A \emph{new goal} gives a robot $i$ a new goal anchor $g_i^\eps\in V_D$. A \emph{keep-out region} is a closed set $R\subset\mathcal W$ that appears at one time and is removed at a later time. While it exists, $R$ acts as an additional obstacle. We write $V_D(R)\subseteq V_D$ for the vertices that pass the two admission tests of Appendix~\ref{app:admission} when $R$ is added to the obstacles.

\myparagraph{Whole-Fleet Replanning} --
On a change, the session selects a \emph{splice index} $J_s$ ahead of the frontier. 
The planner then treats $P^{J_s}$ as a new start configuration and computes a replacement plan, with the new goals and on the vertices $V_D(R)$, and the result replaces the current plan from $J_s$ on. The time from the change to the splice is the lead time of the whole-fleet paradigm, and the planner must return within it. The session places $J_s$ at twice the computation time of the initial plan ahead of the frontier. The robots follow the previous plan until the splice.

\myparagraph{Incremental Replanning} --
The incremental paradigm runs one \emph{decision cycle} per tick, listed in \cref{alg:cycle}. A cycle reads the changes that arrived since the last cycle, calls \cref{alg:generator} once from the frontier configuration $P^J$, and commits the result as $P^{J+1}$. A change \emph{takes effect} when the stop-and-go trajectories reach the first configuration decided with it, which occurs between $(L+1)\dt$ and $(L+2)\dt$ after the change arrives. The planner must supply one configuration per tick. \Cref{lem:tick-validity} holds for every budget $B\ge1$, because a robot that is not yet decided can always wait. The cycle uses this property to keep its pace. It stops the root loop of \cref{alg:generator} at a wall-clock budget below the tick duration, and the robots that are not yet decided wait in that tick.

\begin{algorithm}[t]
\caption{Decision cycle of the incremental paradigm}\label{alg:cycle}
\small
\begin{algorithmic}[1]
\Require Committed sequence $P^{0:J}$; goals $G^\eps$; regions $\mathcal R$; halted robots $H$; priority order $\prec$; budget $B$; changes $E$ since the last cycle
\Ensure Committed configuration $P^{J+1}$
\ForAll{changes in $E$, in order of arrival}
 \If{new goal $g$ of robot $i$}
  \State $g_i^\eps\gets g$ and reset the scheduling state of robot $i$
 \ElsIf{region $R$ appears}
  \State $\mathcal R\gets\mathcal R\cup\{R\}$
  \State add to $H$ every robot $i$ with a vertex $p_i^j\notin V_D(R)$ for some $k_t-W/2\le j\le J$
 \ElsIf{region $R$ is removed}
  \State $\mathcal R\gets\mathcal R\setminus\{R\}$ and release the robots that $R$ halted
 \EndIf
\EndFor
\State restrict $C(i)$ to the wait for every $i\in H$
\State remove from $C(i)$, $i\notin H$, the moves that leave $V_D(R)$ for some $R\in\mathcal R$ or conflict with a robot of $H$
\State $P^{J+1}\gets{}$\Call{Generator}{$P^J,\prec,B,\varnothing$} \Comment{\cref{alg:generator}}
\State commit $P^{J+1}$ and update $\prec$ \Comment{Appendix~\ref{app:scheduling}}
\end{algorithmic}
\end{algorithm}

\myparagraph{New Goal} --
A new goal changes the candidate order of its robot (\cref{sec:ordering}) from the next cycle on. The scheduling state of the robot is reset as on an arrival at a goal (Appendix~\ref{app:scheduling}). The retained search of \cref{sec:distance} starts a backward A* from the new goal at the first distance query of the robot, so a new goal requires no distance field over the lattice.

\myparagraph{Keep-Out Region} --
When a region $R$ appears, the vertices outside $V_D(R)$ are removed from the lattice. The motion of every robot is committed up to the frontier. A robot is \emph{halted} when one of the vertices that still enter its reconstruction, from the index $k_t-W/2$ to the frontier, lies outside $V_D(R)$. A halted robot waits from the frontier on and holds its position until the region is removed, as a robot stopped for maintenance would. Every other robot loses the candidates of \cref{eq:candidates} that lie outside $V_D(R)$ or conflict with the held position of a halted robot, and it keeps the wait. The goal of a robot is \emph{suspended} when the region removes the goal or cuts it off. Such a robot is scheduled as a robot at its goal until the removal, so it waits and yields to the others. The goal distances of the remaining robots are searched on the reduced lattice. Under the whole-fleet paradigm, the halting rule applies to the vertices from the index $J_s-W/2$ to $J_s$.

\myparagraph{Certified Execution} --
The guarantee of \cref{cor:pipeline} extends to the motion that the robots execute.

\begin{proposition}[Certified execution under changes]\label{prop:online}
Under the incremental paradigm, for every sequence of changes and every wall-clock budget, the committed configurations form a diamond plan from $S^\eps$, and the reference trajectories of the robots are its reconstruction. The planned motion is therefore admissible and conflict-free at all times. 
\end{proposition}
\begin{proof}
Every candidate set contains the wait, and the proof of \cref{lem:tick-validity} uses no other candidate. The lemma therefore gives $P^{j+1}\in\mathcal Q_D$ and elementary moves for every committed index $j$, and the sequence is a valid plan by \cref{prop:diamond-certified}. At every time the robots need the sequence up to the frontier only, so their reference trajectories are the reconstruction of the committed sequence, and \cref{thm:soundness} applies.
\end{proof}

Under the whole-fleet paradigm, the retained part of the plan and the new plan are valid and share $P^{J_s}$, so the same argument certifies the planned motion, and a change takes effect at the splice. A keep-out region adds one guarantee. A robot that is not halted has every vertex from the index $k_t-W/2$ on in $V_D(R)$, the committed ones by the halting test and the later ones by the restriction of its candidates. From the time at which the region takes effect, its reconstruction averages these vertices only. The obstacle part of \cref{thm:soundness}, applied with $R$ added to the obstacles, then shows that every robot that is not halted keeps a clearance above $r$ from $R$ until the region is removed. 

\subsection{Real-Time Sessions}\label{app:online}

\begin{figure*}[t]
\centering
\includegraphics[width=\textwidth]{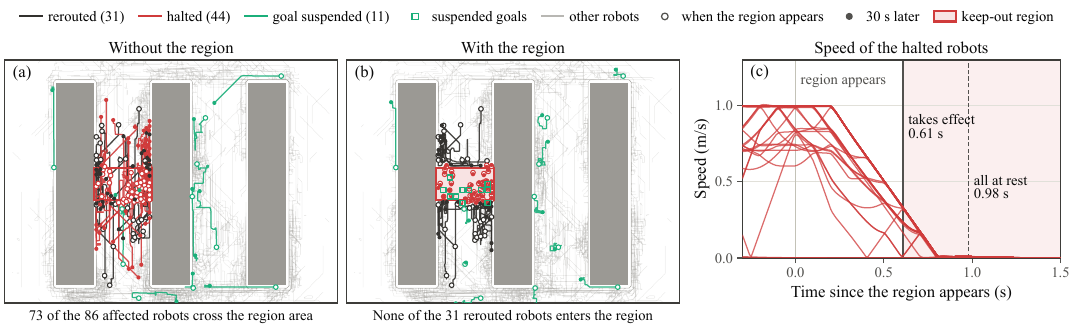}
\caption{Reaction to a keep-out region in the incremental session of \cref{fig:exp-online}(a) with 771 simulated RB-1 Base robots in the rack hall, in which the region appears at 60\,s. (a), (b) Paths, over the 30\,s after the region appears, of the 86 robots that the region affects, in the same session without and with the region. (c) Speed of the halted robots, with the time at which the region takes effect and the time at which all of them are at rest.}\label{fig:online-region}
\end{figure*}

\myparagraph{Fleets and Changes} --
The fleets are instances of \cref{sec:exp-scalability}, and the fleets of 100 and 200 robots are the first 100 and 200 robots of such instances. Each of the ten instances of a fleet has one script of changes with random times, and both paradigms follow the same scripts. In the rack hall and in the empty hall, a band of 18.4\,m by 10\,m appears during the motion and stands for one to two minutes, and every robot later receives its start position as its new goal. In the rack hall, the band closes one of the two aisles between the racks. In the central pillar volume, a box of 2\,m by 2\,m over the full height stands against one side of the pillar for 8 to 40\,s, and every quadrotor later receives its start position as its new goal.

The whole-fleet sessions plan and replan with \acrshort{acr:takt}-PIBT. Each session advances the clock of the fleet by one tick per committed configuration and measures the computation time of every decision cycle and replan on the laptop of \cref{sec:exp-setup}.

\myparagraph{Computation Time per Change and Budgets} --
For each change, we report the computation time of the first configuration that accounts for it. Under the incremental paradigm, it is the time of one decision cycle of \cref{alg:cycle}, including the processing of the change. Under the whole-fleet paradigm, it is the time of one replan, which prepares the instance of the robots that are not halted and runs the planner on it. The wall-clock budget of a cycle is 80\,ms in 2D and 35\,ms in 3D. When less than 20\,ms in 2D or 10\,ms in 3D remain before the end of the tick, the robots whose candidate order has not yet been computed rank their candidates by the Manhattan distance to the goal in that tick, which requires no distance query and keeps the configuration valid.

\begin{table}[t]
\centering\footnotesize
\caption{Computation times against the fleet size, ten sessions per row. Incremental paradigm: mean, 99th percentile, and longest decision cycle, against a tick of 101\,ms in 2D and 43\,ms in 3D. Whole-fleet paradigm: initial plan, and mean and longest replan over the replans after a new keep-out region, after its removal, and after new goals, against the criterion of 0.5\,s.}\label{tab:online-scale}
\setlength{\tabcolsep}{3pt}
\begin{tabular}{@{}lrrrrrrr@{}}
\toprule
Map & $N$ & \multicolumn{3}{c}{Incremental cycle (ms)} & \multicolumn{3}{c}{Whole-fleet (s)}\\
\cmidrule(lr){3-5}\cmidrule(l){6-8}
 & & Mean & 99th & Longest & Initial & Mean & Longest\\
\midrule
Rack hall & 100 & 0.04 & 0.36 & 2.8 & 0.07--0.09 & 0.07 & 0.19\\
Rack hall & 200 & 0.09 & 0.67 & 3.4 & 0.17--0.22 & 0.15 & 0.36\\
Rack hall & 385 & 0.19 & 0.82 & 3.7 & 0.41--0.50 & 0.38 & 0.74\\
Rack hall & 578 & 0.30 & 1.15 & 4.2 & 0.76--0.88 & 0.66 & 1.21\\
Rack hall & 771 & 0.41 & 1.33 & 4.9 & 1.21--1.39 & 1.03 & 1.83\\
Empty hall & 254 & 0.05 & 0.08 & 3.3 & 0.12--0.13 & 0.09 & 0.14\\
Empty hall & 509 & 0.12 & 0.19 & 4.2 & 0.21--0.24 & 0.16 & 0.26\\
Empty hall & 763 & 0.21 & 0.35 & 5.6 & 0.35--0.40 & 0.26 & 0.46\\
Empty hall & 1,018 & 0.31 & 0.62 & 7.0 & 0.52--0.55 & 0.41 & 0.79\\
Pillar & 100 & 0.11 & 0.73 & 29.5 & 0.04--0.09 & 0.07 & 0.19\\
Pillar & 200 & 0.24 & 1.32 & 30.5 & 0.11--0.19 & 0.12 & 0.30\\
Pillar & 335 & 0.48 & 2.35 & 33.4 & 0.22--0.38 & 0.22 & 0.44\\
Pillar & 671 & 0.98 & 3.95 & 34.1 & 0.77--1.29 & 0.50 & 1.38\\
\bottomrule
\end{tabular}
\end{table}

\myparagraph{Computation Times} --
\Cref{tab:online-scale} gives the mean and the longest computation times of the fleets in \cref{fig:exp-online}(c)--(e). At every fleet size, the mean decision cycle takes at most 1\,ms and the 99th percentile at most 4\,ms, against ticks of 101\,ms and 43\,ms. The longest cycles are those that read a new keep-out region or its removal. In 3D, the processing of a new no-fly zone consists mostly of the construction of its lattice vertices, and the generator decides every quadrotor within the rest of the 43\,ms tick.

Under the whole-fleet paradigm, every session splices three replans into the motion, one per change, and \cref{tab:online-scale} reports these 30 replans per fleet. In every session, the fleet comes to rest.

\begin{table}[t]
\centering\footnotesize
\caption{Reaction to the changes in the sessions with 200 robots, ten sessions per row, with the same changes under both paradigms. Computation time: that of the first configuration that accounts for a new keep-out region or for the new goals. Effect: time from the appearance of a keep-out region to the time at which it takes effect.}\label{tab:exp-online}
\setlength{\tabcolsep}{3.5pt}
\begin{tabular}{@{}lrrr@{}}
\toprule
Paradigm & \multicolumn{2}{c}{Computation time (ms)} & Effect\\
\cmidrule(lr){2-3}
 & Keep-out & New goals & (s)\\
\midrule
2D, incremental & 2.0--3.4 & 0.04--0.06 & 0.62--0.69\\
2D, whole-fleet & 57--261 & 146--359 & 0.94--1.10\\
3D, incremental & 29.3--30.5 & 0.07--0.79 & 0.39--0.43\\
3D, whole-fleet & 117--143 & 83--304 & 0.62--0.77\\
\bottomrule
\end{tabular}
\end{table}

\myparagraph{Time to Take Effect} --
\Cref{tab:exp-online} compares the two paradigms on the sessions with 200 robots. Under the incremental paradigm, a keep-out region takes effect between $(L+1)\dt$ and $(L+2)\dt$ after it appears at every fleet size, which is 0.62 to 0.69\,s in the rack hall and 0.39 to 0.43\,s in the central pillar volume. Under the whole-fleet paradigm, every replan after a new region returns before its splice, and the region takes effect at the splice, 0.9 to 1.1\,s after it appears in the rack hall and 0.6 to 0.8\,s after it appears in the central pillar volume.

\myparagraph{Reaction to a Keep-Out Region} --
\Cref{fig:online-region} shows the session of \cref{fig:exp-online}(a) and (b) with 771 robots, executed by the RB-1 Base models of \cref{sec:exp-tracking}. In the same session without the region, 73 robots pass through the area of the band in the 30\,s after it appears. With the region, the planner halts the 44 robots whose committed configurations meet the band, suspends the goals of 11 additional robots, and routes 31 robots around the band. The band takes effect 0.61\,s after it appears, and the halted robots are at rest 0.98\,s after it appears. From the time at which the band takes effect, no robot outside the halted set enters the band, and the closest body stays 32\,cm away from it.

\myparagraph{Execution by Simulated Robots} --
The RB-1 Base and Crazyflie models of \cref{sec:exp-tracking} executed six recorded sessions with their controllers unchanged. The six sessions are one session per paradigm and dimension with 200 robots and the incremental sessions with 771 robots and 335 quadrotors. The tracking errors stay within 0.72\,cm for the bases and 5.1\,cm for the quadrotors, and no unhalted robot enters the region after it takes effect.
Under the incremental paradigm, the halted robots come to rest within 0.98\,s in 2D and 1.03\,s in 3D.

\end{document}